\documentclass[runningheads]{llncs}

\usepackage{eccv}

\usepackage{eccvabbrv}

\usepackage{graphicx}
\usepackage{booktabs}

\usepackage[accsupp]{axessibility}  

\usepackage{hyperref}

\usepackage{orcidlink}

\begin{document}

\title{COMPOSE: Comprehensive Portrait Shadow Editing} 


\author{Andrew Hou\inst{1,2}\thanks{This work was done while Andrew was an intern at Adobe Research.} \and
Zhixin Shu\inst{2} \and Xuaner Zhang\inst{2} \and He Zhang\inst{2} \and Yannick Hold-Geoffroy\inst{2} \and Jae Shin Yoon\inst{2} \and
Xiaoming Liu\inst{1}}

\authorrunning{A. Hou et al.}

\institute{Michigan State University \\ \email{\{houandr1,liuxm\}@msu.edu} 
\and Adobe Research \\
\email{\{zshu,cezhang,hezhan,holdgeof,jaeyoon\}@adobe.com}}

\newcommand{\vardbtilde}[1]{\tilde{\raisebox{0pt}[0.85\height]{$\tilde{#1}$}}}
\newcommand{\defeq}{\coloneqq}
\newcommand{\grad}{\nabla}
\newcommand{\E}{\mathbb{E}}
\newcommand{\Var}{\mathrm{Var}}
\newcommand{\Cov}{\mathrm{Cov}}
\newcommand{\Ea}[1]{\E\left[#1\right]}
\newcommand{\Eb}[2]{\E_{#1}\!\left[#2\right]}
\newcommand{\Vara}[1]{\Var\left[#1\right]}
\newcommand{\Varb}[2]{\Var_{#1}\left[#2\right]}
\newcommand{\kl}[2]{D_{\mathrm{KL}}\!\left(#1 ~ \| ~ #2\right)}
\newcommand{\pdata}{{p_\mathrm{data}}}
\newcommand{\bA}{\mathbf{A}}
\newcommand{\bI}{\mathbf{I}}
\newcommand{\bJ}{\mathbf{J}}
\newcommand{\bH}{\mathbf{H}}
\newcommand{\bL}{\mathbf{L}}
\newcommand{\bM}{\mathbf{M}}
\newcommand{\bQ}{\mathbf{Q}}
\newcommand{\bR}{\mathbf{R}}
\newcommand{\bzero}{\mathbf{0}}
\newcommand{\bone}{\mathbf{1}}
\newcommand{\bb}{\mathbf{b}}
\newcommand{\bu}{\mathbf{u}}
\newcommand{\bv}{\mathbf{v}}
\newcommand{\bw}{\mathbf{w}}
\newcommand{\bx}{\mathbf{x}}
\newcommand{\by}{\mathbf{y}}
\newcommand{\bz}{\mathbf{z}}
\newcommand{\bxh}{\hat{\mathbf{x}}}
\newcommand{\btheta}{{\boldsymbol{\theta}}}
\newcommand{\bphi}{{\boldsymbol{\phi}}}
\newcommand{\bepsilon}{{\boldsymbol{\epsilon}}}
\newcommand{\bmu}{{\boldsymbol{\mu}}}
\newcommand{\bnu}{{\boldsymbol{\nu}}}
\newcommand{\bSigma}{{\boldsymbol{\Sigma}}}

\maketitle

\begin{abstract}
  Existing portrait relighting methods struggle with precise control over facial shadows, particularly when faced with challenges such as handling hard shadows from directional light sources or adjusting shadows while remaining in harmony with existing lighting conditions.
In many situations, completely altering input lighting 
is undesirable for portrait retouching applications: one may want to preserve some authenticity in the captured environment. 
Existing shadow editing methods typically restrict their application to just the facial region and often offer limited lighting control options, such as shadow softening or rotation.
In this paper, we introduce COMPOSE: a novel shadow editing pipeline for human portraits, offering precise control over shadow attributes such as shape, intensity, and position, all while preserving the original environmental illumination of the portrait. This level of disentanglement and controllability is obtained thanks to a novel decomposition of the environment map representation into ambient light and an editable gaussian dominant light source. 
COMPOSE is a four-stage pipeline that consists of light estimation and editing, light diffusion, shadow synthesis, and finally shadow editing. 
We define facial shadows as the result of a dominant light source, encoded using our novel gaussian environment map representation. Utilizing an OLAT dataset, we have trained models to: (1) predict this light source representation from images, and (2) generate realistic shadows using this representation. We also demonstrate comprehensive and intuitive shadow editing with our pipeline.  Through extensive quantitative and qualitative evaluations, we have demonstrated the robust capability of our system in shadow editing.
\end{abstract}

\begin{figure}
\begin{center}
\begin{minipage}[t]{0.19\linewidth}
\centering
\includegraphics[width=\linewidth]{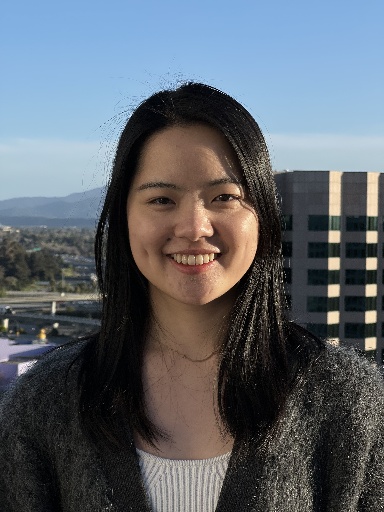} \\
\small a) Source Image
\end{minipage}
\begin{minipage}[t]{0.19\linewidth}
\centering
\includegraphics[width=\linewidth]{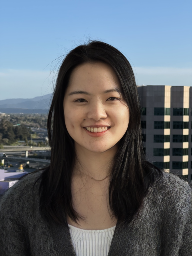} \\
\small b) Soften Shadow
\end{minipage}
\begin{minipage}[t]{0.19\linewidth}
\centering
\includegraphics[width=\linewidth]{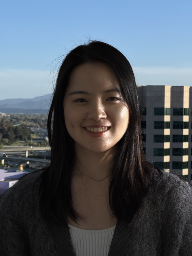} \\
\small c) Intensify Shadow
\end{minipage}
\begin{minipage}[t]{0.19\linewidth}
\centering
\includegraphics[width=\linewidth]{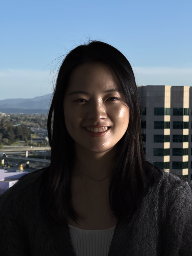} \\
\small d) Modify Light Size
\end{minipage}
\begin{minipage}[t]{0.19\linewidth}
\centering
\includegraphics[width=\linewidth]{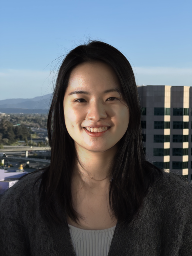} \\
\small e) Rotate Shadow
\end{minipage}


\end{center}
\vspace{-5mm}
\small \caption{\textbf{Overview}. COMPOSE is the first single-image portrait shadow editing method that achieves complete shadow editing control, {\it i.e.}~adjusting shadow intensity, modifying light size, and changing shadow positions, all while preserving the source image's other lighting attributes (\textit{e.g.} ambient light).}
\label{fig:Overview} 
\vspace{-5mm}
\end{figure}

\section{Introduction}
\label{sec:intro}
Single-image portrait relighting is a well-studied problem with steady interest from the computer vision and graphics communities. Portrait relighting has a wide range of applications in computational photography~\cite{TotalRelighting}, AR/VR~\cite{SIRA, RANA, bi2021avatar}, and downstream tasks in the face domain~\cite{controllable-light-diffusion,HaLeWACV19}. 
While many methods for single image portrait relighting have been proposed in recent years~\cite{PhysicsGuidedRelighting,UCSDSingleImagePortraitRelighting,TotalRelighting,DPR,diffusionfacerelighting,face-relighting-with-geometrically-consistent-shadows, towards-high-fidelity-face-relighting-with-realistic-shadows, yeh2022learning}, 
they often face challenges when dealing with facial shadows.

One popular genre of face relighting methods~\cite{TotalRelighting, UCSDSingleImagePortraitRelighting} uses low resolution HDR environment maps to represent lighting. This lighting representation is used in a conditional image translation framework to control the lighting effect of the output face image, thereby achieving facial relighting.
The requirement for an environment map restricts users seeking to fine-tune lighting, such as minor modifications to shadow smoothness or direction. Additionally, a low-resolution environment map often lacks the precision needed for accurately depicting complex, high-frequency shadow effects. Conversely, several approaches use low-dimensional spherical harmonics (SH) for their lighting representation~\cite{DPR,SfSNet}. These methods, based on simplified lighting and reflectance models, frequently produce unrealistic relit images. Moreover, they overlook occlusions in the image formation process, disallowing the generation and editing of cast shadows.
SunStage~\cite{wang2023sunstage} generates realistic shadows on faces but lacks generalizability and requires videos as input.
Some recent works have focused on shadow editing tasks~\cite{controllable-light-diffusion, PortraitShadowManipulation, blind-removal-of-facial-foreign-shadows} using state-of-the-art image generators such as diffusion models. However, achieving precise control with these sophisticated generators remains a challenging task. Consequently, these methods often provide limited control in shadow synthesis, typically handling only shadow softening or completely removing the shadow, without precise editing of shadow intensity, shape, or position. 

To overcome the challenges in modeling and controlling facial shadows in portrait relighting, we introduce COMPOSE, a system that separates shadow modeling from ambient light in the lighting representation. COMPOSE enables flexible manipulation of shadow attributes, including position (lighting direction), smoothness, and intensity (See Fig.~\ref{fig:Overview}). The key innovation in COMPOSE is its shadow representation, which is compatible with environment map-based representations, making it easily integrable into existing relighting networks.
Building on the concept of image-based lighting representations (\textit{e.g.} HDR environment maps), our approach divides lighting effects into two components: shadows caused by dominant light sources and ambient lighting. We simplify the shadow component by attributing it to a single dominant light source on an environment map. This is encoded using a 2D isotropic Gaussian, with variables representing light position, standard deviation (indicating light size or diffusion), and scale (light intensity). The remaining lighting effects are attributed to a diffused environment map, modeling the image's ambient lighting.

For a given face image, we first predict the lighting as a panorama environment map, from which we estimate the dominant light (shadow) parameters. The system then removes shadows to obtain a diffused, albedo-like image.  Utilizing this diffuse image and the shadow representation, we train a network to synthesize shadows on the subject in the input image. This shadow synthesis process is adaptable, accepting various shadow-related parameters for controllable shadow synthesis, including position and shape. The synthesized shadow can be linearly blended with the diffuse lighting to create shadows of varying intensities.

To train models for shadow diffusion and synthesis, we utilize an OLAT (one-light-at-a-time) dataset, captured through a light stage system. This dataset, combined with the environment map-based lighting representation, ensures our system's compatibility with previously proposed face relighting techniques. Our approach also uniquely offers the capability to accurately model shadows.

Our contributions are thus as follows: 

$\diamond$ We propose the first single image portrait shadow editing method with precise control over various shadow attributes, including shadow shape, intensity, and position. Our system provides users with the flexibility to control shading while also preserving the other lighting attributes of the input images.

$\diamond$ We propose a novel lighting decomposition into ambient light and an editable dominant gaussian light source, which enables disentangled shadow editing while preserving the remaining lighting attributes. 

$\diamond$ We achieve state-of-the-art relighting performance quantitatively and qualitatively, including on a light-stage-captured test set and in-the-wild images. 

\section{Related Work}
\label{sec:relatedwork}

\subsection{Portrait Relighting}
Many portrait relighting methods have been proposed in recent years that generally fall into one of two categories: focusing on reducing the data requirement and improving the relighting quality as much as possible with limited data~\cite{yeh2022learning,face-relighting-with-geometrically-consistent-shadows,towards-high-fidelity-face-relighting-with-realistic-shadows,DPR,diffusionfacerelighting,SfSNet,FaceLit,ding2023diffusionrig,wang2023sunstage, Flickr,SIPRExplicitMultiple,delighting,MassTransport,temporallyconsistentvideorelighting,lightstagedesk,volux-gan,geometryawarehumanrelighting,Lagunas2021humanrelighting} or using light stage data but modeling more complex lighting effects~\cite{TotalRelighting, PhysicsGuidedRelighting, UCSDSingleImagePortraitRelighting,lightpainter,therelightables,neuralvideorelighting,NLT,bi2021avatar,sun2021nelf,BareSkinNet}. 
Among single image portrait relighting methods, a substantial limitation is that they cannot preserve the original environmental lighting while editing shadows, and instead opt to relight by replacing the original lighting with a completely new environment~\cite{controllable-light-diffusion}, which is undesirable in many portrait editing applications. 
Existing methods also struggle to handle more directed lights and fail in the presence of substantial hard or intense shadows~\cite{DPR,SfSNet,UCSDSingleImagePortraitRelighting}. 

Among portrait relighting methods that take multiple images or videos as input, the method SunStage~\cite{wang2023sunstage} is able to maintain the lighting attributes of the input video's environment when performing relighting. However, the method must be trained using a video of a subject on a sunny day and involves an expensive optimization procedure. The model is also subject-specific and must be retrained or fine-tuned even for the same subject if they change their hairstyle, accessories, or clothing. COMPOSE has the advantage of being a single-image portrait relighting method, and is both generalizable and practical for in-the-wild photo editing applications. 

\begin{figure*}[t]
\begin{center}
   \includegraphics[width=1.\linewidth]{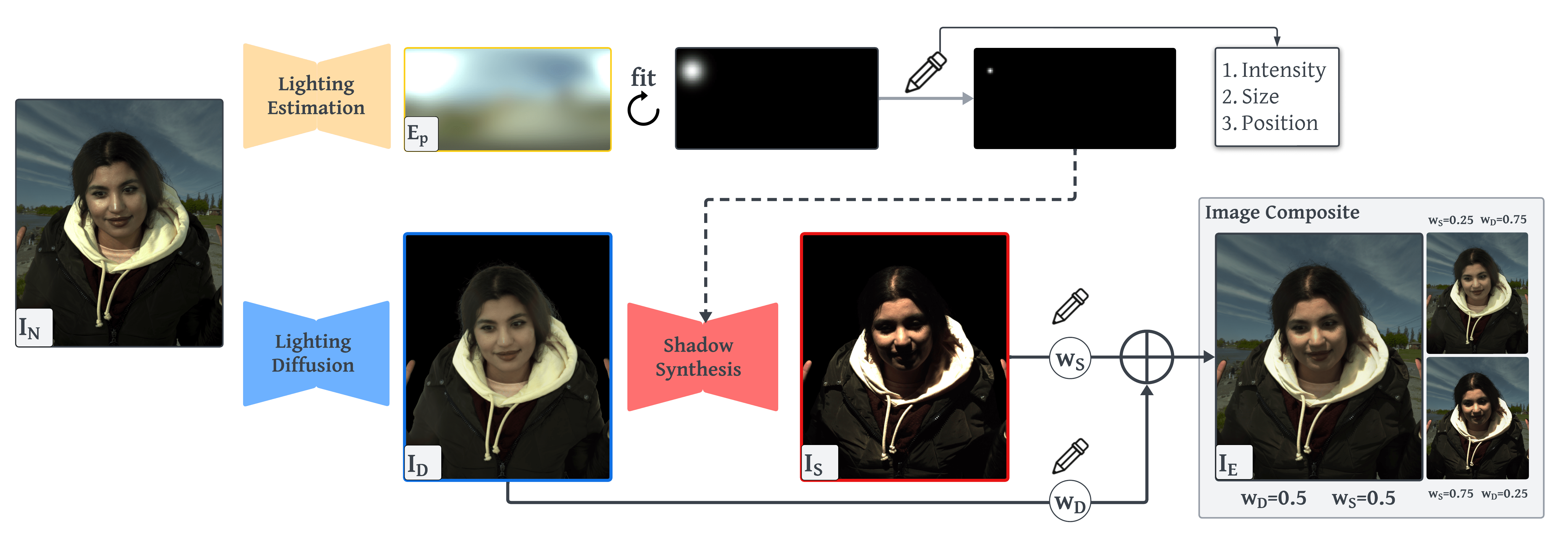} \\
\caption{\small \textbf{Method Overview}. COMPOSE is a $4$-stage shadow editing method consisting of single-image lighting estimation, light diffusion, shadow synthesis, and image compositing. By first estimating the dominant light position using the environment map regressed from the input image, COMPOSE can control shadow shape and intensity by controlling light spread and light intensity respectively as well as shadow position by changing the location of the dominant light. By estimating a diffuse image $\mathbf{I}_{D}$ as well as a shadowed image $\mathbf{I}_{S}$, COMPOSE can generate the final edited image $\mathbf{I}_{E}$ through image compositing. }
\label{fig:Architecture}
\end{center} 
\vspace{-8mm}
\end{figure*}


\subsection{Shadow Editing}
Another line of work closely related to COMPOSE are methods that edit only facial shadows, including shadow removal~\cite{blind-removal-of-facial-foreign-shadows, PortraitShadowManipulation} and shadow softening~\cite{controllable-light-diffusion}. Most similar to our work is the method of~\cite{controllable-light-diffusion}, which is able to directly control the degree of shadow softening while preserving the original environmental lighting. 
However, unlike COMPOSE, they are restricted to only shadow softening and cannot handle other shadow editing operations like shadow intensifying, modifying shadow shape, and shadow rotation. 
Shadow removal methods~\cite{blind-removal-of-facial-foreign-shadows, PortraitShadowManipulation,he21unsupervised} are often even more limited since they cannot control the degree of shadow softening and attempt to remove the shadow completely. 

\subsection{Lighting Representations}
Existing portrait relighting methods generally utilize one of three lighting representations: Spherical Harmonics (SH)~\cite{DPR, SfSNet, FaceLit, towards-high-fidelity-face-relighting-with-realistic-shadows, diffusionfacerelighting}, dominant lighting direction~\cite{face-relighting-with-geometrically-consistent-shadows, PhysicsGuidedRelighting}, and environment maps~\cite{UCSDSingleImagePortraitRelighting, TotalRelighting, yeh2022learning}. SH lighting tends to only model the lower frequencies of lighting effects, and thus is unsuitable for modeling non-Lambertian hard shadows and specularities and is better suited for diffuse lightings. Representing the light as a dominant lighting direction can often handle the non-Lambertian effects of directional and point light sources, but they often leave other lighting attributes such as light size out of the equation (\textit{e.g.} Hou \textit{et al.}~\cite{face-relighting-with-geometrically-consistent-shadows}). Moreover, they often do not handle diffuse lightings well and are intended for directional lights~\cite{face-relighting-with-geometrically-consistent-shadows, PhysicsGuidedRelighting}. The environment map representation is more flexible and is able to model light intensity, light size, and light position and can handle both diffuse and directional lights. However, what is largely missing in this domain is a way to properly decompose the environment map that enables precise controllability over different lighting components. Existing methods only perform relighting by completely changing the scene illumination with a new environment map without disentangling the effects of ambient, diffuse, and directional light~\cite{TotalRelighting, UCSDSingleImagePortraitRelighting, yeh2022learning}. This can be undesirable for many computational photography applications where the user wishes to preserve the ambience of the scene and edit only the facial shadow alone~\cite{controllable-light-diffusion}. Moreover, the default environment map representation is unsuitable for some shadow editing applications such as shrinking the light. While enlarging the light can be performed by applying a gaussian blur, there is no equivalent operation that can be performed to shrink the light on an environment map. Our lighting representation decomposes the environment map into the ambient light and an editable gaussian dominant light source, which is adaptable in light intensity, size, and position. COMPOSE is thus able to perform a full suite of shadow editing operations thanks to its editable gaussian component and can edit the shadow alone without disrupting other lighting attributes (\textit{e.g.} ambient light).             

\section{Methodology}

\subsection{Problem Formulation}

In our framework, shadows are treated as a composable lighting effect, which can be independently predicted and altered based on a specific face image. This shadow representation is integrated into the overall lighting representation, and our framework is designed to (1) accurately predict the shadow from an image, and (2) apply a controllable shadow onto a shadow-free face image.
We define the portrait shadow editing problem based on properties of shadows that users can manipulate, namely shadow position, shadow intensity, and shadow shape. These shadow properties are connected to the lighting attribute that is easily controllable. Specifically, 
given source image $\mathbf{I}_{N}$ and desired edited output image $\mathbf{I}_{E}$ with modified shadows, our problem can be written as: 
\begin{equation}
\begin{split}
    \mathbf{I}_{E}=F_{\theta}(\mathbf{I}_{N}, x, y, \sigma, \gamma),
\end{split} 
\end{equation}
where $F_{\theta}$ is our shadow editing method, $x$ and $y$ are the desired light position on an environment map, $\sigma$ is the light size, and $\gamma$ is the light intensity. 
By editing the parameters, ($x$, $y$), $\sigma$, and $\gamma$, we can control the shadow position, shape, and intensity respectively. 
We describe our solution COMPOSE in detail in Sec.~\ref{sec:framework}. 

\subsection{COMPOSE Framework}
\label{sec:framework}
COMPOSE is a $4$-stage shadow editing framework, consisting of single image lighting estimation, light diffusion, shadow synthesis, and finally image compositing. See Fig.~\ref{fig:Architecture} for an overview of our method. 
\vspace{-5mm}
\subsubsection{Lighting Estimation} In the lighting estimation stage, we train a variational autoencoder (VAE)~\cite{VAE} to estimate an environment map (LDR) from a single portrait image as its lighting representation $\mathbf{I}_{N}$. From this environment map, we then estimate the dominant light position. Naturally, the dominant light, such as the sun, represents the brightest intensity on a image-based lighting representation. Therefore we can simply obtain the dominant light position by fitting a 2D isotropic Gaussian on the predicted environment map. The center of the Gaussian will indicate the position of the dominant light on the 2D environment map. 
This estimated light position corresponds to the shadow information on the input image, and therefore is useful for ``in-place'' shadow editing (\textit{e.g.}, shadow softening, shadow intensifying) in the later stage that does not alter the direction of the dominant light. We argue this is desirable in many portrait editing applications where the user only wants to edit existing shadows and does not want to alter the ambient lighting.
\vspace{-5mm}
\subsubsection{Light Diffusion} In the light diffusion stage, we remove all existing hard shadows and specular highlights from the input image and output a ``diffused'' image $\mathbf{I}_{D}$ with very smooth shading. This represents the input face under only an ambient illumination condition. Our light diffusion network consists of a hierarchical transformer encoder that generates multi-level features, which are fed to a decoder with transposed convolutional layers. Our network takes as input the original input image, a body parsing mask, and a binary foreground mask. 
\vspace{-5mm}
\subsubsection{Shadow Synthesis} 
\begin{figure*}[t]
\begin{center}
   \includegraphics[width=1.\linewidth]{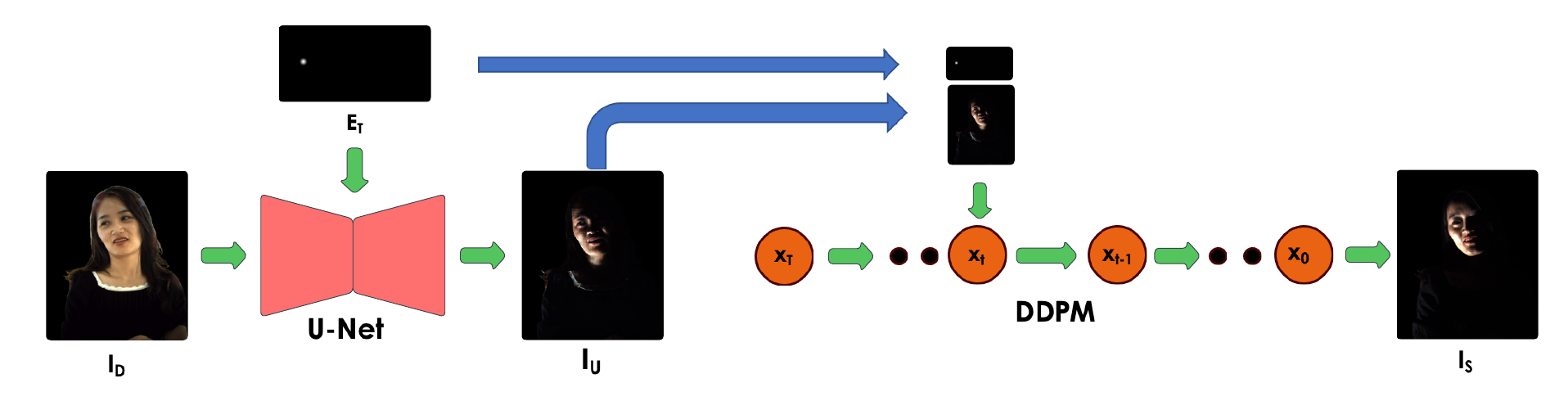} \\
   \vspace{-4mm}
\caption{\small \textbf{Shadow Synthesis Pipeline}. Our shadow synthesis step consists of two stages: a U-Net followed by a conditional DDPM. The U-Net takes ambient image $\mathbf{I}_{D}$ and environment map $\mathbf{E}_{T}$ as input and outputs the coarse relit image $\mathbf{I}_{U}$. With $\mathbf{I}_{U}$ and $\mathbf{E}_{T}$ as spatial conditions, the conditional DDPM outputs the final shadowed image $\mathbf{I}_{S}$, refining shadow boundaries and improving image quality. }
\label{fig:ShadowSynthesis}
\end{center} 
\vspace{-8mm}
\end{figure*}

In the shadow synthesis stage, we aim to produce a shadowed image $\mathbf{I}_{S}$ illuminated by an edited light source given the diffuse image $\mathbf{I}_{D}$ estimated from the light diffusion stage and an edited environment map. 
Using the estimated dominant light position from the light estimation stage, we have full control over light attributes such as light size, intensity, and the dominant light position. 
We can produce a new environment map using a Gaussian to represent the dominant light, where the light size is modeled by the standard deviation of the Gaussian (a larger standard deviation represents a larger, more diffuse light) and the light intensity can be adjusted by multiplying the Gaussian with a scalar. This allows our method to control shadow shape, light intensity, and position. To regress the shadowed image $\mathbf{I}_{S}$, we propose a two-stage architecture: a U-Net followed by a conditional DDPM (See Fig.~\ref{fig:ShadowSynthesis}). For the first stage, we adopt the U-Net architecture of~\cite{UCSDSingleImagePortraitRelighting}.
During training, instead of using a reshaped environment map-like representation for the dominant light, we design a feature map-like representation that's parameterized by the shadow attributes.
Specifically, the lighting representation for shadows has 4 parameters.
The first two parameters $x$ and $y$ encode the coordinates of the light center on the environment map, the third parameter $\sigma$ encodes the light size, and the fourth parameter $\gamma$ encodes the light intensity. All four parameters are normalized between $0$ and $1$ and repeated spatially as $32\times32$ channels to be suitable inputs for the U-Net. We find that this lighting representation leads to faster convergence, especially when the desired light mimics a point light and occupies a tiny portion of the environment map. 
We feed the diffuse image $\mathbf{I}_{D}$ and our desired lighting feature map to the U-Net, and the output is a relit image $\mathbf{I}_{U}$. While the U-Net architecture has merits such as complete geometric consistency for shadow synthesis, one demerit is that the synthesized shadow boundaries are often not very sharp and the overall image quality could be improved. We thus employ a DDPM~\cite{DDPM,ilvr_adm} as the second stage of our shadow synthesis model. The role of the DDPM here is to take the output image of the U-Net $\mathbf{I}_{U}$ as a spatial condition along with the lighting parameters and perform image refinement to generate the final shadowed image $\mathbf{I}_{S}$. The training objective of our shadow synthesis DDPM thus becomes: 
\begin{equation}
 \Eb{t, \bx_0, \bepsilon}{ \left\| \bepsilon - \bepsilon_\theta(\mathbf{x}_{t}, t, \mathbf{I}_{U}, x, y, \sigma, \gamma) \right\|^2}. \label{eq:training_objective_conditional_ours}
\end{equation}
Our condition $\mathbf{I}_{U}$ is spatially concatenated with $\mathbf{x}_{t}$ and our lighting parameters are repeated spatially as channels of the same resolution and similarly spatially concatenated, where
 $\mathbf{x}_{t}= \sqrt{\bar\alpha_t}\bx_0 + \sqrt{1-\bar\alpha_t}\bepsilon$. We find that the quality of the edited images is noticeably enhanced by the DDPM compared to the U-Net alone and that adding the DDPM sharpens the shadow boundaries.
\vspace{-5mm}

\subsubsection{Image Compositing} With $\mathbf{I}_{D}$ from our light diffusion stage and $\mathbf{I}_{S}$ from our shadow synthesis stage, we can then perform image compositing between $\mathbf{I}_{D}$ and $\mathbf{I}_{S}$ to achieve our final edited image $\mathbf{I}_{E}$, by assigning weights $\omega_{D}$ and $\omega_{S}$ to $\mathbf{I}_{D}$ and $\mathbf{I}_{S}$ respectively. 
The final edited image $\mathbf{I}_{E}$ is thus generated as: 
\begin{equation}
\begin{split}
\mathbf{I}_{E}=\omega_{D}\mathbf{I}_{D}+\omega_{S}\mathbf{I}_{S},
\end{split} 
\end{equation}
where $\omega_{D}+\omega_{S}=1$. By adjusting $\omega_{D}$ and $\omega_{S}$, we can further tune shadow intensity: a higher $\omega_{D}$ softens shadows and a higher $\omega_{S}$ results in darker shadows. 

\subsection{Loss Functions}
To train our single image lighting estimation network, we apply two loss functions commonly used in VAE training~\cite{VAE}: $\mathcal{L}_\text{recon}$ and $\mathcal{L}_\text{KLD}$. $\mathcal{L}_\text{recon}$ is a reconstruction loss between the predicted environment map $\mathbf{E}_{P}$ and the groundtruth environment map $\mathbf{E}_{G}$ defined as follows: 
\begin{equation}
    \mathcal{L}_\text{recon} = \frac{1}{3HW}(\|\mathbf{E}_{P}-\mathbf{E}_{G}\|^{2}_\text{2}),
\end{equation}
where $H$ and $W$ are the size of the environment maps. 
The Kullback–Leibler divergence (KLD) loss $\mathcal{L}_\text{KLD}$ is: 
\begin{equation}
    \mathcal{L}_\text{KLD} = -\frac{1}{2N}\sum\limits_{i=1}^{N}(1+\mbox{log}(\sigma_{i}^{2})-\mu_{i}^{2}-\sigma_{i}^{2}),
\end{equation}
where $N$ is the dimensionality of the latent vector, $\mu$ is the batch mean, and $\sigma^{2}$ is the batch variance. The final loss for our lighting estimation network is thus: 
\begin{equation}
    \mathcal{L}_\text{LE} = \lambda_{1}\mathcal{L}_\text{recon}+\lambda_{2}\mathcal{L}_\text{KLD},
\end{equation}
where $\lambda_{1}=1$ and $\lambda_{2}=2.5\times10^{-4}$. 

Our light diffusion network is also trained with two losses $\mathcal{L}_\text{recon}$ and $\mathcal{L}_\text{perceptual}$. 
\begin{equation}
    \mathcal{L}_\text{recon} = \frac{1}{3HW}(\|\mathbf{I}_{D}-\mathbf{I}_{G}\|_\text{1}),
\end{equation}
where $H$ and $W$ are the size of the training images, $\mathbf{I}_{D}$ is the predicted diffuse image, and $\mathbf{I}_{G}$ is the groundtruth diffuse image. $\mathcal{L}_\text{perceptual}$ enforces visual similarity~\cite{LPIPS} and is computed as the distance between VGG~\cite{VGG} features computed for $\mathbf{I}_{D}$ and $\mathbf{I}_{G}$. The final loss for the light diffusion network is thus: 
\begin{equation}
    \mathcal{L}_\text{LD} = \mathcal{L}_\text{recon}+\mathcal{L}_\text{perceptual}.
\end{equation}

When training the U-Net of our shadow synthesis pipeline, we employ a single reconstruction loss $\mathcal{L}_\text{recon}$ between the predicted relit image $\mathbf{I}_{S}$ and the groundtruth $\mathbf{I}_{G}$, defined as follows:  
\begin{equation}
    \mathcal{L}_\text{recon} = \frac{1}{3HW}(\|\mathbf{I}_{S}-\mathbf{I}_{G}\|^{2}_\text{2}).
\end{equation}



\subsection{Data Collection and Generation}
To train out networks, we use a light stage to capture one-light-at-a-time (OLAT) images of $107$ subjects with diverse skin tones, body poses, expressions, and accessories using a light stage~\cite{paullightstage} with $160$ lights. Each subject is captured under $10$-$20$ sessions with varied body poses, expressions, and accessories. Each session is captured by $4$ cameras with varying camera poses. 

To generate input images to train our lighting estimation VAE and our light diffusion network, we render our OLAT images using a diverse set of environment maps collected from the Polyhaven and Laval Outdoor HDR datasets~\cite{LavalOutdoor}. During rendering, we apply random augmentations to the environment map including rotations and adjusting intensity to improve our model's generalization to different lighting. 
The groundtruth $\mathbf{E}_{G}$ for our lighting estimation VAE is simply the environment map used for rendering, and the groundtruth $\mathbf{I}_{G}$ for the light diffusion network is generated by rendering OLAT images with heavily blurred, diffuse environment maps. 

To generate images for our shadow synthesis stage, we render groundtruth relit images by using environment maps consisting of Gaussian lights with various positions, intensities, and sizes (See Sec.~\ref{sec:framework}) along with our OLAT images. As the input to the shadow synthesis stage is meant to be the diffuse image $\mathbf{I}_{D}$ predicted by the light diffusion stage, we generate diffuse images for training our shadow synthesis networks by rendering OLAT images with heavily blurred environment maps in the Polyhaven and Laval Outdoor HDR datasets. 

\subsection{Implementation Details}
We implement all components of COMPOSE using PyTorch~\cite{Pytorch}. 
When training our lighting estimation VAE, we use a batch size of $128$ and a learning rate of $10^{-4}$ on $4$ $24$GB A$10$G GPUs. 
For our light diffusion network, we train with a batch size of $88$ and a learning rate of $10^{-4}$ using $8$ A$100$ GPUs. 
For our shadow synthesis U-Net, we train with a batch size of $100$ and a learning rate of $10^{-4}$ using $4$ $24$GB A$10$G GPUs. 
For all networks, we train using the Adam Optimizer~\cite{AdamOptimizer}. 
The image resolution used to train the lighting estimation and shadow synthesis networks is $256\times256$ and the resolution used to train the light diffusion network is $768\times768$. Please see the supplementary materials for architectural details of the lighting estimation network, the lighting diffusion network, and the shadow synthesis network.

\section{Experiments}


\begin{figure*}[t]
\begin{center}
   \begin{minipage}[t]{0.1601\linewidth}
\centering
\includegraphics[width=\linewidth]{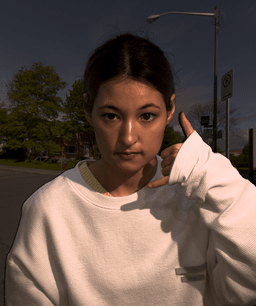} \\
\end{minipage}
\begin{minipage}[t]{0.1601\linewidth}
\centering
\includegraphics[width=\linewidth]{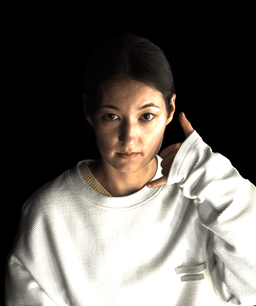} \\
\end{minipage}
\begin{minipage}[t]{0.1601\linewidth}
\centering
\includegraphics[width=\linewidth]{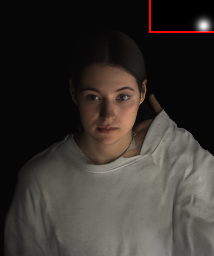} \\
\end{minipage}
\begin{minipage}[t]{0.1601\linewidth}
\centering
\includegraphics[width=\linewidth]{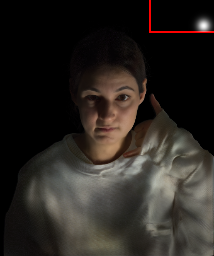} \\
\end{minipage}
\begin{minipage}[t]{0.1601\linewidth}
\centering
\includegraphics[width=\linewidth]{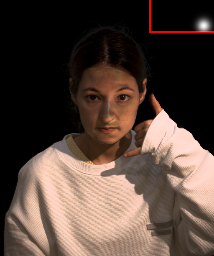} \\
\end{minipage}

\begin{minipage}[t]{0.1601\linewidth}
\centering
\includegraphics[width=\linewidth]{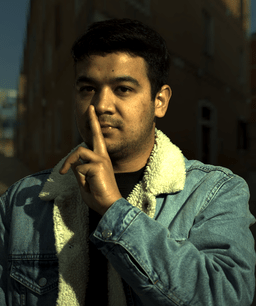} \\
\end{minipage}
\begin{minipage}[t]{0.1601\linewidth}
\centering
\includegraphics[width=\linewidth]{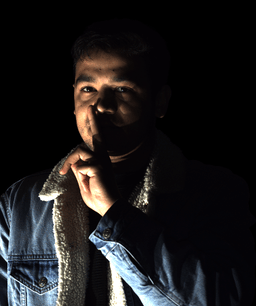} \\
\end{minipage}
\begin{minipage}[t]{0.1601\linewidth}
\centering
\includegraphics[width=\linewidth]{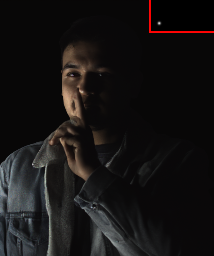} \\
\end{minipage}
\begin{minipage}[t]{0.1601\linewidth}
\centering
\includegraphics[width=\linewidth]{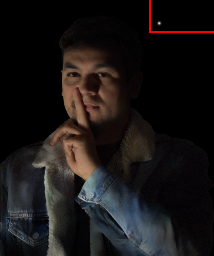} \\
\end{minipage}
\begin{minipage}[t]{0.1601\linewidth}
\centering
\includegraphics[width=\linewidth]{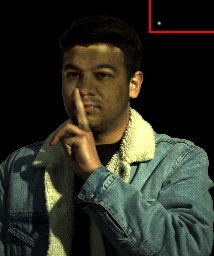} \\
\end{minipage}

\begin{minipage}[t]{0.1601\linewidth}
\centering
\includegraphics[width=\linewidth]{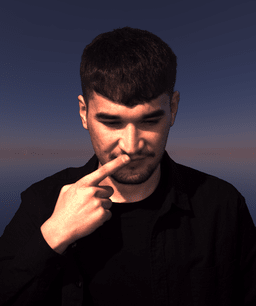} \\
\end{minipage}
\begin{minipage}[t]{0.1601\linewidth}
\centering
\includegraphics[width=\linewidth]{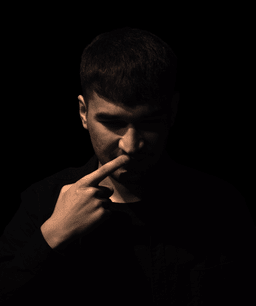} \\
\end{minipage}
\begin{minipage}[t]{0.1601\linewidth}
\centering
\includegraphics[width=\linewidth]{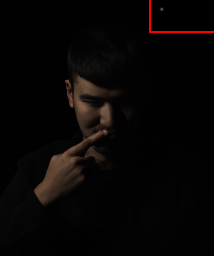} \\
\end{minipage}
\begin{minipage}[t]{0.1601\linewidth}
\centering
\includegraphics[width=\linewidth]{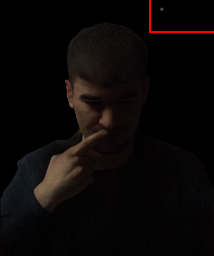} \\
\end{minipage}
\begin{minipage}[t]{0.1601\linewidth}
\centering
\includegraphics[width=\linewidth]{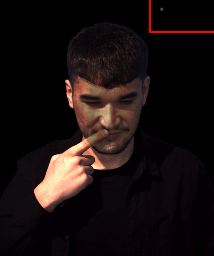} \\
\end{minipage}

\begin{minipage}[t]{0.1601\linewidth}
\centering
\includegraphics[width=\linewidth]{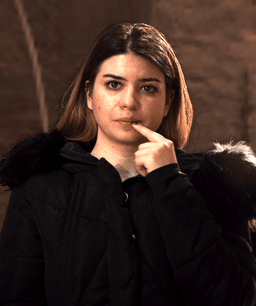} \\
\small a) Source 
\end{minipage}
\begin{minipage}[t]{0.1601\linewidth}
\centering
\includegraphics[width=\linewidth]{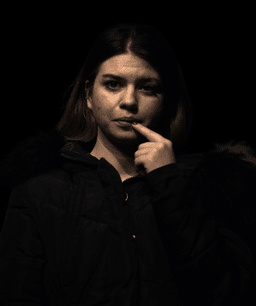} \\
\small b) Target 
\end{minipage}
\begin{minipage}[t]{0.1601\linewidth}
\centering
\includegraphics[width=\linewidth]{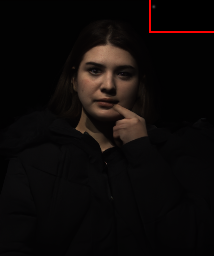} \\
\small c) Ours
\end{minipage}
\begin{minipage}[t]{0.1601\linewidth}
\centering
\includegraphics[width=\linewidth]{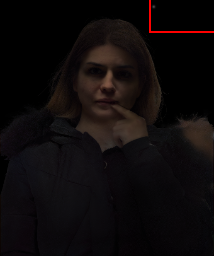} \\
\small d) TR~\cite{TotalRelighting}
\end{minipage}
\begin{minipage}[t]{0.1601\linewidth}
\centering
\includegraphics[width=\linewidth]{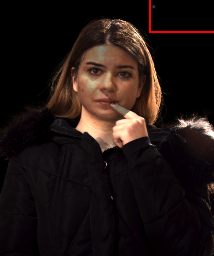} \\
\small e)~\cite{face-relighting-with-geometrically-consistent-shadows}
\end{minipage}

\vspace{-2mm}
\caption{\small \textbf{Shadow Synthesis}. Our model synthesizes more plausible shadows than the baselines, and, unlike~\cite{face-relighting-with-geometrically-consistent-shadows}, is also able to properly remove shadows from the source image before relighting. Environment maps are shown to help visualize each test lighting. }
\label{fig:shadowsynthesis}
\end{center} \vspace{-5mm}
\end{figure*}

\subsection{Datasets}
To train our lighting estimation and shadow synthesis networks, we generate our training sets using light stage images and $521$ outdoor environment maps. Our training set consists of $82$ subjects with significant variation in factors such as skin tones, pose, expressions, and accessories.

For evaluation, we use two separate evaluation sets: the first is generated using our light stage images and has corresponding groundtruth, and the second consists of in-the-wild images that represent more realistic application settings and do not have groundtruth. Our light stage evaluation set consists of $48$ images from a total of $16$ subjects with diverse skin tones, body poses, expressions, and accessories. All evaluation subjects and lightings are held out and unseen during training. Our in-the-wild evaluation set also consists of unseen test subjects and applies unseen lighting conditions in all qualitative results.  

\begin{table}[t!]

\caption{\small
\textbf{Shadow Editing Performance}. We evaluate shadow editing using unseen lightings with varied intensity and size on $16$ held out highly diverse light stage test subjects with groundtruth relit images. COMPOSE outperforms all baselines quantitatively in shadow editing performance across all metrics (mean $\pm$ standard deviation). 
}\label{tab:ShadowEditingPerformance}
\vspace{-5mm}
\begin{center}
\scalebox{0.75}{
\setlength\tabcolsep{8pt}  
\begin{tabular}{c c c c c}
\hline
 & MAE & MSE & SSIM & LPIPS \\
\hline
Hou \textit{et al.}~\cite{face-relighting-with-geometrically-consistent-shadows} & $0.1172\pm0.0754$ & $0.0438\pm0.0503$ & $0.7059\pm0.1061$ & $0.2371\pm0.0871$\\
\hline
Total Relighting~\cite{TotalRelighting} & $0.1008\pm0.0743$ & $0.0379\pm0.0447$ & $0.7754\pm0.0738$ & $0.2120\pm0.0572$ \\
\hline
COMPOSE (Ours) & $\mathbf{0.0965\pm0.0686}$ & $\mathbf{0.0349\pm0.0368}$ & $\mathbf{0.7780\pm0.0826}$ & $\mathbf{0.1973\pm0.0678}$\\
\hline
\end{tabular}}
\end{center}

\vspace{-8mm}
\end{table}


\subsection{Shadow Editing Evaluation}
We evaluate our shadow editing performance (synthesis and removal) by using randomly sampled, unseen lighting positions as the target lights to generate relit light stage images of our $16$ test subjects, which serve as groundtruth. 

\begin{figure}[t]
\begin{center}

\begin{minipage}[t]{0.1601\linewidth}
\centering
\includegraphics[width=\linewidth]{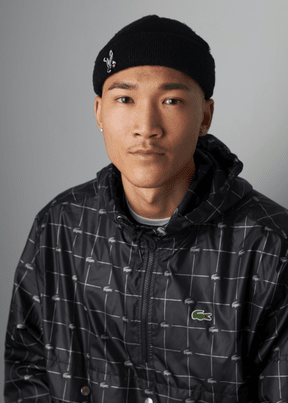} \\
\end{minipage}
\begin{minipage}[t]{0.1601\linewidth}
\centering
\includegraphics[width=\linewidth]{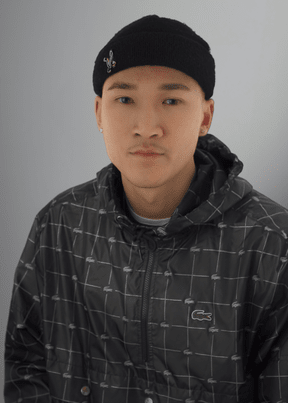} \\
\end{minipage}
\begin{minipage}[t]{0.1601\linewidth}
\centering
\includegraphics[width=\linewidth]{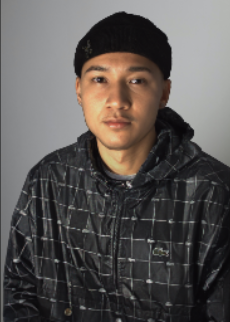} \\
\end{minipage}
\begin{minipage}[t]{0.1601\linewidth}
\centering
\includegraphics[width=\linewidth]{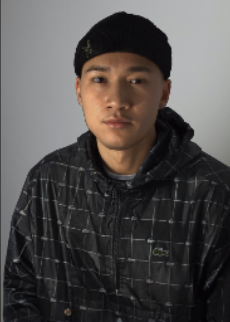} \\
\end{minipage}
\begin{minipage}[t]{0.1601\linewidth}
\centering
\includegraphics[width=\linewidth]{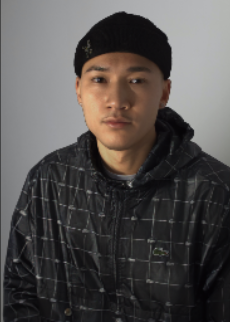} \\
\end{minipage}

\begin{minipage}[t]{0.1601\linewidth}
\centering
\includegraphics[width=\linewidth]{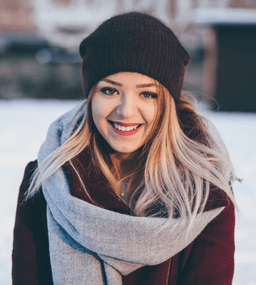} \\
\end{minipage}
\begin{minipage}[t]{0.1601\linewidth}
\centering
\includegraphics[width=\linewidth]{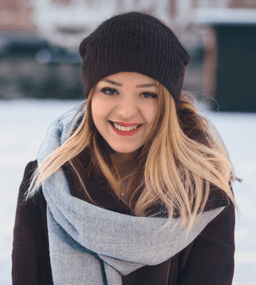} \\
\end{minipage}
\begin{minipage}[t]{0.1601\linewidth}
\centering
\includegraphics[width=\linewidth]{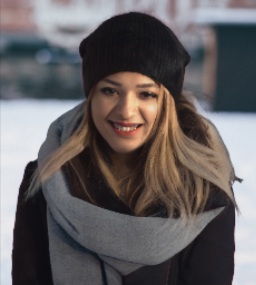} \\
\end{minipage}
\begin{minipage}[t]{0.1601\linewidth}
\centering
\includegraphics[width=\linewidth]{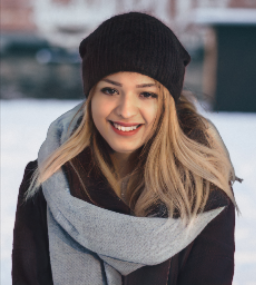} \\
\end{minipage}
\begin{minipage}[t]{0.1601\linewidth}
\centering
\includegraphics[width=\linewidth]{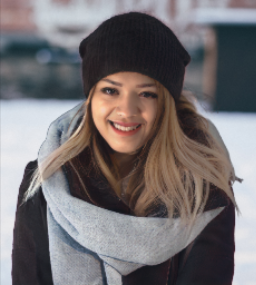} \\
\end{minipage}

\begin{minipage}[t]{0.1601\linewidth}
\centering
\includegraphics[width=\linewidth]{Figures/Shadow_Editing/albert-dera-ILip77SbmOE-unsplash0_input_hdr_smaller.png} \\
\end{minipage}
\begin{minipage}[t]{0.1601\linewidth}
\centering
\includegraphics[width=\linewidth]{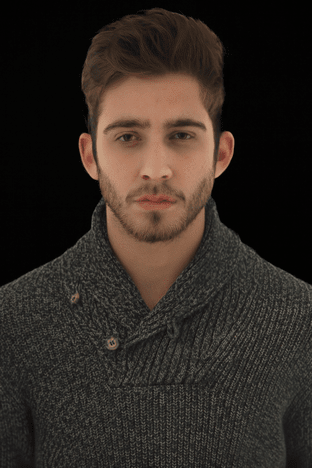} \\
\end{minipage}
\begin{minipage}[t]{0.1601\linewidth}
\centering
\includegraphics[width=\linewidth]{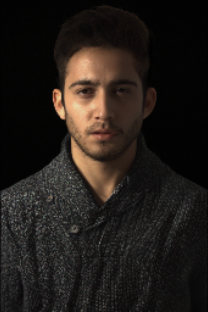} \\
\end{minipage}
\begin{minipage}[t]{0.1601\linewidth}
\centering
\includegraphics[width=\linewidth]{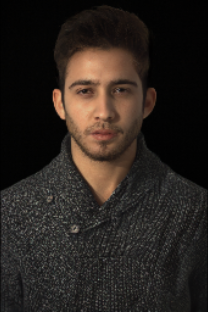} \\
\end{minipage}
\begin{minipage}[t]{0.1601\linewidth}
\centering
\includegraphics[width=\linewidth]{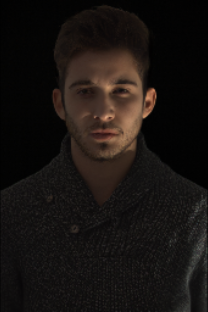} \\
\end{minipage}

\begin{minipage}[t]{0.1601\linewidth}
\centering
\includegraphics[width=\linewidth]{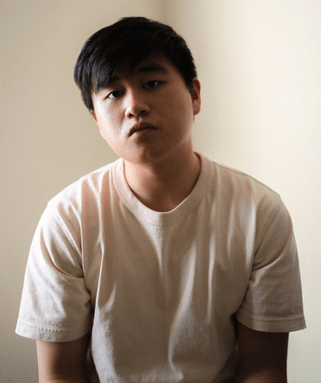} \\
\small a) Source Image
\end{minipage}
\begin{minipage}[t]{0.1601\linewidth}
\centering
\includegraphics[width=\linewidth]{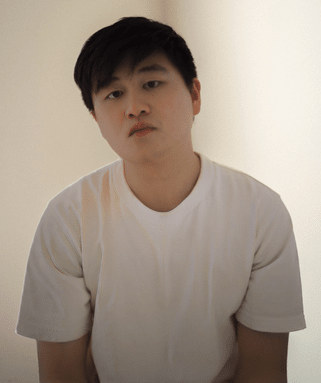} \\
\small b) Soften Shadows
\end{minipage}
\begin{minipage}[t]{0.1601\linewidth}
\centering
\includegraphics[width=\linewidth]{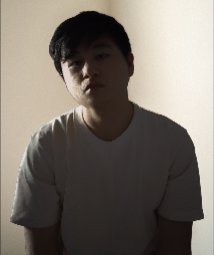} \\
\small c) Intensify Shadows
\end{minipage}
\begin{minipage}[t]{0.1601\linewidth}
\centering
\includegraphics[width=\linewidth]{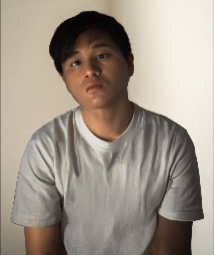} \\
\small d) Modify Light Size
\end{minipage}
\begin{minipage}[t]{0.1601\linewidth}
\centering
\includegraphics[width=\linewidth]{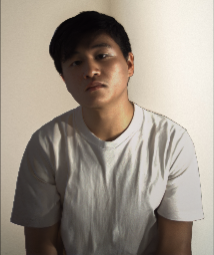} \\
\small e) Rotate Shadows
\end{minipage}


\vspace{-2mm}
\caption{\small \textbf{Shadow Editing}. Our method achieves complete shadow editing control, including softening/intensifying shadows, changing light size to alter shape and intensity, and changing light position, all while preserving the source image's ambient light. 
}\label{fig:ShadowEditing}
\end{center}\vspace{-8mm}
\end{figure}

Each input image is rendered with a randomly selected environment map out of $125$ unseen outdoor environment maps and is first passed to our deshading network to generate diffuse image $\mathbf{I}_{D}$. $\mathbf{I}_{D}$ and target environment map $\mathbf{E}_{T}$ are then fed to our two-stage shadow synthesis pipeline to generate the newly shadowed image $\mathbf{I}_{S}$, which is evaluated against the groundtruth light stage image. 

Tab.~\ref{tab:ShadowEditingPerformance} compares our shadow editing performance against prior relighting methods. For both baselines~\cite{TotalRelighting, face-relighting-with-geometrically-consistent-shadows}, the test results were provided by the authors. 
Our method achieves state-of-the-art results on all metrics (MAE, MSE, SSIM~\cite{SSIM}, and LPIPS~\cite{LPIPS}). As seen in Fig.~\ref{fig:shadowsynthesis}, our model can synthesize appropriate shadows for various light positions. The method of Hou \textit{et al.}~\cite{face-relighting-with-geometrically-consistent-shadows} cannot remove existing shadows in the source image and the shadow traces carry over as artifacts to the relit images. 
In addition, Hou \textit{et al.}~\cite{face-relighting-with-geometrically-consistent-shadows} only models the lighting direction and does not model the light size as a parameter, which leads to inaccurate shadow shape when the light size is varied. Total Relighting~\cite{TotalRelighting} is often unable to synthesize physically plausible shadows, and will sometimes overshadow (Fig.~\ref{fig:shadowsynthesis}, rows $1$, $3$, and $4$) or undershadow the image (Fig.~\ref{fig:shadowsynthesis}, row $2$). Moreover, their shadows are often blurry and not as sharp as our method. Compared to the baselines, COMPOSE can properly remove existing shadows from the source image and synthesize geometrically plausible and realistic shadows for a wide variety of lighting conditions. 

\vspace{-3mm}


\begin{figure*}[t]
\begin{center}

\begin{minipage}[t]{0.1601\linewidth}
\centering
\includegraphics[width=\linewidth]{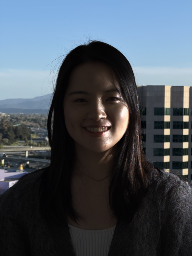} \\
\end{minipage}
\begin{minipage}[t]{0.1601\linewidth}
\centering
\includegraphics[width=\linewidth]{Figures/Shadow_Editing_supp/img00003.png} \\
\end{minipage}
\begin{minipage}[t]{0.1601\linewidth}
\centering
\includegraphics[width=\linewidth]{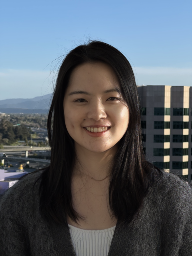} \\
\end{minipage}
\begin{minipage}[t]{0.1601\linewidth}
\centering
\includegraphics[width=\linewidth]{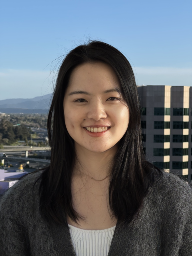} \\
\end{minipage}
\begin{minipage}[t]{0.1601\linewidth}
\centering
\includegraphics[width=\linewidth]{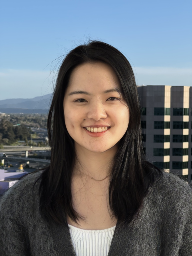} \\
\end{minipage}

\begin{minipage}[t]{0.1601\linewidth}
\centering
\includegraphics[width=\linewidth]{Figures/Shadow_Editing_supp/NXT_img00003_shape.png} \\
\end{minipage}
\begin{minipage}[t]{0.1601\linewidth}
\centering
\includegraphics[width=\linewidth]{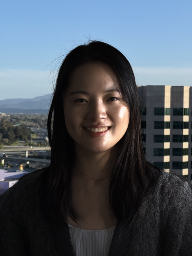} \\
\end{minipage}
\begin{minipage}[t]{0.1601\linewidth}
\centering
\includegraphics[width=\linewidth]{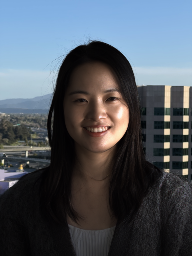} \\
\end{minipage}
\begin{minipage}[t]{0.1601\linewidth}
\centering
\includegraphics[width=\linewidth]{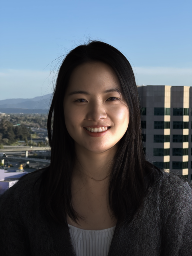} \\
\end{minipage}
\begin{minipage}[t]{0.1601\linewidth}
\centering
\includegraphics[width=\linewidth]{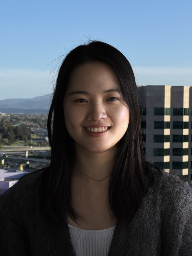} \\
\end{minipage}

\begin{minipage}[t]{0.1601\linewidth}
\centering
\includegraphics[width=\linewidth]{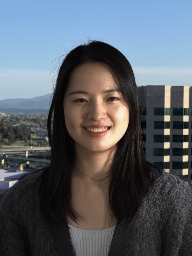} \\
\end{minipage}
\begin{minipage}[t]{0.1601\linewidth}
\centering
\includegraphics[width=\linewidth]{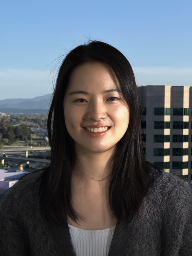} \\
\end{minipage}
\begin{minipage}[t]{0.1601\linewidth}
\centering
\includegraphics[width=\linewidth]{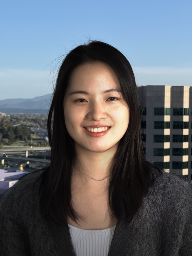} \\
\end{minipage}
\begin{minipage}[t]{0.1601\linewidth}
\centering
\includegraphics[width=\linewidth]{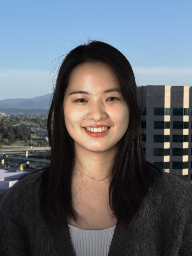} \\
\end{minipage}
\begin{minipage}[t]{0.1601\linewidth}
\centering
\includegraphics[width=\linewidth]{Figures/Shadow_Editing_supp/NXT_img00027.png} \\
\end{minipage}

\vspace{-2mm}
\caption{\small \textbf{Degree of Shadow Editing}. We demonstrate COMPOSE's controllable shadow editing by varying shadow intensity (top row), light diffusion for modifying shadow shape and intensity (middle row), and light position (bottom row). 
}\label{fig:ShadowEditingDegree}
\end{center}\vspace{-8mm}
\end{figure*}

\subsection{Portrait Shadow Editing Applications}
We demonstrate the full performance of COMPOSE by performing all forms of shadow editing, including shadow softening, shadow intensifying, modifying light size (which changes both shadow shape and intensity), and shadow rotation by changing the dominant light position. For each image, we first estimate the diffuse image $\mathbf{I}_{D}$ and the shadowed image $\mathbf{I}_{S}$ and perform image compositing between these two images to generate the final edited image $\mathbf{I}_{E}$. As seen in Fig.~\ref{fig:ShadowEditing}, COMPOSE properly edits the shadows of in-the-wild images. To soften or intensify the shadow without changing the shadow shape, we rely on our image compositing coefficients $\omega_{d}$ and $\omega_{s}$, which control the contributions of $\mathbf{I}_{D}$ and $\mathbf{I}_{S}$ respectively. We increase the weight of $\omega_{d}$ to soften the shadow and increase the weight of $\omega_{s}$ to strengthen the shadow. We further show that we can modify light size by tuning the Gaussian spread $\sigma$, which changes both shadow shape and intensity. Enlarging the light makes the shadow more diffuse and less intense, whereas shrinking the light generates a harder, more intense shadow. For the subjects in rows $2$, $3$, and $4$ we enlarged the light, resulting in a more diffuse shadow especially noticeable around the nose. For the subject in row $1$ we instead shrunk the light and appropriately produce an edited image with more prominent and intense hard shadows. Finally, we also demonstrate shadow rotation by changing the dominant light position in $\mathbf{I}_{S}$'s environment map. 

To demonstrate COMPOSE can properly control shadow editing, we visualize various degrees of shadow softening/intensifying, light diffusion, and shadow rotation in Fig.~\ref{fig:ShadowEditingDegree}. The first row shows varying shadow intensity, where the shadows become less intense from left to right using $\omega_{d}$ and $\omega_{s}$. Our results demonstrate that these flexible weights allow for a wide range of shadow intensities. The second row shows increasing degrees of light diffusion from left to right, which corresponds to increasing the light size. The shadows in the generated images correctly transition from harder shadows to more diffuse shadows as the light size increases. The third row shows shadow rotation, where the light is rotated horizontally along the environment map and around the subject. COMPOSE is able to produce physically plausible shadows as the light position changes. 

\begin{figure}[t]
\begin{center}

\begin{minipage}[t]{0.1601\linewidth}
\centering
\includegraphics[width=\linewidth]{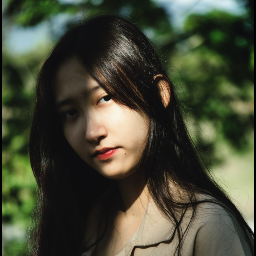} \\
\end{minipage}
\begin{minipage}[t]{0.1601\linewidth}
\centering
\includegraphics[width=\linewidth]{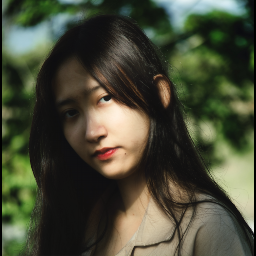} \\
\end{minipage}
\begin{minipage}[t]{0.1601\linewidth}
\centering
\includegraphics[width=\linewidth]{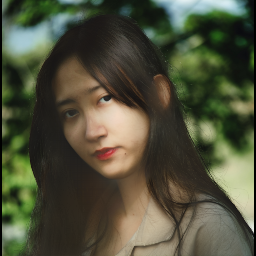} \\
\end{minipage}
\begin{minipage}[t]{0.1601\linewidth}
\centering
\includegraphics[width=\linewidth]{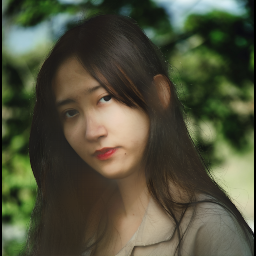} \\
\end{minipage}
\begin{minipage}[t]{0.1601\linewidth}
\centering
\includegraphics[width=\linewidth]{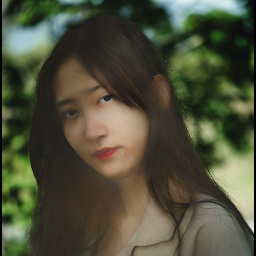} \\
\end{minipage}

\begin{minipage}[t]{0.1601\linewidth}
\centering
\includegraphics[width=\linewidth]{Figures/Shadow_Softening/00000098_source_256.png} \\
\end{minipage}
\begin{minipage}[t]{0.1601\linewidth}
\centering
\includegraphics[width=\linewidth]{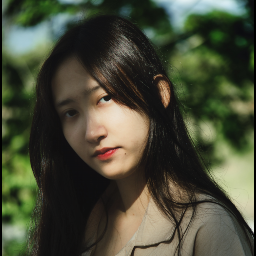} \\
\end{minipage}
\begin{minipage}[t]{0.1601\linewidth}
\centering
\includegraphics[width=\linewidth]{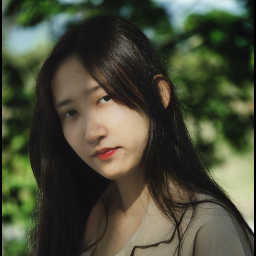} \\
\end{minipage}
\begin{minipage}[t]{0.1601\linewidth}
\centering
\includegraphics[width=\linewidth]{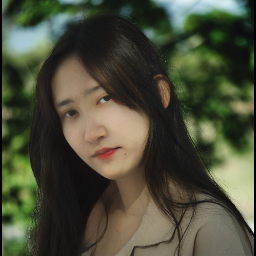} \\
\end{minipage}
\begin{minipage}[t]{0.1601\linewidth}
\centering
\includegraphics[width=\linewidth]{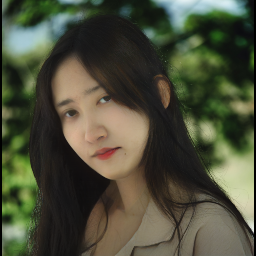} \\
\end{minipage}

\begin{minipage}[t]{0.1601\linewidth}
\centering
\includegraphics[width=\linewidth]{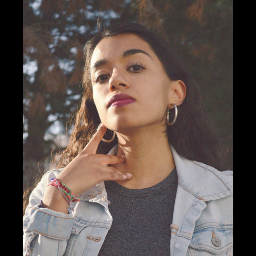} \\
\end{minipage}
\begin{minipage}[t]{0.1601\linewidth}
\centering
\includegraphics[width=\linewidth]{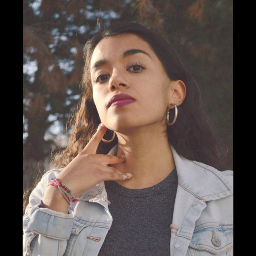} \\
\end{minipage}
\begin{minipage}[t]{0.1601\linewidth}
\centering
\includegraphics[width=\linewidth]{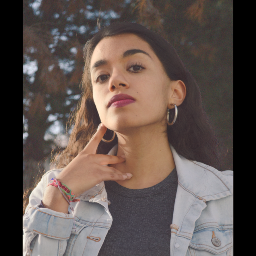} \\
\end{minipage}
\begin{minipage}[t]{0.1601\linewidth}
\centering
\includegraphics[width=\linewidth]{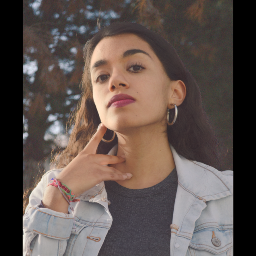} \\
\end{minipage}
\begin{minipage}[t]{0.1601\linewidth}
\centering
\includegraphics[width=\linewidth]{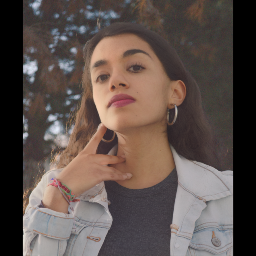} \\
\end{minipage}

\begin{minipage}[t]{0.1601\linewidth}
\centering
\includegraphics[width=\linewidth]{Figures/Shadow_Softening/00000105_source_256.png} \\
\small a) Source  
\end{minipage}
\begin{minipage}[t]{0.1601\linewidth}
\centering
\includegraphics[width=\linewidth]{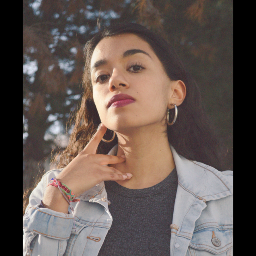} \\
\small b) $0.75$
\end{minipage}
\begin{minipage}[t]{0.1601\linewidth}
\centering
\includegraphics[width=\linewidth]{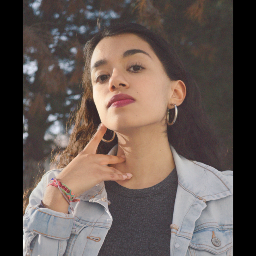} \\
\small c) $0.5$
\end{minipage}
\begin{minipage}[t]{0.1601\linewidth}
\centering
\includegraphics[width=\linewidth]{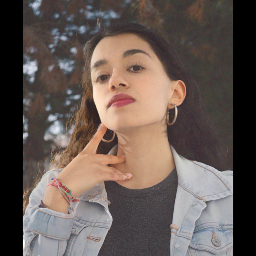} \\
\small d) $0.2$
\end{minipage}
\begin{minipage}[t]{0.1601\linewidth}
\centering
\includegraphics[width=\linewidth]{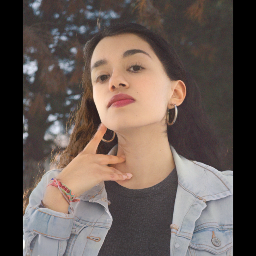} \\
\small e) $0.0$ 
\end{minipage}
\vspace{-3mm}
\caption{\small \textbf{Controllable Shadow Softening}. For each subject, we show the shadow softening results of~\cite{controllable-light-diffusion} in the first row and COMPOSE in the second. COMPOSE leaves less shadow traces at level $0.0$, where the goal is to remove the shadows completely. 
}\label{fig:ShadowSoftening}
\end{center}\vspace{-7mm}
\end{figure}

It is important to note that this level of controllable shadow editing is not achievable by existing portrait shadow editing methods~\cite{controllable-light-diffusion, PortraitShadowManipulation} that are only able to handle shadow softening or complete shadow removal as well as existing portrait relighting methods that completely modify the existing environment and cannot preserve the original scene illumination~\cite{TotalRelighting}, do not properly model all shadow attributes (\textit{e.g.} intensity, shape, and position)~\cite{face-relighting-with-geometrically-consistent-shadows, diffusionfacerelighting} or are not designed to handle lighting conditions with more substantial shadows~\cite{UCSDSingleImagePortraitRelighting, DPR, SfSNet}. 
\vspace{-3mm}
\subsection{Shadow Softening Comparison} We compare on in-the-wild shadow softening with the state-of-the-art portrait shadow softening method Futschik \textit{et al.}~\cite{controllable-light-diffusion}. Using results provided by the authors, we visualize different degrees of shadow softening for~\cite{controllable-light-diffusion} by varying the light diffusion parameter ($0.75$, $0.5$, $0.2$, and $0.0$) and compare with equivalent degrees of shadow softening by COMPOSE, as shown in Fig.~\ref{fig:ShadowSoftening}. Higher parameter values correspond to less softening whereas lower values correspond to more softening. At $0.0$, the goal is to completely remove the source image's shadows. Across all degrees of light diffusion, both methods gradually soften the shadow, but COMPOSE leaves less shadow traces than~\cite{controllable-light-diffusion} when completely removing the source image's shadows at level $0.0$, especially for images with darker shadows. This is thanks to our hierarchical transformer design for our light diffusion network, which better handles removing the effects of shadows at all scales. 

\section{Conclusion} We have proposed COMPOSE: the first single image portrait shadow editing method that can perform comprehensive variations of shadow editing including editing shadow intensity, light size, and shadow position while preserving the source image's lighting environment. This is largely enabled by our novel lighting decomposition into ambient light and an editable gaussian dominant light, which enables new applications like shrinking the light and intensifying shadows as well as disentangled shadow editing that preserves the scene ambience. We have demonstrated the novel photo editing applications enabled by our method over prior work qualitatively and that our method achieves state-of-the-art shadow editing performance quantitatively over prior relighting methods. We hope that our work can inspire and motivate more research in the exciting new research area of controllable and comprehensive portrait shadow editing. 
\vspace{-5mm}
\subsubsection{\textbf{Limitations.}} Good directions for future work include handling lighting environments with multiple light sources and indoor environments, as our framework is intended mostly for outdoor settings, where the sun is the single dominant light. 
In addition, while $\mathbf{I}_{D}$ can preserve the ambience, we still notice there are sometimes color shifts with the environmental lighting during shadow editing, largely due to the compositing step with shadowed image $\mathbf{I}_{S}$ (which adopts the light stage's environmental lighting). 
This could be alleviated by modeling the diffuse lighting tint similar to~\cite{controllable-light-diffusion} or by training with white balanced data to remove the color bias of the light stage. 
We also believe that one can continue to improve the generalizability of our model. Similar to other relighting methods that rely on light stage data~\cite{controllable-light-diffusion, TotalRelighting, UCSDSingleImagePortraitRelighting}, we train with a limited number of subjects. 
Future work may incorporate synthetic data to expand the training set or adopt latent diffusion models, which naturally have high generalizability~\cite{stablediffusion}. 
Finally, one can further improve the identity preservation of our method. 
We find that high frequency details are sometimes lost at the initial U-Net stage of shadow synthesis, and ways to better preserve the identity include passing $\mathbf{I}_{D}$ or the input image $\mathbf{I}_{N}$, which contain high frequency details, as additional conditions to the DDPM or to use stronger backbones than~\cite{UCSDSingleImagePortraitRelighting}.  
\vspace{-5mm}
\subsubsection{\textbf{Potential Societal Impacts.}} In image editing, there are always concerns of generating malicious deepfakes. However, we only alter illumination, not subject identity. Even so, there may be concerns that altering the illumination and especially adding dark shadows to faces could be used as a malicious attack to worsen the performance of downstream tasks such as face recognition. Users can detect these attacks using existing deepfake detection methods~\cite{malp, reverse-engineering-of-generative-models-inferring-model-hyperparameters-from-generated-images, hierarchical-fine-grained-image-forgery-detection-and-localization, interpretabledeepfake, intrinsicdeepfake, icasspdeepfake, revisitingdeepfake, universaldeepfake} or by training one using the edited images generated by COMPOSE. State-of-the-art shadow removal methods~\cite{PortraitShadowManipulation, blind-removal-of-facial-foreign-shadows} are also an option to counter these potential attacks since they can remove any facial shadows that COMPOSE synthesizes.

%
%
\bibliographystyle{splncs04}
\bibliography{main}

\begin{thebibliography}{10}
\providecommand{\url}[1]{\texttt{#1}}
\providecommand{\urlprefix}{URL }
\providecommand{\doi}[1]{https://doi.org/#1}

\bibitem{interpretabledeepfake}
Aghasanli, A., Kangin, D., Angelov, P.: Interpretable-through-prototypes deepfake detection for diffusion models. In: ICCVW (2023)

\bibitem{malp}
Asnani, V., Yin, X., Hassner, T., Liu, X.: {MaLP}: Manipulation localization using a proactive scheme. In: CVPR (2023)

\bibitem{reverse-engineering-of-generative-models-inferring-model-hyperparameters-from-generated-images}
Asnani, V., Yin, X., Hassner, T., Liu, X.: Reverse engineering of generative models: Inferring model hyperparameters from generated images. PAMI  (2023)

\bibitem{bi2021avatar}
Bi, S., Lombardi, S., Saito, S., Simon, T., Wei, S.E., McPhail, K., Ramamoorthi, R., Sheikh, Y., Saragih, J.: Deep relightable appearance models for animatable faces. TOG  (2021)

\bibitem{SIRA}
Caselles, P., Ramon, E., Garcia, J., i~Nieto, X.G., Moreno-Noguer, F., Triginer, G.: {SIRA}: Relightable avatars from a single image. In: WACV (2023)

\bibitem{temporallyconsistentvideorelighting}
Chandran, S., Hold-Geoffroy, Y., Sunkavalli, K., Shu, Z., Jayasuriya, S.: Temporally consistent relighting for portrait videos. In: WACV (2022)

\bibitem{ilvr_adm}
Choi, J., Kim, S., Jeong, Y., Gwon, Y., Yoon, S.: {ILVR}: Conditioning method for denoising diffusion probabilistic models. In: ICCV (2021)

\bibitem{icasspdeepfake}
Corvi, R., Cozzolino, D., Zingarini, G., Poggi, G., Nagano, K., Verdoliva, L.: On the detection of synthetic images generated by diffusion models. In: ICASSP (2023)

\bibitem{paullightstage}
Debevec, P., Hawkins, T., Tchou, C., Duiker, H.P., Sarokin, W., Sagar, M.: Acquiring the reflectance field of a human face. In: SIGGRAPH (2000)

\bibitem{ding2023diffusionrig}
Ding, Z., Zhang, C., Xia, Z., Jebe, L., Tu, Z., Zhang, X.: {DiffusionRig}: Learning personalized priors for facial appearance editing. In: CVPR (2023)

\bibitem{controllable-light-diffusion}
Futschik, D., Ritland, K., Vecore, J., Fanello, S., Orts-Escolano, S., Curless, B., Sýkora, D., Pandey, R.: Controllable light diffusion for portraits. In: CVPR (2023)

\bibitem{therelightables}
Guo, K., Lincoln, P., Davidson, P., Busch, J., Yu, X., Whalen, M., Harvey, G., Orts-Escolano, S., Pandey, R., Dourgarian, J., DuVall, M., Tang, D., Tkach, A., Kowdle, A., Cooper, E., Dou, M., Fanello, S., Fyffe, G., Rhemann, C., Taylor, J., Debevec, P., Izadi, S.: The {Relightables}: Volumetric performance capture of humans with realistic relighting. In: SIGGRAPH Asia (2019)

\bibitem{hierarchical-fine-grained-image-forgery-detection-and-localization}
Guo, X., Liu, X., Ren, Z., Grosz, S., Masi, I., Liu, X.: Hierarchical fine-grained image forgery detection and localization. In: CVPR (2023)

\bibitem{he21unsupervised}
He, Y., Xing, Y., Zhang, T., Chen, Q.: Unsupervised portrait shadow removal via generative priors. In: MM (2021)

\bibitem{DDPM}
Ho, J., Jain, A., Abbeel, P.: Denoising diffusion probabilistic models. In: NeurIPS (2020)

\bibitem{LavalOutdoor}
Hold-Geoffroy, Y., Athawale, A., Lalonde, J.F.: Deep sky modeling for single image outdoor lighting estimation. In: CVPR (2019)

\bibitem{face-relighting-with-geometrically-consistent-shadows}
Hou, A., Sarkis, M., Bi, N., Tong, Y., Liu, X.: Face relighting with geometrically consistent shadows. In: CVPR (2022)

\bibitem{towards-high-fidelity-face-relighting-with-realistic-shadows}
Hou, A., Zhang, Z., Sarkis, M., Bi, N., Tong, Y., Liu, X.: Towards high fidelity face relighting with realistic shadows. In: CVPR (2021)

\bibitem{RANA}
Iqbal, U., Caliskan, A., Nagano, K., Khamis, S., Molchanov, P., Kautz, J.: {RANA}: Relightable articulated neural avatars. In: ICCV (2023)

\bibitem{geometryawarehumanrelighting}
Ji, C., Yu, T., Guo, K., Liu, J., Liu, Y.: Geometry-aware single-image full-body human relighting. In: ECCV (2022)

\bibitem{revisitingdeepfake}
Kamat, S., Agarwal, S., Darrell, T., Rohrbach, A.: Revisiting generalizability in deepfake detection: Improving metrics and stabilizing transfer. In: ICCVW (2023)

\bibitem{AdamOptimizer}
Kingma, D., Adam, J.B.: A method for stochastic optimization. In: ICLR (2014)

\bibitem{VAE}
Kingma, D.P., Welling, M.: Auto-encoding variational bayes . In: ICLR (2014)

\bibitem{Lagunas2021humanrelighting}
Lagunas, M., Sun, X., Yang, J., Villegas, R., Zhang, J., Shu, Z., Masia, B., Gutierrez, D.: Single-image full-body human relighting. In: EGSR (2021)

\bibitem{HaLeWACV19}
Le, H., Kakadiaris, I.: Illumination-invariant face recognition with deep relit face images. In: WACV (2019)

\bibitem{blind-removal-of-facial-foreign-shadows}
Liu, Y., Hou, A., Huang, X., Ren, L., Liu, X.: Blind removal of facial foreign shadows. In: BMVC (2022)

\bibitem{intrinsicdeepfake}
Lorenz, P., Durall, R.L., Keuper, J.: Detecting images generated by deep diffusion models using their local intrinsic dimensionality. In: ICCVW (2023)

\bibitem{lightpainter}
Mei, Y., Zhang, H., Zhang, X., Zhang, J., Shu, Z., Wang, Y., Wei, Z., Yan, S., Jung, H., Patel, V.M.: {LightPainter}: Interactive portrait relighting with freehand scribble. In: CVPR (2023)

\bibitem{PhysicsGuidedRelighting}
Nestmeyer, T., Lalonde, J.F., Matthews, I., Lehrmann, A.: Learning physics-guided face relighting under directional light. In: CVPR (2020)

\bibitem{universaldeepfake}
Ojha, U., Li, Y., Lee, Y.J.: Towards universal fake image detectors that generalize across generative models. In: CVPR (2023)

\bibitem{TotalRelighting}
Pandey, R., Orts-Escolano, S., LeGendre, C., Haene, C., Bouaziz, S., Rhemann, C., Debevec, P., Fanello, S.: Total relighting: Learning to relight portraits for background replacement. In: SIGGRAPH (2021)

\bibitem{Pytorch}
Paszke, A., Gross, S., Chintala, S., Chanan, G., Yang, E., DeVito, Z., Lin, Z., Desmaison, A., Antiga, L., Lerer, A.: Automatic differentiation in {P}y{T}orch. In: NeurIPSW (2017)

\bibitem{diffusionfacerelighting}
Ponglertnapakorn, P., Tritrong, N., Suwajanakorn, S.: {DiFaReli}: Diffusion face relighting. In: ICCV (2023)

\bibitem{FaceLit}
Ranjan, A., Yi, K.M., Chang, J.H.R., Tuzel, O.: {FaceLit}: Neural {3D} relightable faces. In: CVPR (June 2023)

\bibitem{stablediffusion}
Rombach, R., Blattmann, A., Lorenz, D., Esser, P., Ommer, B.: High-resolution image synthesis with latent diffusion models. In: CVPR (2022)

\bibitem{lightstagedesk}
Sengupta, R., Curless, B., Kemelmacher-Shlizerman, I., Seitz, S.: A light stage on every desk. In: ICCV (2021)

\bibitem{SfSNet}
Sengupta, S., Kanazawa, A., Castillo, C.D., Jacobs, D.W.: Sf{S}{N}et: {L}earning shape, refectance and illuminance of faces in the wild. In: CVPR (2018)

\bibitem{Flickr}
Shih, Y., Paris, S., Barnes, C., Freeman, W.T., Durand, F.: Style transfer for headshot portraits. TOG  (2014)

\bibitem{MassTransport}
Shu, Z., Hadap, S., Shechtman, E., Sunkavalli, K., Paris, S., Samaras, D.: Portrait lighting transfer using a mass transport approach. TOG  (2017)

\bibitem{VGG}
Simonyan, K., Zisserman, A.: Very deep convolutional networks for large-scale image recognition. In: ICLR (2015)

\bibitem{UCSDSingleImagePortraitRelighting}
Sun, T., Barron, J.T., Tsai, Y.T., Xu, Z., Yu, X., Fyffe, G., Rhemann, C., Busch, J., Debevec, P., Ramamoorthi, R.: Single image portrait relighting. In: SIGGRAPH (2019)

\bibitem{sun2021nelf}
Sun, T., Lin, K.E., Bi, S., Xu, Z., Ramamoorthi, R.: {NeLF}: Neural light-transport field for portrait view synthesis and relighting. In: EGSR (2021)

\bibitem{volux-gan}
Tan, F., Fanello, S., Meka, A., OrtsEscolano, S., Tang, D., Pandey, R., Taylor, J., Tan, P., Zhang., Y.: {VoLux-GAN}: A generative model for {3D} face synthesis with {HDRI} relighting. In: SIGGRAPH (2022)

\bibitem{wang2023sunstage}
Wang, Y., Holynski, A., Zhang, X., Zhang, X.: {Sunstage}: Portrait reconstruction and relighting using the sun as a light stage. In: CVPR (2023)

\bibitem{SIPRExplicitMultiple}
Wang, Z., Yu, X., Lu, M., Wang, Q., Qian, C., Xu, F.: Single image portrait relighting via explicit multiple reflectance channel modeling. TOG  (2020)

\bibitem{SSIM}
Wang, Z., Bovik, A.C., Sheikh, H.R., Simoncelli, E.P.: Image quality assessment: from error visibility to structural similarity. TIP  (2004)

\bibitem{delighting}
Weir, J., Zhao, J., Chalmers, A., Rhee, T.: Deep portrait delighting. In: ECCV (2022)

\bibitem{BareSkinNet}
Yang, X., Taketomi, T.: {BareSkinNet}: Demakeup and de-lighting via {3D} face reconstruction. CGF  (2022)

\bibitem{yeh2022learning}
Yeh, Y.Y., Nagano, K., Khamis, S., Kautz, J., Liu, M.Y., Wang, T.C.: Learning to relight portrait images via a virtual light stage and synthetic-to-real adaptation. TOG  (2022)

\bibitem{neuralvideorelighting}
Zhang, L., Zhang, Q., Wu, M., Yu, J., Xu, L.: Neural video portrait relighting in real-time via consistency modeling. In: ICCV (2021)

\bibitem{LPIPS}
Zhang, R., Isola, P., Efros, A.A., Shechtman, E., Wang, O.: The unreasonable effectiveness of deep features as a perceptual metric. In: CVPR (2018)

\bibitem{NLT}
Zhang, X., Fanello, S., Tsai, Y.T., Sun, T., Xue, T., Pandey, R., Orts-Escolano, S., Davidson, P., Rhemann, C., Debevec, P., Barron, J.T., Ramamoorthi, R., Freeman, W.T.: Neural light transport for relighting and view synthesis. TOG  (2021)

\bibitem{PortraitShadowManipulation}
Zhang, X., Barron, J.T., Tsai, Y.T., Pandey, R., Zhang, X., Ng, R., Jacobs, D.E.: Portrait {S}hadow {M}anipulation. TOG  (2020)

\bibitem{DPR}
Zhou, H., Hadap, S., Sunkavalli, K., Jacobs, D.W.: Deep single portrait image relighting. In: ICCV (2019)

\end{thebibliography}

\clearpage
\setcounter{equation}{0}
\setcounter{figure}{0}
\setcounter{table}{0}
\setcounter{section}{0}
\title{COMPOSE: Comprehensive Portrait Shadow Editing Supplementary Materials} 

\titlerunning{COMPOSE: Comprehensive Portrait Shadow Editing}

\author{Andrew Hou\inst{1,2} \and
Zhixin Shu\inst{2} \and Xuaner Zhang\inst{2} \and He Zhang\inst{2} \and Yannick Hold-Geoffroy\inst{2} \and Jae Shin Yoon\inst{2} \and
Xiaoming Liu\inst{1}}

\authorrunning{A. Hou et al.}

\institute{Michigan State University \\ \email{\{houandr1,liuxm\}@msu.edu} 
\and Adobe Research \\
\email{\{zshu,cezhang,hezhan,holdgeof,jaeyoon\}@adobe.com}}

\maketitle

\section{Additional Discussions Regarding Prior Methods}

We delve into face relighting and shadow editing methods in greater detail to highlight the novel contributions of COMPOSE. We again emphasize that our core contribution is our decomposition of the environment map representation into ambient light and an editable gaussian dominant light source that can manipulate all shadow attributes (intensity, light size, and position). This also enables preserving the ambient light in the original source image through compositing with $\mathbf{I}_{D}$. As emphasized in the main paper, existing relighting methods~\cite{UCSDSingleImagePortraitRelighting, TotalRelighting, yeh2022learning} that use environment maps as the lighting representation will completely change the lighting environment to perform shadow editing by providing an entirely new environment map. This is unsuitable for many computational photography applications where the desire is to edit the facial shadows alone and to preserve the ambient light of the original environment. However, aside from these methods, some relighting and shadow editing methods are able to preserve the ambient light but lack a comprehensive set of shadow editing capabilities. Zhang \textit{et al.}~\cite{PortraitShadowManipulation} controls the degree of shadow softening but can't intensify shadows. They don't consider shadows resulting from decreasing light size and can't alter shadow position. Futschik \textit{el al.}~\cite{controllable-light-diffusion} are similarly able to control the degree of shadow softening while maintaining the ambience of the original photo, but they're completely limited to shadow softening. Hou \textit{et al.}~\cite{face-relighting-with-geometrically-consistent-shadows} model shadow position/intensity, but don't model light size and can't edit shadow shape. They also don't properly disentangle existing light sources from albedo and source image shadows will remain baked in. Nestmeyer \textit{et al.}~\cite{PhysicsGuidedRelighting} use a single light direction (xyz) and thus can model shadow position but not intensity or light size. \textbf{Notably, none of these methods can handle shrinking the light to both intensify the shadow and change its shape or need to completely modify the photo's environment in order to do so.} This is due to the lack of a suitable lighting decomposition, which is enabled by our method's separation of the environment map into ambient light and an editable gaussian dominant light source. The editable gaussian is what enables flexible control over all shadow attributes (including shrinking the light size), while the ambient light estimation helps preserve the remaining lighting attributes of the original photo. Another consideration that motivates our decomposition is that while it is possible to enlarge the light size of an existing environment map with a gaussian blur, the converse operation of shrinking the light size does not have a good solution. 

\section{Model Architectures}

We describe our model architectures for our light estimation VAE, our light diffusion hierarchical transformer, and our shadow synthesis pipeline in greater detail here.

\subsubsection{Light Estimation VAE}

Our light estimation VAE consists of a $6$-layer convolutional encoder followed by two fully-connected layers $FC_{\mu}$ and $FC_{\sigma}$ that predict $\mu$ and $\sigma$ respectively. We then sample $z=\epsilon\sigma+\mu$, where $\epsilon \sim N(0, 1)$ using the reparameterization trick, which becomes the input for our $4$-layer convolutional decoder. The decoder produces the final predicted environment map $\mathbf{E}_{P}$. Please see Tab.~\ref{tab:LightEstimationVAE} for more details. 

\renewcommand{\thetable}{2}
\begin{table}
\centering
\caption{\small
\textbf{Structure of Light Estimation VAE}. We describe in detail the structure of our lighting estimation VAE. $z$ is obtained by applying the reparameterization trick to the output of $FC_{mu}$ and $FC_{var}$. We apply batch normalization after all convolutional layers and all LeakyReLU layers use a slope of $0.2$. 
}\label{tab:LightEstimationVAE}
\scalebox{0.75}{
\setlength\tabcolsep{3pt}  
\begin{tabular}{c c c c c c c c}
\hline
Layer & Input & Type & Kernel Size & Stride & Input Features & Output Features & Activation \\
\hline
$conv1$ & $\mathbf{I}_{N}$ & Convolution & $3$ & $2$ & $3$ & $16$ & LeakyReLU \\
\hline
$conv2$ & $conv1$ & Convolution & $3$ & $2$ & $16$ & $32$ & LeakyReLU \\
\hline
$conv3$ & $conv2$ & Convolution & $3$ & $2$ & $32$ & $64$ & LeakyReLU \\
\hline
$conv4$ & $conv3$ & Convolution & $3$ & $2$ & $64$ & $128$ & LeakyReLU \\
\hline
$conv5$ & $conv4$ & Convolution & $3$ & $2$ & $128$ & $256$ & LeakyReLU \\
\hline
$conv6$ & $conv5$ & Convolution & $3$ & $2$ & $256$ & $512$ & LeakyReLU \\
\hline
$FC_{\mu}$ & $conv6$ & Fully Connected & - & - & $512*10*8$ & $512$ & None \\
\hline
$FC_{\sigma}$ & $conv6$ & Fully Connected & - & - & $512*10*8$ & $512$ & None \\
\hline
$FC_{z}$ & $z$ & Fully Connected & - & - & $512$ & $256*8*4$ & LeakyReLU \\
\hline
$deconv1$ & $FC_{z}$ & Transposed Convolution & $3$ & $2$ & $256$ & $128$ & LeakyReLU \\
\hline
$deconv2$ & $deconv1$ & Transposed Convolution & $3$ & $2$ & $128$ & $64$ & LeakyReLU \\
\hline
$deconv3$ & $deconv2$ & Transposed Convolution & $3$ & $2$ & $64$ & $32$ & LeakyReLU \\
\hline
$deconv4$ & $deconv3$ & Transposed Convolution & $3$ & $2$ & $32$ & $3$ & Sigmoid \\
\hline
\end{tabular}}
\end{table}

\subsubsection{Light Diffusion Transformer}

The light diffusion transformer's goal is to predict the ambient image $\mathbf{I}_{D}$. As such, it removes shadows and specularities from the source image $\mathbf{I}_{N}$. To accomplish this, we leverage an encoder-decoder structure with $3$ inputs: input image $\mathbf{I}_{N}$, a binary segmentation mask with the portrait foreground, and a body parsing mask. We leverage a hierarchical transformer encoder and divide the inputs into $4$x$4$ patches, which are more suitable for harmonization and shadow removal tasks than larger $16$x$16$ patches. We then obtain multi-level features using the transformer encoder at $\frac{1}{4}$, $\frac{1}{8}$, $\frac{1}{16}$, and $\frac{1}{32}$ of the original image resolution using multiple transformer blocks. These multi-level features are then passed to our decoder, which consists of transposed convolutional layers and generates the final ambient image $\mathbf{I}_{D}$.  

\subsubsection{Shadow Synthesis U-Net}

Our stage $1$ U-Net in our shadow synthesis pipeline is largely adopted from SIPR's~\cite{UCSDSingleImagePortraitRelighting} architecture with minor modifications. The main difference is in how we inject the lighting representation to the bottleneck of the U-Net. Instead of passing a $3$-channel environment map, we pass our $4$-channel spatial lighting representation $(x, y, \sigma, \gamma)$, as mentioned in the main paper. We find that this helps the network when training on smaller, concentrated lights compared to passing the standard $3$-channel environment map, likely because a small focused light occupies only a small portion of the environment map in pixel space and may thus provide a weak learning signal. Our method of encoding the light parameters numerically and replicating them spatially avoids this issue. 

\subsubsection{Shadow Synthesis DDPM}

For our stage $2$ conditional DDPM, we leverage the model of~\cite{ilvr_adm}. The role of the DDPM here is to take the output image of the U-Net $\mathbf{I}_{U}$ as a spatial condition along with the lighting parameters and perform image refinement to generate the final shadowed image $\mathbf{I}_{S}$. The training objective of our shadow synthesis DDPM thus becomes: 
\begin{equation}
 \Eb{t, \bx_0, \bepsilon}{ \left\| \bepsilon - \bepsilon_\theta(\mathbf{x}_{t}, t, \mathbf{I}_{U}, x, y, \sigma, \gamma) \right\|^2}. \label{eq:training_objective_conditional_ours}
\end{equation}
Our condition $\mathbf{I}_{U}$ is spatially concatenated with $\mathbf{x}_{t}$ and our lighting parameters are repeated spatially as channels of the same resolution and similarly spatially concatenated, where
 $\mathbf{x}_{t}= \sqrt{\bar\alpha_t}\bx_0 + \sqrt{1-\bar\alpha_t}\bepsilon$.

\renewcommand{\thefigure}{8}
 \begin{figure}
    \centering
    \includegraphics[width=\linewidth]{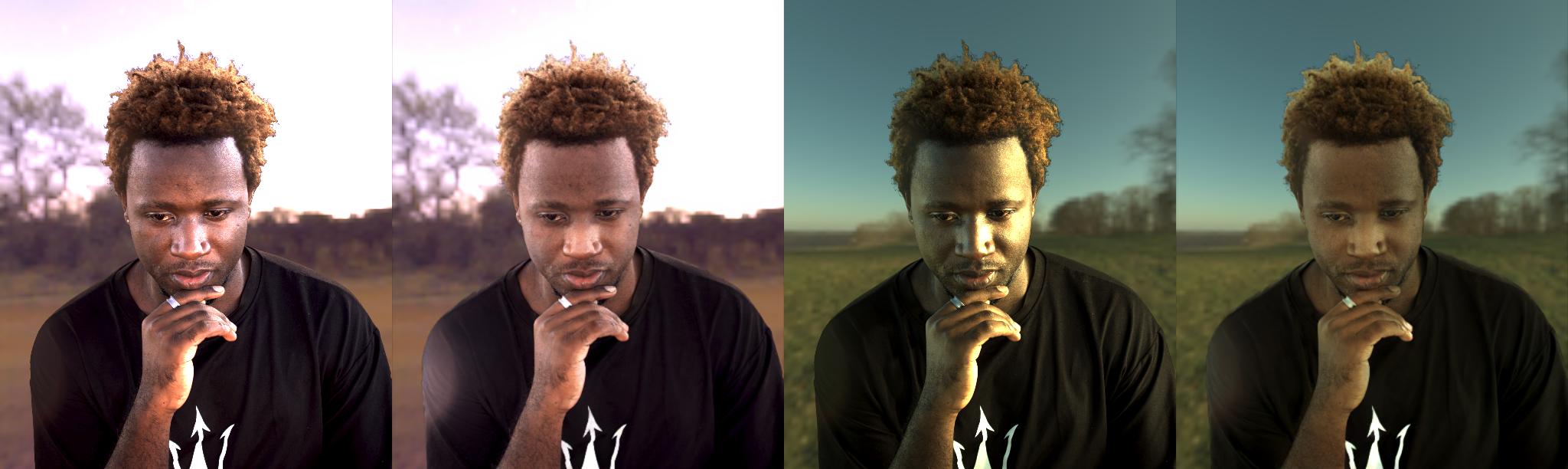} 
\vspace{-6mm}
\caption{\small \textbf{Ambient Light}. COMPOSE is able to preserve the ambient light of the original portrait thanks to the hierarchical transformer used in the light diffusion stage. The light diffusion stage removes specularities and existing shadows caused by strong directional lights, leaving behind only the ambient light. Each pair of images is left: input, right: ambient prediction. 
}\label{fig:AmbientLight}
\end{figure}

\renewcommand{\thetable}{3}
\begin{table}[t!]
    \centering
    \caption{\small
\textbf{Skin Tone Fairness}. COMPOSE performs well on darker skin tones (Dark) as well as intermediate intensity skin tones (Tan). On light skin tones (Light), SSIM/LPIPS are comparable with Dark and Tan and MAE/MSE are slightly worse, mostly due to the larger errors accrued from incorrect shadow boundaries on light skin subjects. 
}\label{tab:SkinTonePerformance}
    \begin{tabular}{c c c c c}
\hline
 & MAE & MSE & SSIM & LPIPS \\
\hline
Light & $0.1141$ & $0.0449$ & $0.7756$ & $0.1880$\\
\hline
Tan & $0.0724$ & $0.0239$ & $0.7821$ & $0.2074$ \\
\hline
Dark & $0.0755$ & $0.0207$ & $0.7798$ & $0.2034$\\
\hline
All Skin Tones & $0.0965$ & $0.0349$ & $0.7780$ & $0.1973$\\
\hline
    \end{tabular}
\end{table}

\section{Skin Tone Fairness} As with any face editing work, the ability to represent and generalize to diverse subjects is of the upmost importance. We therefore quantitatively evaluate how well COMPOSE performs across different skin tones. See Tab.~\ref{tab:SkinTonePerformance}, where we separated our light stage test subjects into $3$ groups (Light, Tan, and Dark) based on mean skin tone values. Performance on Dark/Tan is good whereas Light has comparable SSIM/LPIPS but worse MAE/MSE. This is because errors involving shadows are maximized for light skin since the difference between shadows and light skin is higher than for darker skin. This can be addressed by increasing the number of light skin training subjects or increasing loss weights in shadow regions to prevent large errors caused by slightly misaligned shadow boundaries. 

\section{Ambient Light} Since preserving the ambient light of the portrait is an important component of COMPOSE, we demonstrate some results from our light diffusion network. Ambient light is respected by our light diffusion network, and should be different for the same face illuminated by $2$ different environment maps (See Fig.~\ref{fig:AmbientLight}, left:input, right:ambient). 

\section{Foreign Shadow Removal} Our light diffusion network is able to remove not just self shadows, but also foreign shadows and can thus serve as a shadow removal network. Please see Fig.~\ref{fig:ForeignShadowRemoval} for some sample results on portraits with substantial foreign shadows. 
\vspace{-2mm}
\renewcommand{\thefigure}{9}
\begin{figure}[t]
\vspace{-2mm}
\begin{center}

\begin{minipage}[t]{0.13\linewidth}
\centering
\includegraphics[width=\linewidth]{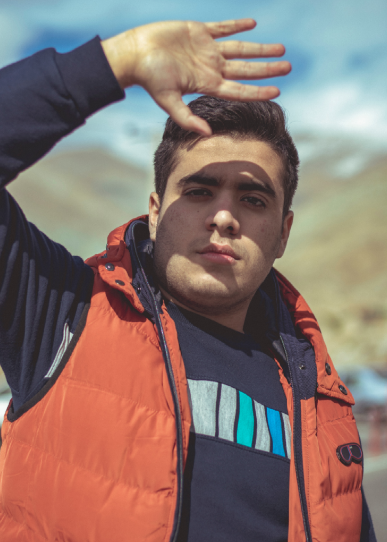} \\
\end{minipage}
\begin{minipage}[t]{0.13\linewidth}
\centering
\includegraphics[width=\linewidth]{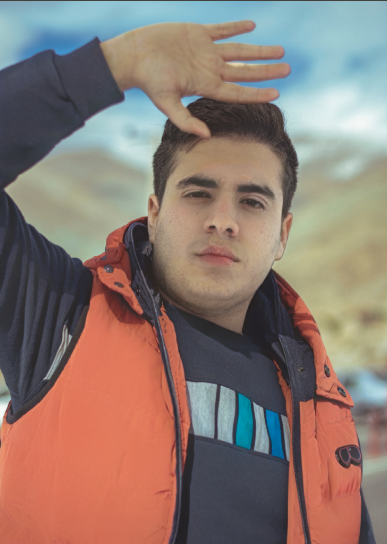} \\
\end{minipage}
\begin{minipage}[t]{0.18\linewidth}
\centering
\includegraphics[width=\linewidth]{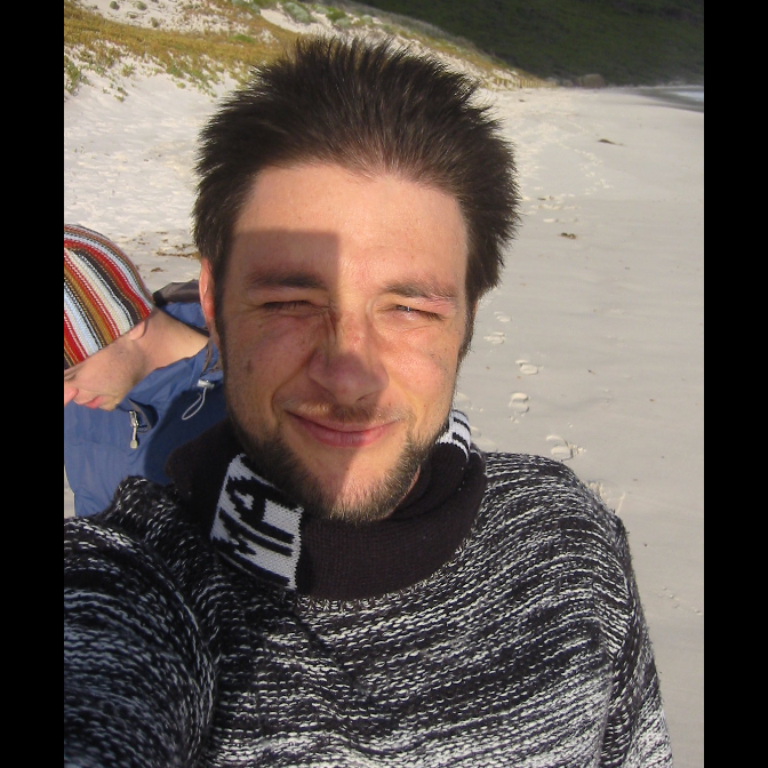} \\
\end{minipage}
\begin{minipage}[t]{0.18\linewidth}
\centering
\includegraphics[width=\linewidth]{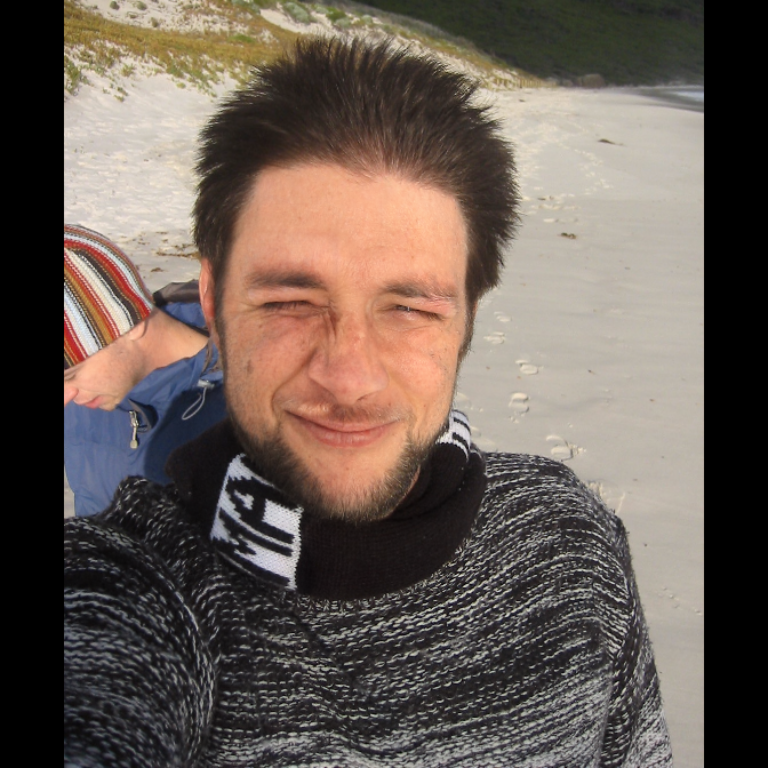} \\
\end{minipage}
\begin{minipage}[t]{0.155\linewidth}
\centering
\includegraphics[width=\linewidth]{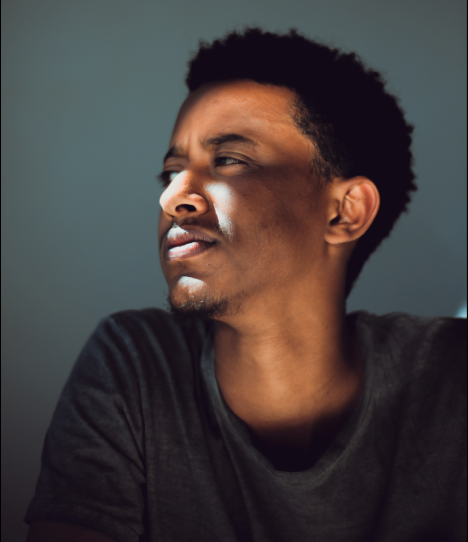} \\
\end{minipage}
\begin{minipage}[t]{0.155\linewidth}
\centering
\includegraphics[width=\linewidth]{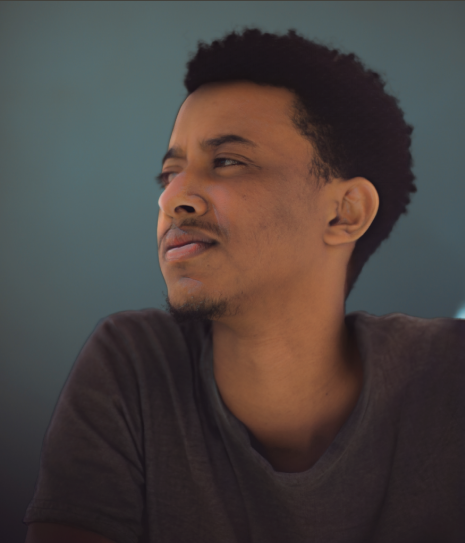} \\
\end{minipage}

\vspace{-2mm}
\caption{\small \textbf{Foreign Shadow Removal}. Our light diffusion network can serve as a shadow removal method, and even handles foreign shadow removal well. 
}\label{fig:ForeignShadowRemoval}
\end{center}
\vspace{-5mm}
\end{figure}

\renewcommand{\thefigure}{10}
\begin{figure}[t]
\begin{center}

\begin{minipage}[t]{0.15\linewidth}
\centering
\includegraphics[width=\linewidth]{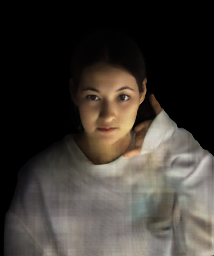} \\
\end{minipage}
\begin{minipage}[t]{0.15\linewidth}
\centering
\includegraphics[width=\linewidth]{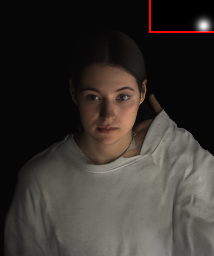} \\
\end{minipage}
\begin{minipage}[t]{0.15\linewidth}
\centering
\includegraphics[width=\linewidth]{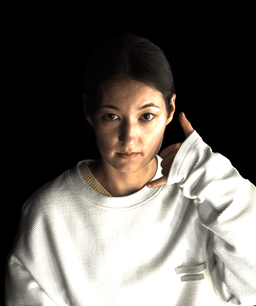} \\
\end{minipage}
\begin{minipage}[t]{0.15\linewidth}
\centering
\includegraphics[width=\linewidth]{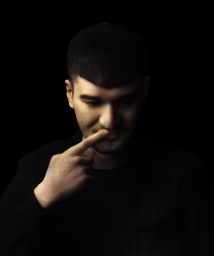} \\
\end{minipage}
\begin{minipage}[t]{0.15\linewidth}
\centering
\includegraphics[width=\linewidth]{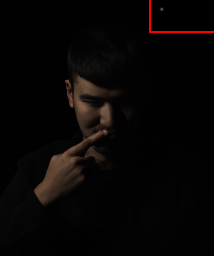} \\
\end{minipage}
\begin{minipage}[t]{0.15\linewidth}
\centering
\includegraphics[width=\linewidth]{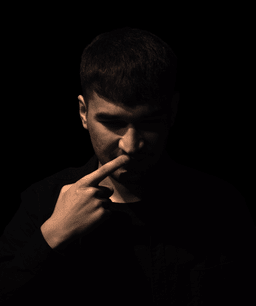} \\
\end{minipage}

\vspace{-2mm}
\caption{\small \textbf{Diffusion Ablation}. We ablate whether adding a DDPM after our shadow synthesis U-Net improves the shadow editing performance. Each triplet represents U-Net only, U-Net+DDPM, and groundtruth target image. The DDPM substantially improves the visual quality of the images and sharpens the shadow boundaries. The target lighting condition is visualized in the middle image of each triplet. 
}\label{fig:DiffusionAblation}
\end{center}
\vspace{-5mm}
\end{figure}

\renewcommand{\thetable}{4}
\begin{table}[t!]
    \centering
    \caption{\small
\textbf{Diffusion Ablation}. Adding the DDPM to the shadow synthesis stage slightly worsens the quantitative metrics, but greatly enhances visual quality. 
}\label{tab:DiffusionAblation}
 \begin{tabular}{c c c c c}
\hline
 & MAE & MSE & SSIM & LPIPS \\
\hline
U-Net Only & $0.0831$ & $0.0226$ & $0.7977$ & $0.1734$\\
\hline
U-Net+DDPM & $0.0965$ & $0.0349$ & $0.7780$ & $0.1973$\\
\hline
\end{tabular}
\end{table}

\section{Diffusion Ablation} See Fig.~\ref{fig:DiffusionAblation}, where every triplet is U-Net only, U-Net+DDPM, and groundtruth. Adding the DDPM slightly worsens our metrics (See Tab.~\ref{tab:DiffusionAblation}) but greatly enhances the visual quality of the images and sharpens shadow boundaries. 

\section{Additional Shadow Removal Results}

\renewcommand{\thefigure}{11}
\begin{figure*}[t]
\vspace{-2mm}
\begin{center}

\begin{minipage}[t]{0.15\linewidth}
\centering
\includegraphics[width=\linewidth]{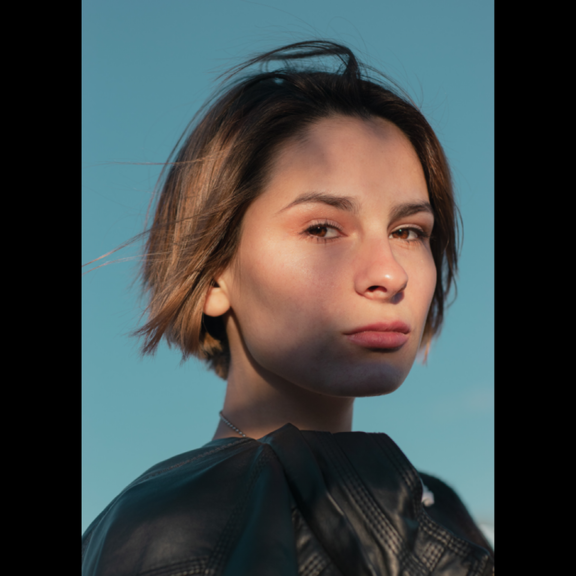} \\ 
\end{minipage}
\begin{minipage}[t]{0.15\linewidth}
\centering
\includegraphics[width=\linewidth]{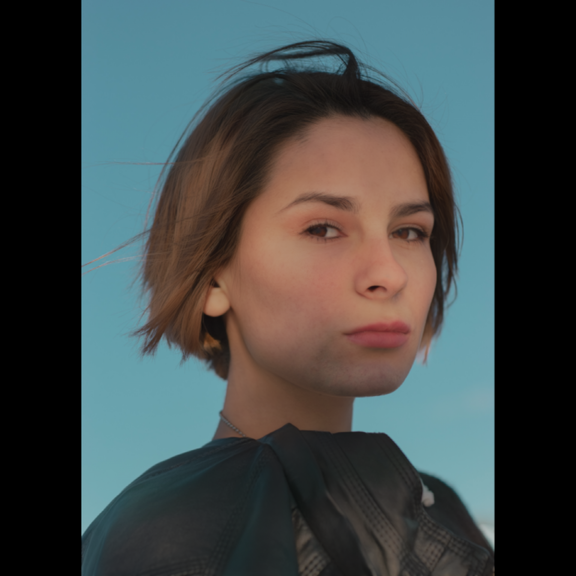} \\
\end{minipage}
\begin{minipage}[t]{0.15\linewidth}
\centering
\includegraphics[width=\linewidth]{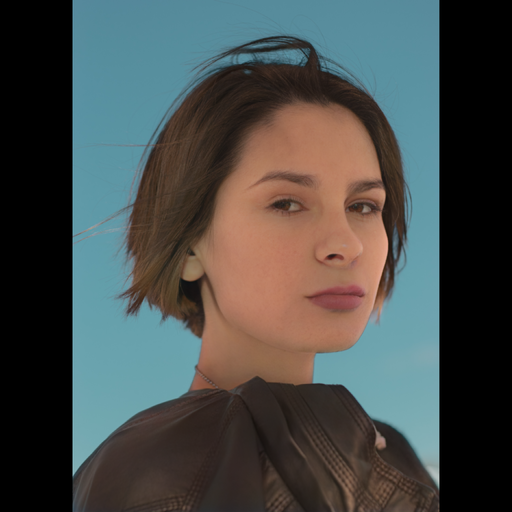} \\
\end{minipage}
\begin{minipage}[t]{0.15\linewidth}
\centering
\includegraphics[width=\linewidth]{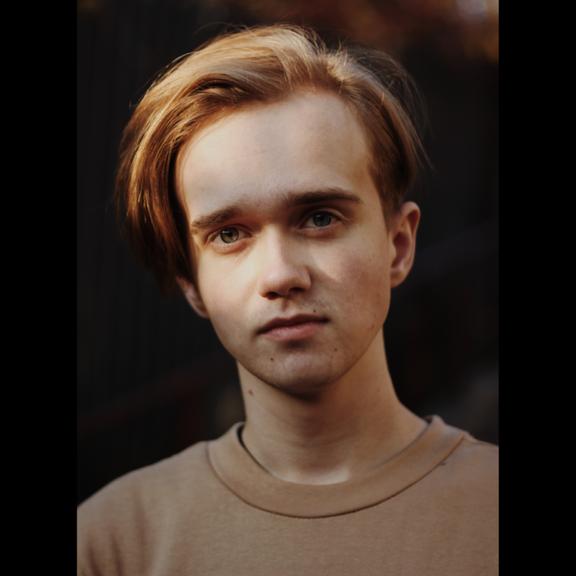} \\
\end{minipage}
\begin{minipage}[t]{0.15\linewidth}
\centering
\includegraphics[width=\linewidth]{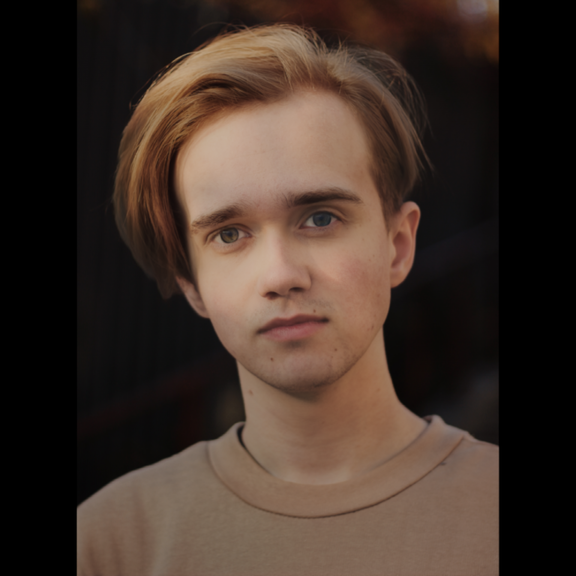} \\
\end{minipage}
\begin{minipage}[t]{0.15\linewidth}
\centering
\includegraphics[width=\linewidth]{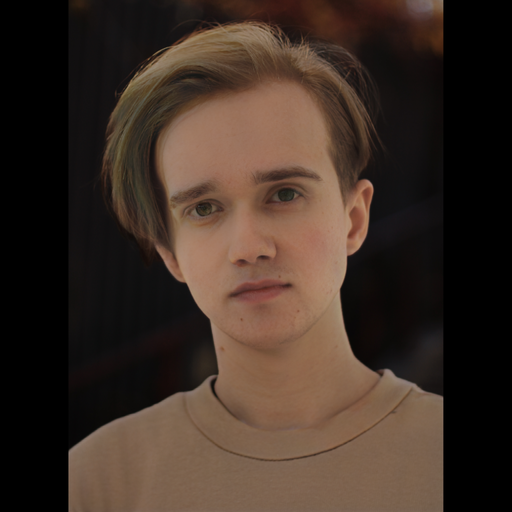} \\
\end{minipage}

\begin{minipage}[t]{0.15\linewidth}
\centering
\includegraphics[width=\linewidth]{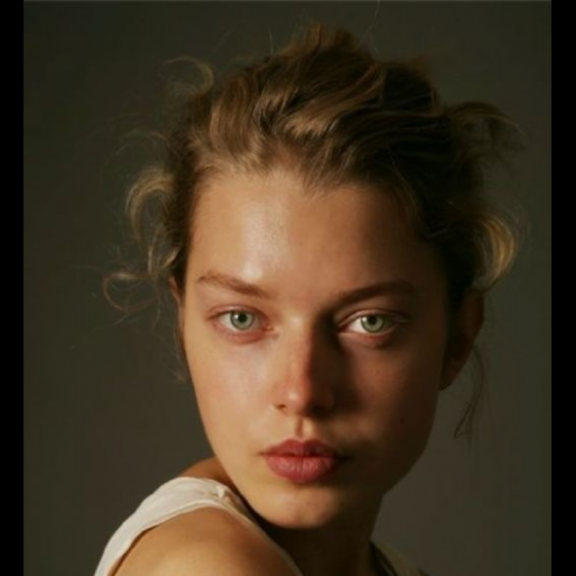} \\
\end{minipage}
\begin{minipage}[t]{0.15\linewidth}
\centering
\includegraphics[width=\linewidth]{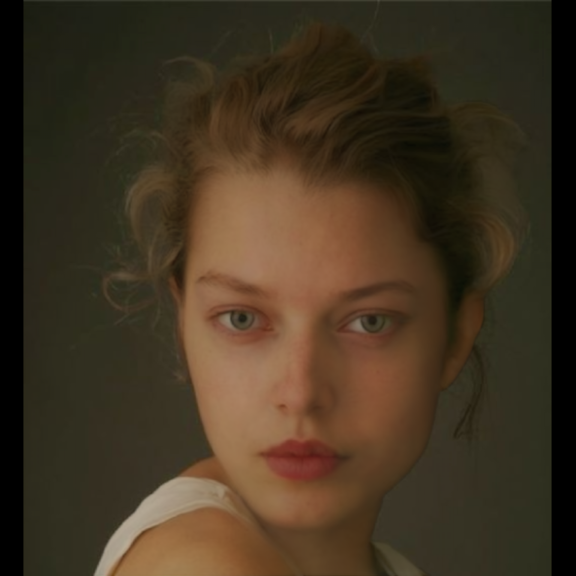} \\
\end{minipage}
\begin{minipage}[t]{0.15\linewidth}
\centering
\includegraphics[width=\linewidth]{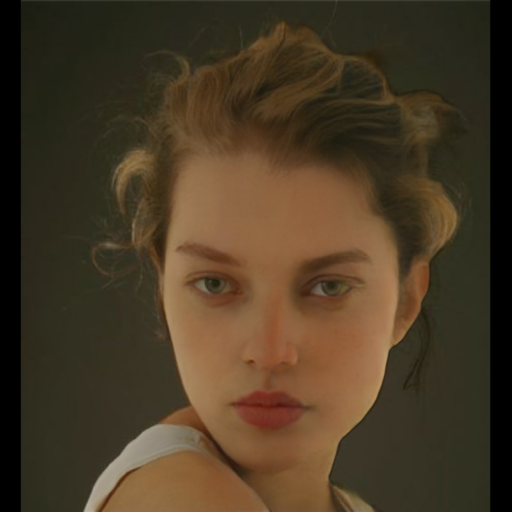} \\
\end{minipage}
\begin{minipage}[t]{0.15\linewidth}
\centering
\includegraphics[width=\linewidth]{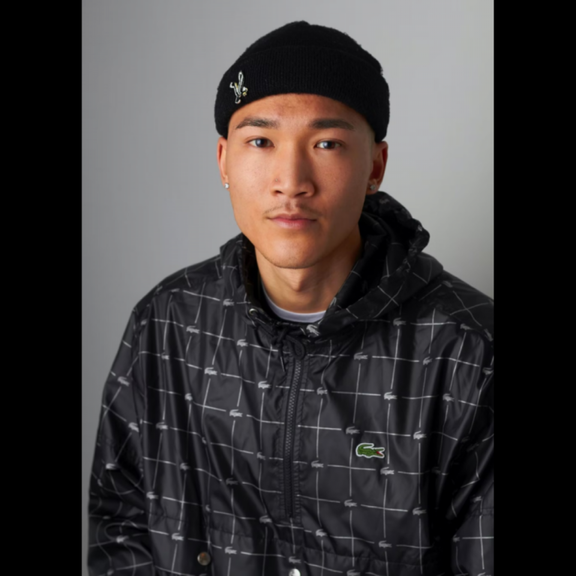} \\
\end{minipage}
\begin{minipage}[t]{0.15\linewidth}
\centering
\includegraphics[width=\linewidth]{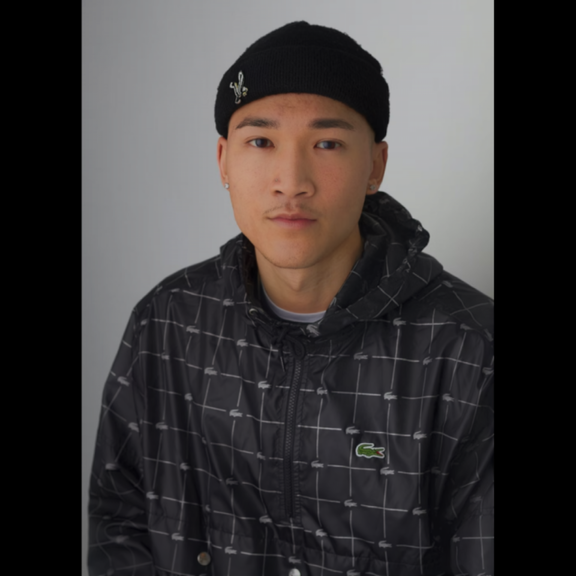} \\
\end{minipage}
\begin{minipage}[t]{0.15\linewidth}
\centering
\includegraphics[width=\linewidth]{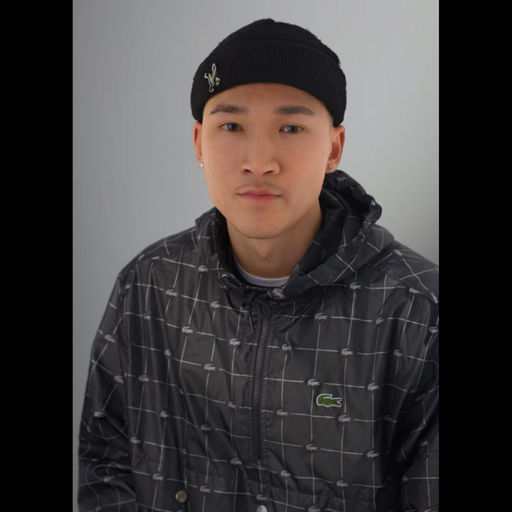} \\
\end{minipage}

\begin{minipage}[t]{0.15\linewidth}
\centering
\includegraphics[width=\linewidth]{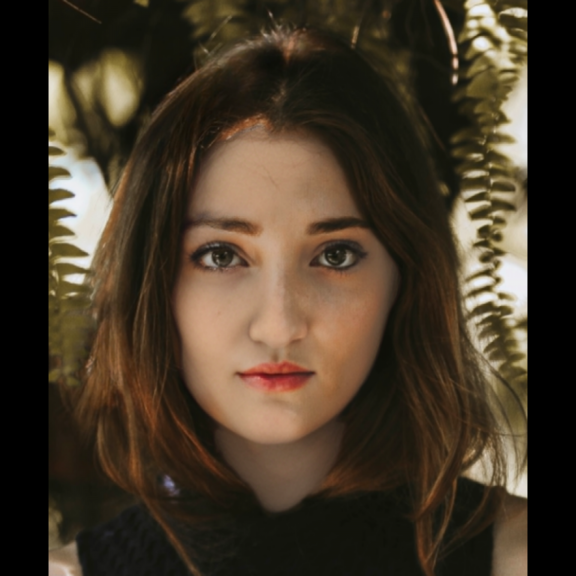} \\ 
\end{minipage}
\begin{minipage}[t]{0.15\linewidth}
\centering
\includegraphics[width=\linewidth]{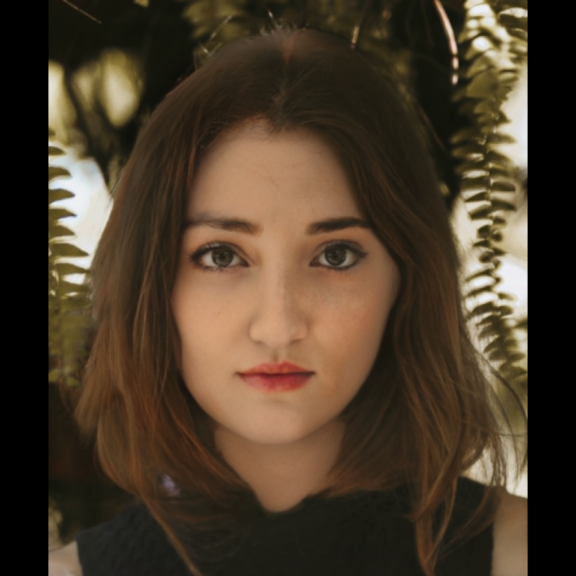} \\
\end{minipage}
\begin{minipage}[t]{0.15\linewidth}
\centering
\includegraphics[width=\linewidth]{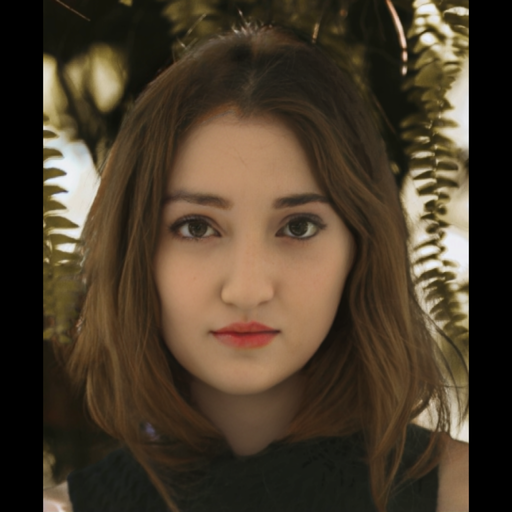} \\
\end{minipage}
\begin{minipage}[t]{0.15\linewidth}
\centering
\includegraphics[width=\linewidth]{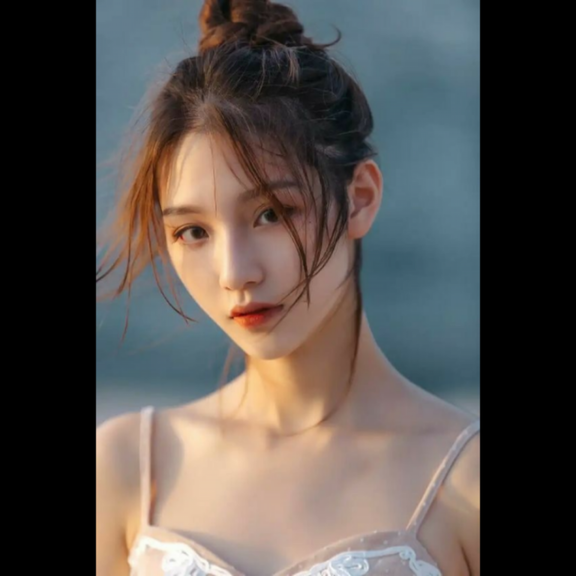} \\
\end{minipage}
\begin{minipage}[t]{0.15\linewidth}
\centering
\includegraphics[width=\linewidth]{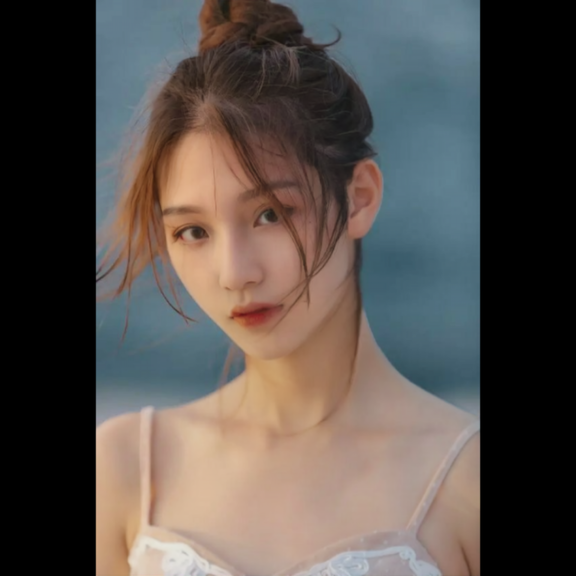} \\
\end{minipage}
\begin{minipage}[t]{0.15\linewidth}
\centering
\includegraphics[width=\linewidth]{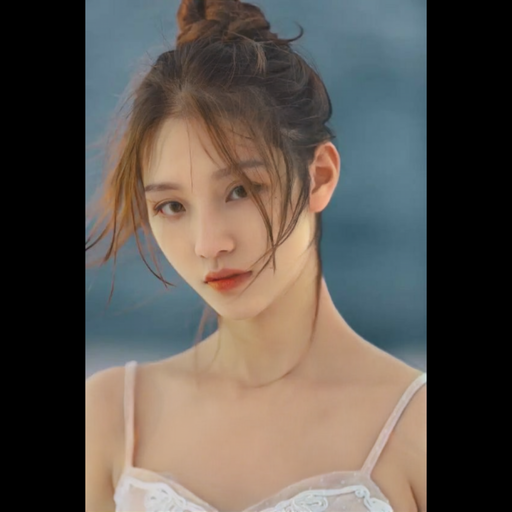} \\
\end{minipage}
\begin{minipage}[t]{0.15\linewidth}
\centering
\includegraphics[width=\linewidth]{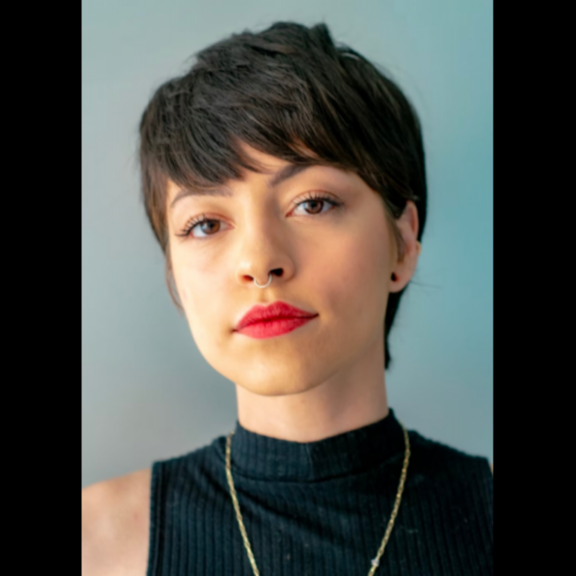} \\
\end{minipage}
\begin{minipage}[t]{0.15\linewidth}
\centering
\includegraphics[width=\linewidth]{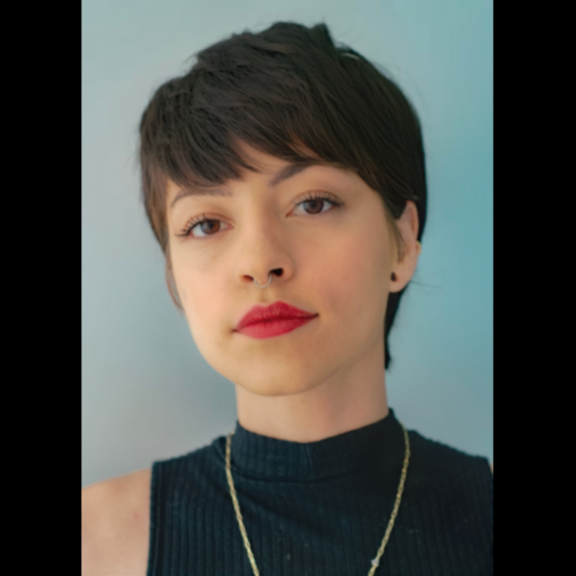} \\
\end{minipage}
\begin{minipage}[t]{0.15\linewidth}
\centering
\includegraphics[width=\linewidth]{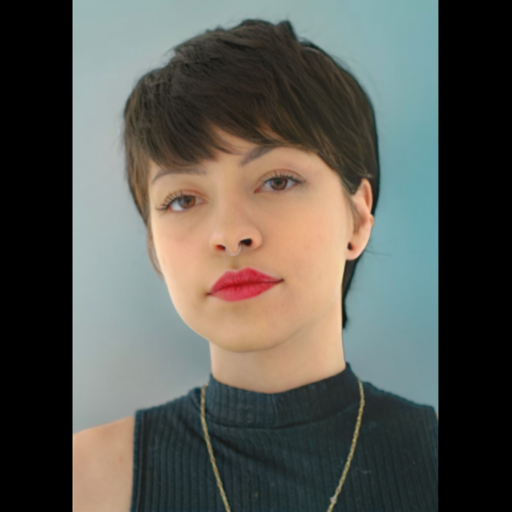} \\
\end{minipage}
\begin{minipage}[t]{0.15\linewidth}
\centering
\includegraphics[width=\linewidth]{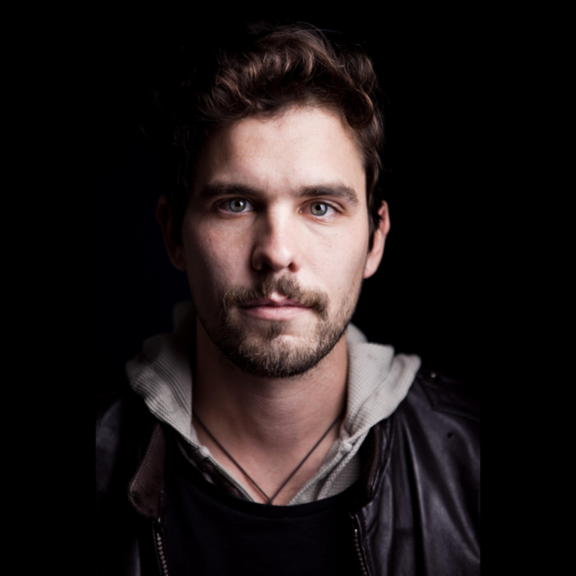} \\
\end{minipage}
\begin{minipage}[t]{0.15\linewidth}
\centering
\includegraphics[width=\linewidth]{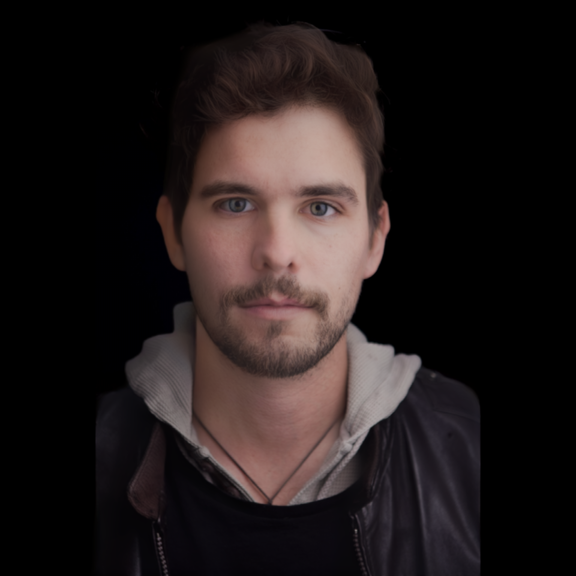} \\
\end{minipage}
\begin{minipage}[t]{0.15\linewidth}
\centering
\includegraphics[width=\linewidth]{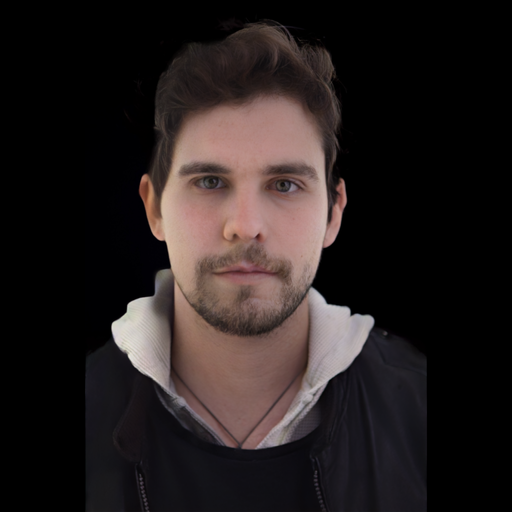} \\
\end{minipage}

\begin{minipage}[t]{0.15\linewidth}
\centering
\includegraphics[width=\linewidth]{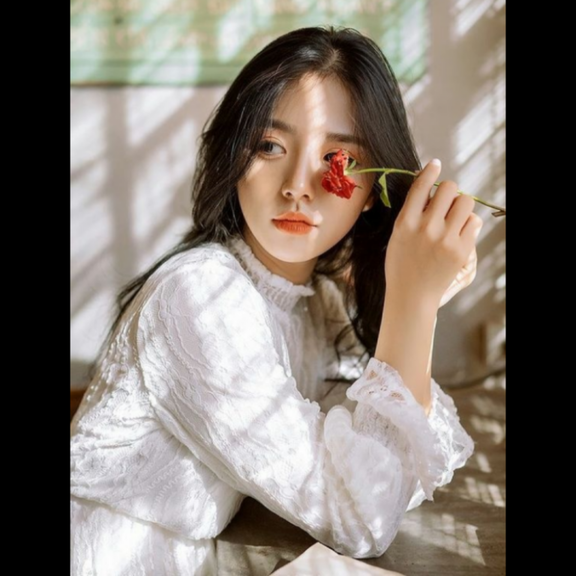} \\ 
\end{minipage}
\begin{minipage}[t]{0.15\linewidth}
\centering
\includegraphics[width=\linewidth]{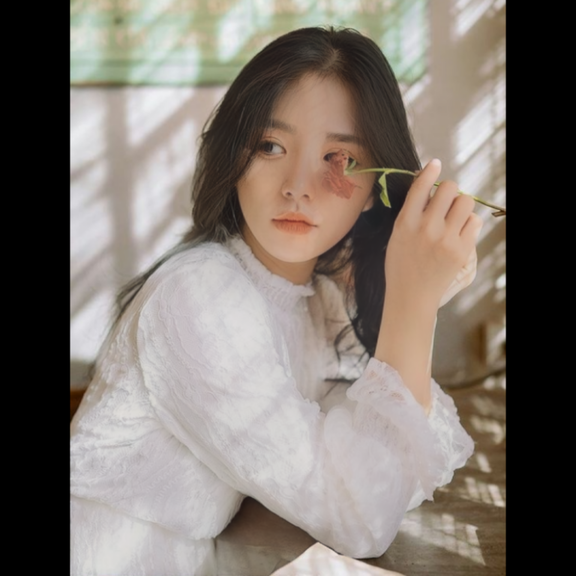} \\
\end{minipage}
\begin{minipage}[t]{0.15\linewidth}
\centering
\includegraphics[width=\linewidth]{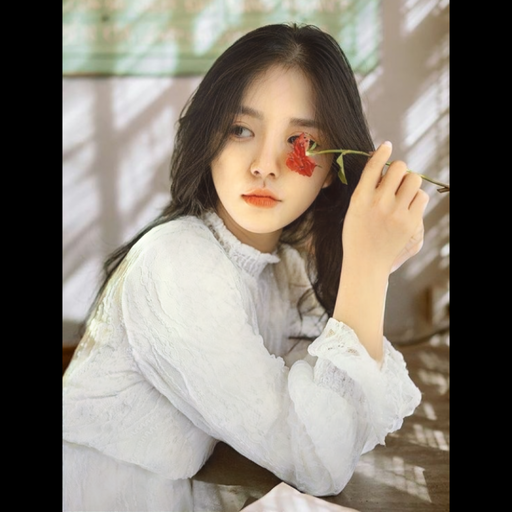} \\
\end{minipage}
\begin{minipage}[t]{0.15\linewidth}
\centering
\includegraphics[width=\linewidth]{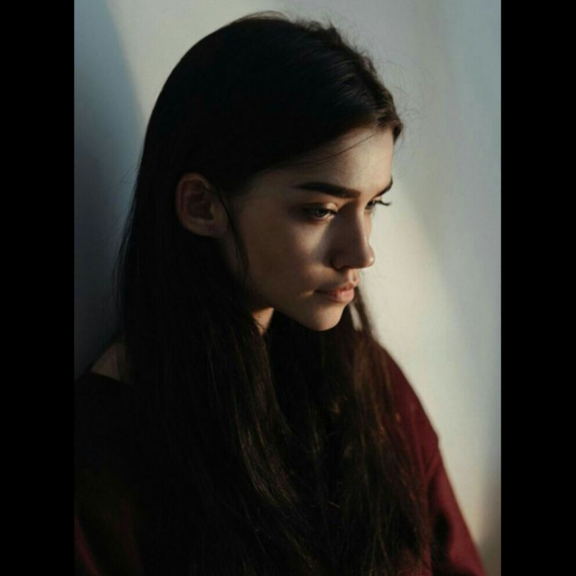} \\
\end{minipage}
\begin{minipage}[t]{0.15\linewidth}
\centering
\includegraphics[width=\linewidth]{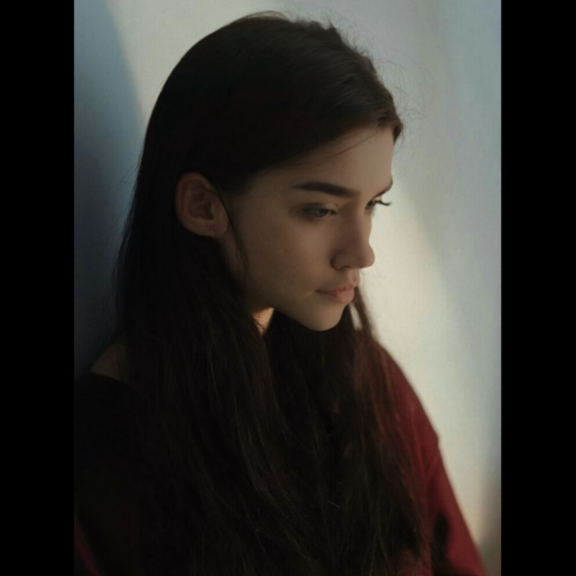} \\
\end{minipage}
\begin{minipage}[t]{0.15\linewidth}
\centering
\includegraphics[width=\linewidth]{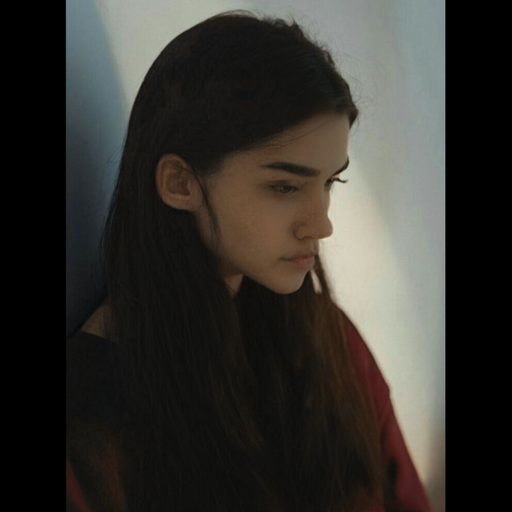} \\
\end{minipage}
\begin{minipage}[t]{0.15\linewidth}
\centering
\includegraphics[width=\linewidth]{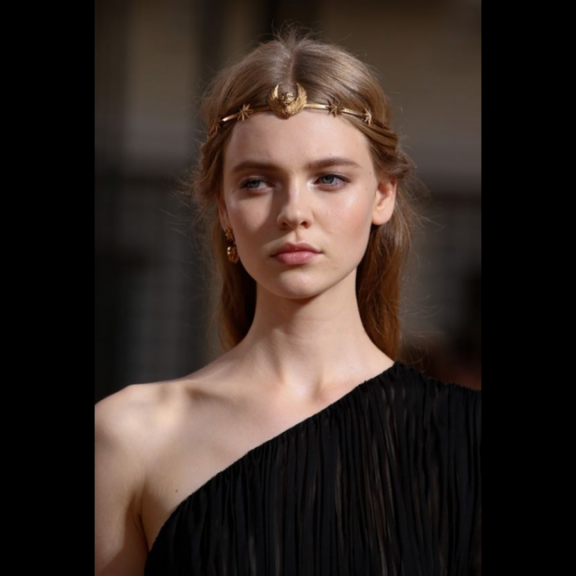} \\
\end{minipage}
\begin{minipage}[t]{0.15\linewidth}
\centering
\includegraphics[width=\linewidth]{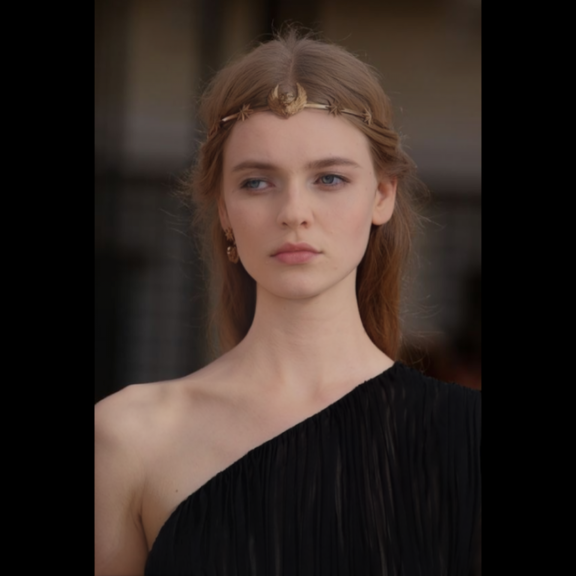} \\
\end{minipage}
\begin{minipage}[t]{0.15\linewidth}
\centering
\includegraphics[width=\linewidth]{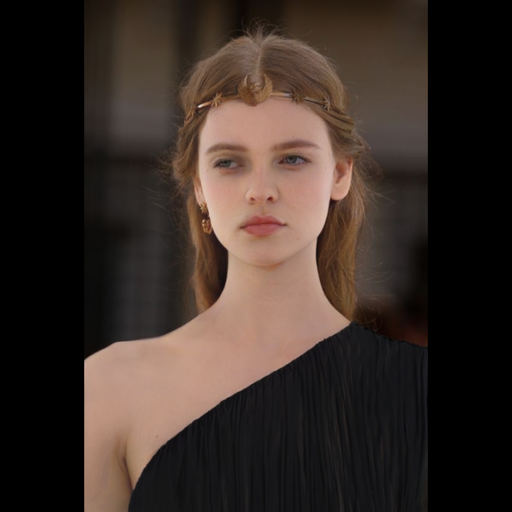} \\
\end{minipage}
\begin{minipage}[t]{0.15\linewidth}
\centering
\includegraphics[width=\linewidth]{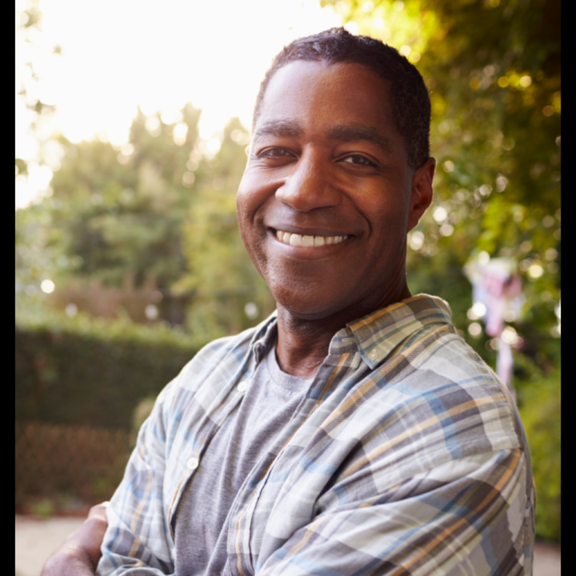} \\
\end{minipage}
\begin{minipage}[t]{0.15\linewidth}
\centering
\includegraphics[width=\linewidth]{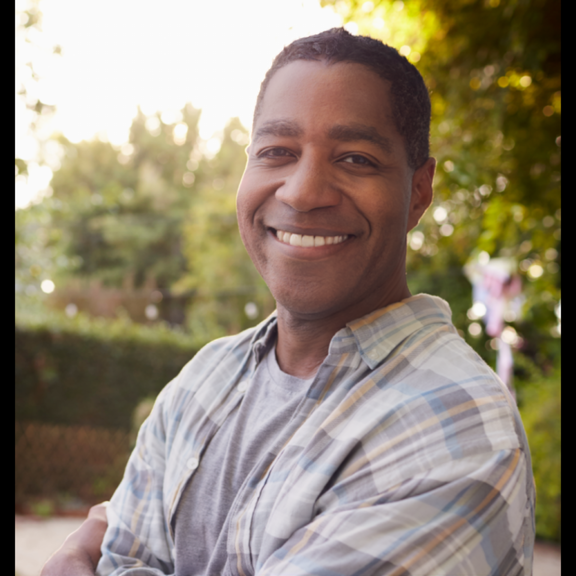} \\
\end{minipage}
\begin{minipage}[t]{0.15\linewidth}
\centering
\includegraphics[width=\linewidth]{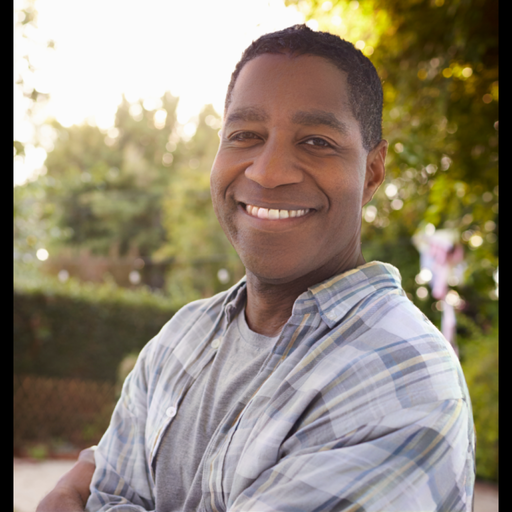} \\
\end{minipage}

\begin{minipage}[t]{0.15\linewidth}
\centering
\includegraphics[width=\linewidth]{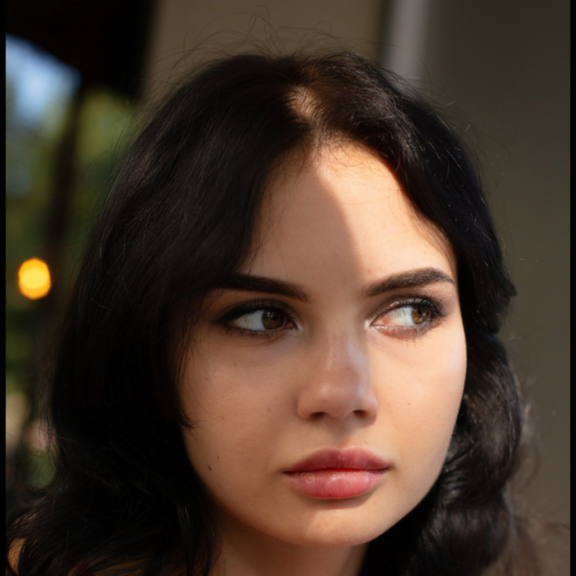} \\ 
\end{minipage}
\begin{minipage}[t]{0.15\linewidth}
\centering
\includegraphics[width=\linewidth]{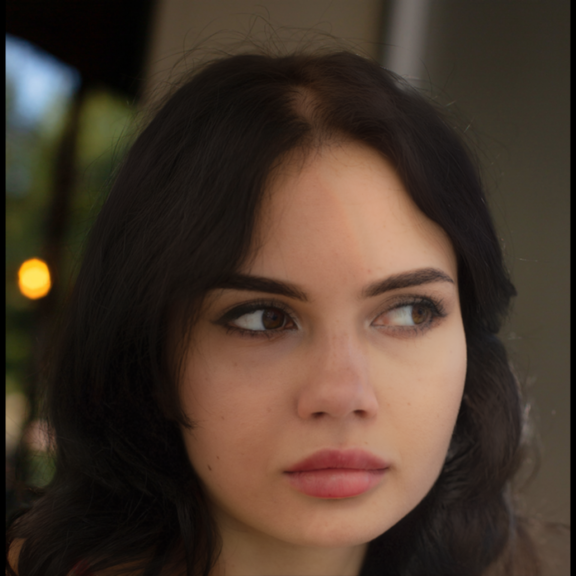} \\
\end{minipage}
\begin{minipage}[t]{0.15\linewidth}
\centering
\includegraphics[width=\linewidth]{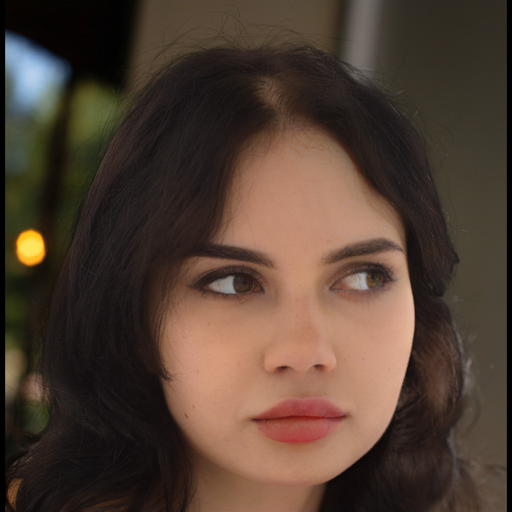} \\
\end{minipage}
\begin{minipage}[t]{0.15\linewidth}
\centering
\includegraphics[width=\linewidth]{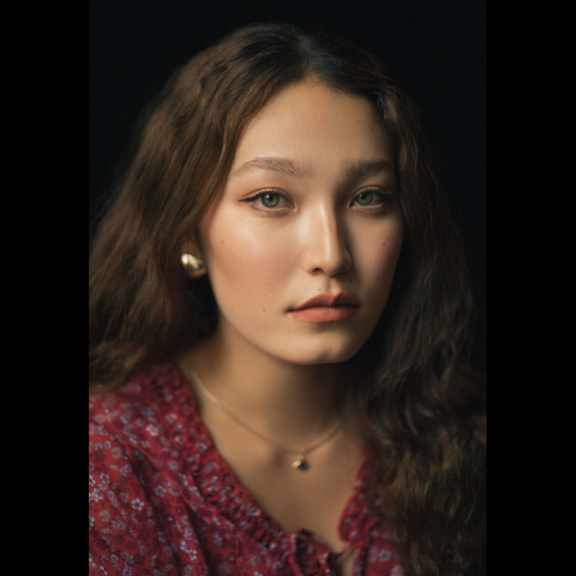} \\
\end{minipage}
\begin{minipage}[t]{0.15\linewidth}
\centering
\includegraphics[width=\linewidth]{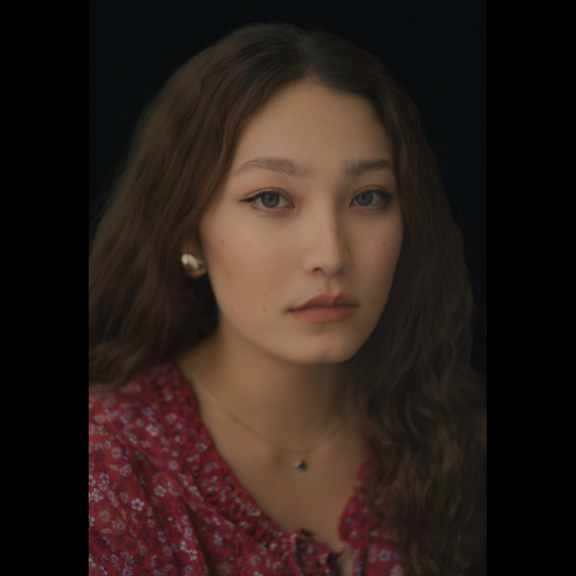} \\
\end{minipage}
\begin{minipage}[t]{0.15\linewidth}
\centering
\includegraphics[width=\linewidth]{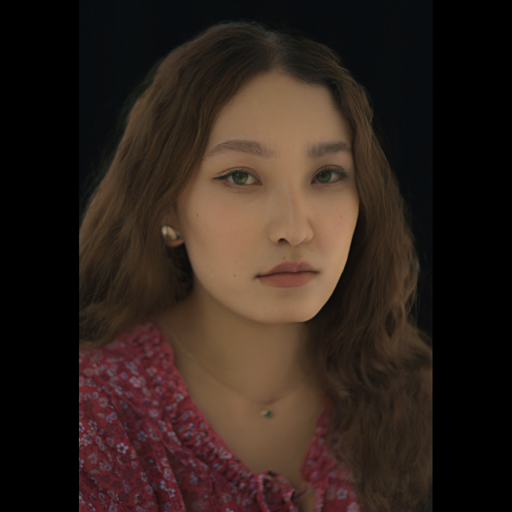} \\
\end{minipage}

\begin{minipage}[t]{0.15\linewidth}
\centering
\includegraphics[width=\linewidth]{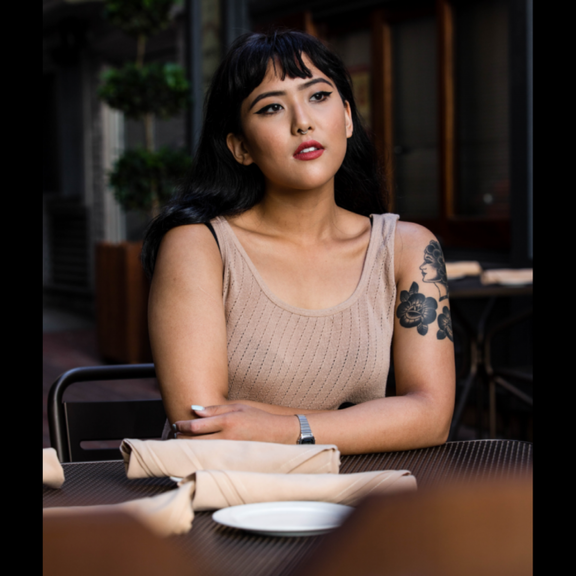} \\
\small a) Source
\end{minipage}
\begin{minipage}[t]{0.15\linewidth}
\centering
\includegraphics[width=\linewidth]{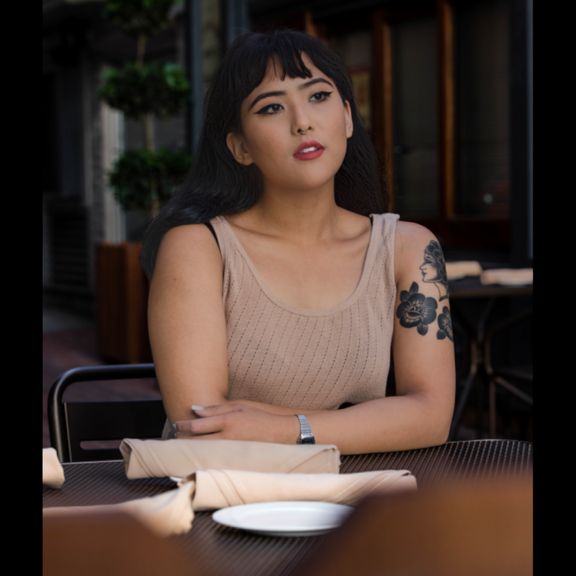} \\
\small b) ~\cite{controllable-light-diffusion}
\end{minipage}
\begin{minipage}[t]{0.15\linewidth}
\centering
\includegraphics[width=\linewidth]{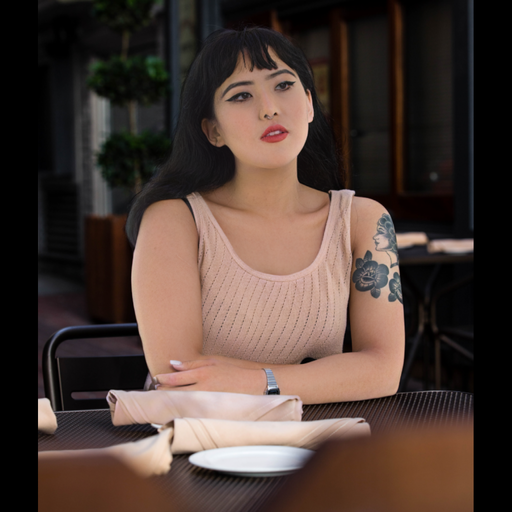} \\
\small c) Ours
\end{minipage}
\begin{minipage}[t]{0.15\linewidth}
\centering
\includegraphics[width=\linewidth]{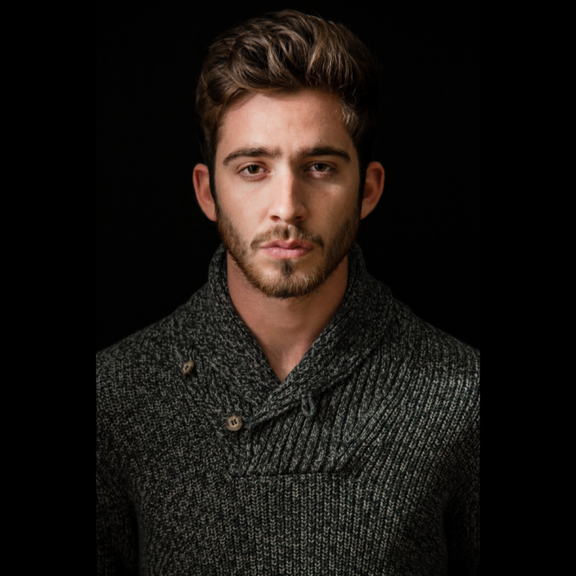} \\
\small d) Source
\end{minipage}
\begin{minipage}[t]{0.15\linewidth}
\centering
\includegraphics[width=\linewidth]{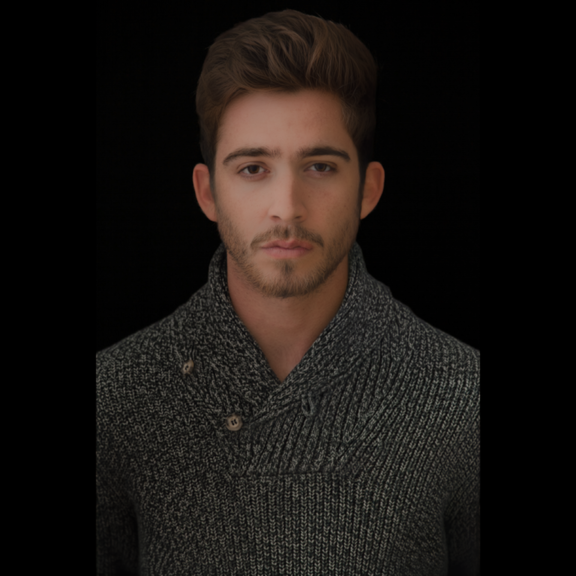} \\
\small e) ~\cite{controllable-light-diffusion}
\end{minipage}
\begin{minipage}[t]{0.15\linewidth}
\centering
\includegraphics[width=\linewidth]{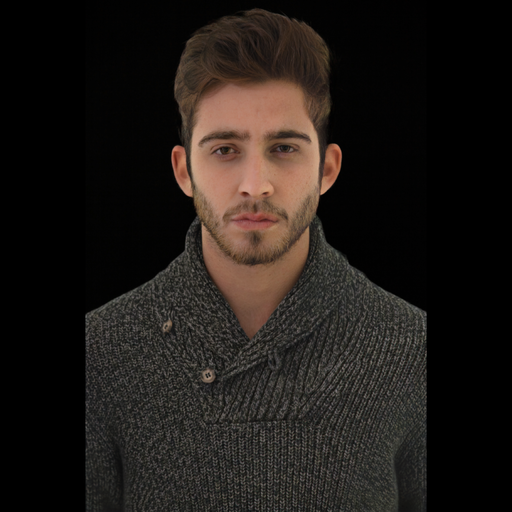} \\
\small f) Ours
\end{minipage}

\vspace{-2mm}
\caption{\small \textbf{Shadow Removal Comparisons with Futschik \textit{et al.}~\cite{controllable-light-diffusion}}. We demonstrate all qualitative shadow removal results against Futschik \textit{et al.}~\cite{controllable-light-diffusion} without cherry picking. COMPOSE is able to leave less of a shadow trace, especially for darker shadows when the goal is complete shadow removal. 
}\label{fig:ShadowRemoval}
\end{center}\vspace{-4mm}
\end{figure*}

\renewcommand{\thefigure}{11}
\begin{figure*}[t]
\vspace{-2mm}
\begin{center}

\begin{minipage}[t]{0.15\linewidth}
\centering
\includegraphics[width=\linewidth]{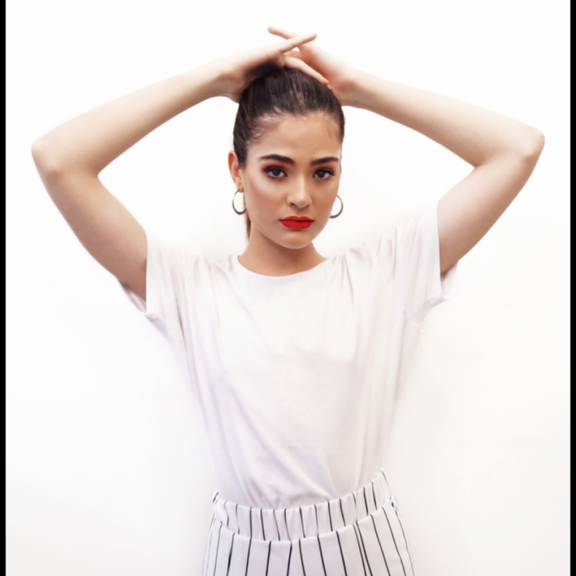} \\ 
\end{minipage}
\begin{minipage}[t]{0.15\linewidth}
\centering
\includegraphics[width=\linewidth]{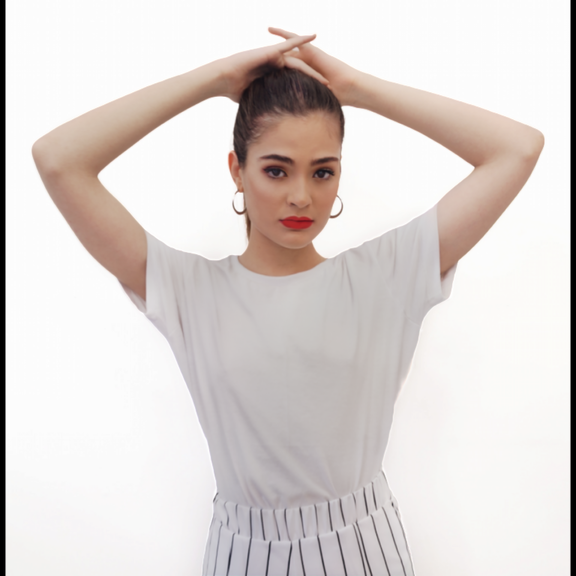} \\
\end{minipage}
\begin{minipage}[t]{0.15\linewidth}
\centering
\includegraphics[width=\linewidth]{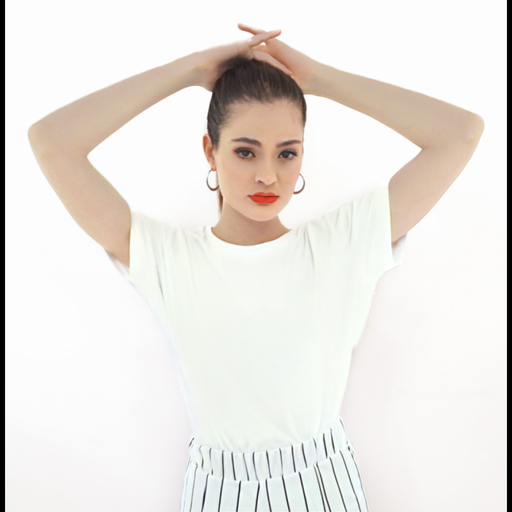} \\
\end{minipage}
\begin{minipage}[t]{0.15\linewidth}
\centering
\includegraphics[width=\linewidth]{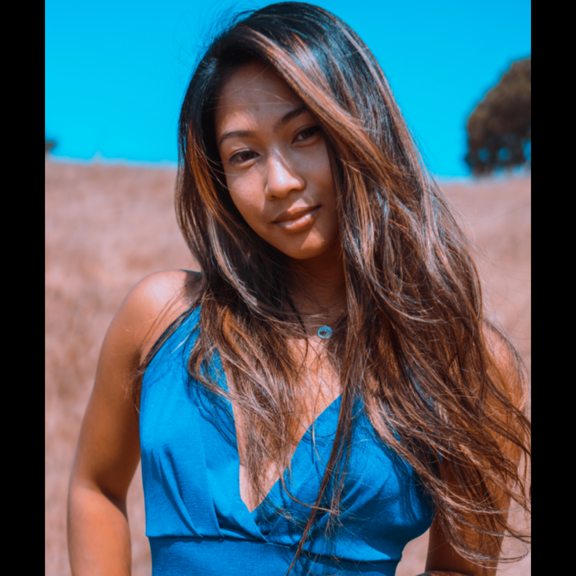} \\
\end{minipage}
\begin{minipage}[t]{0.15\linewidth}
\centering
\includegraphics[width=\linewidth]{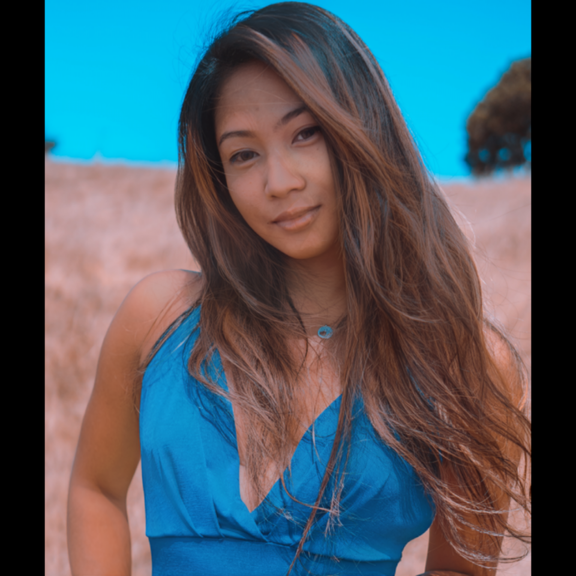} \\
\end{minipage}
\begin{minipage}[t]{0.15\linewidth}
\centering
\includegraphics[width=\linewidth]{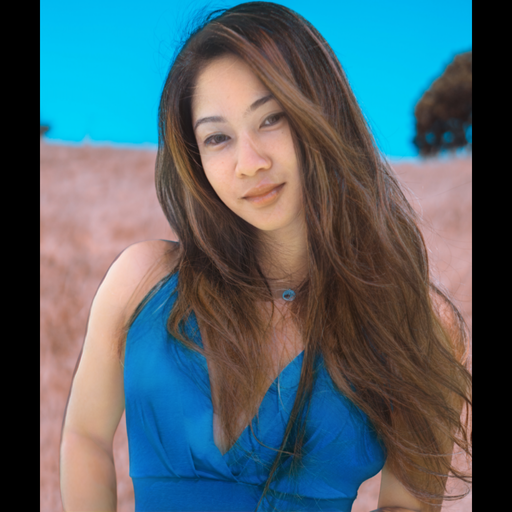} \\
\end{minipage}
\begin{minipage}[t]{0.15\linewidth}
\centering
\includegraphics[width=\linewidth]{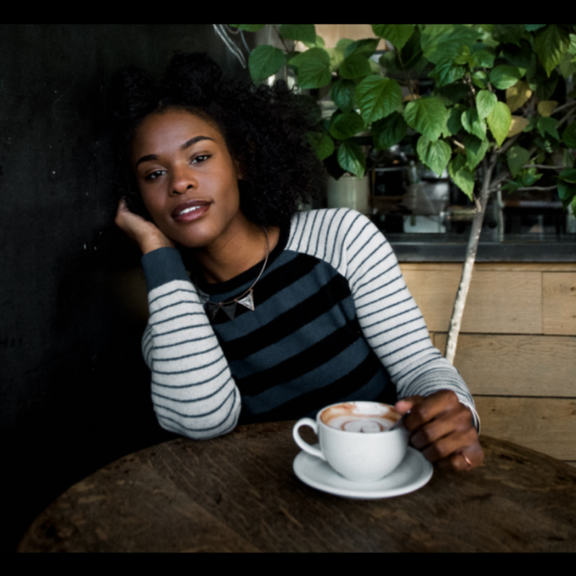} \\
\end{minipage}
\begin{minipage}[t]{0.15\linewidth}
\centering
\includegraphics[width=\linewidth]{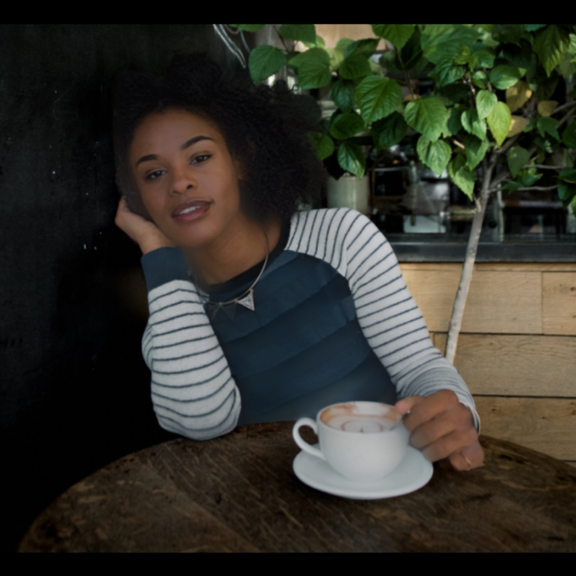} \\
\end{minipage}
\begin{minipage}[t]{0.15\linewidth}
\centering
\includegraphics[width=\linewidth]{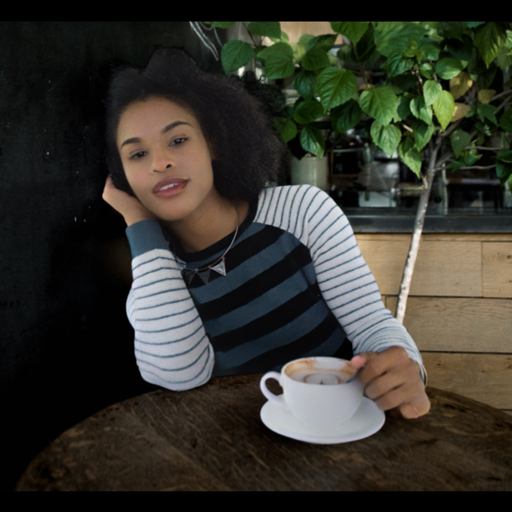} \\
\end{minipage}
\begin{minipage}[t]{0.15\linewidth}
\centering
\includegraphics[width=\linewidth]{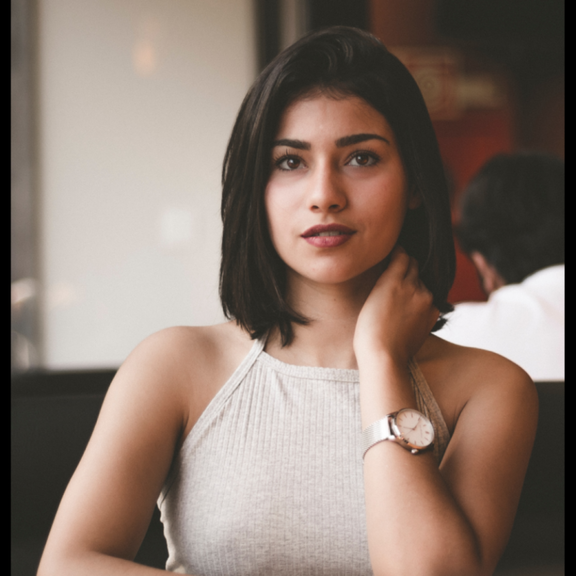} \\
\end{minipage}
\begin{minipage}[t]{0.15\linewidth}
\centering
\includegraphics[width=\linewidth]{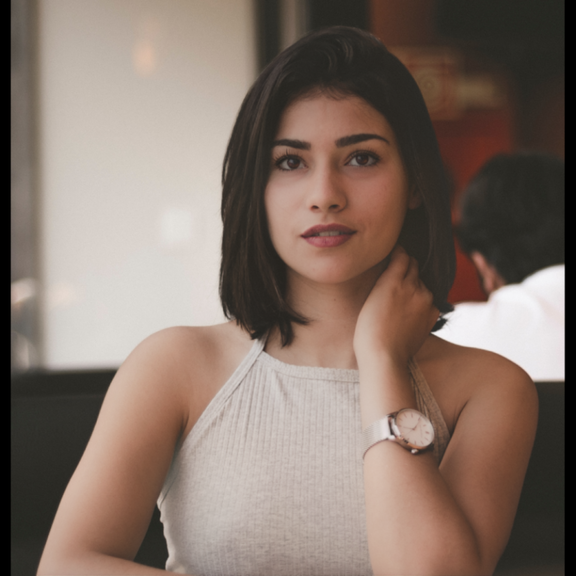} \\
\end{minipage}
\begin{minipage}[t]{0.15\linewidth}
\centering
\includegraphics[width=\linewidth]{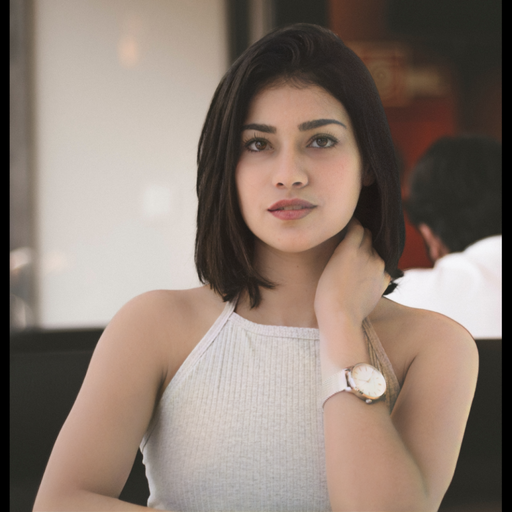} \\
\end{minipage}

\begin{minipage}[t]{0.15\linewidth}
\centering
\includegraphics[width=\linewidth]{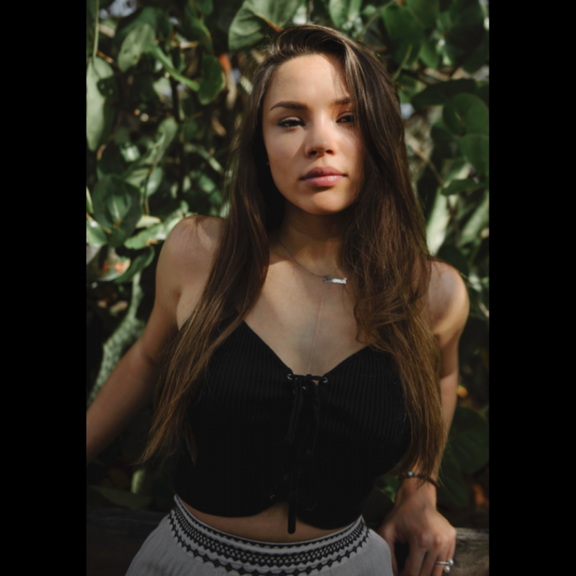} \\ 
\end{minipage}
\begin{minipage}[t]{0.15\linewidth}
\centering
\includegraphics[width=\linewidth]{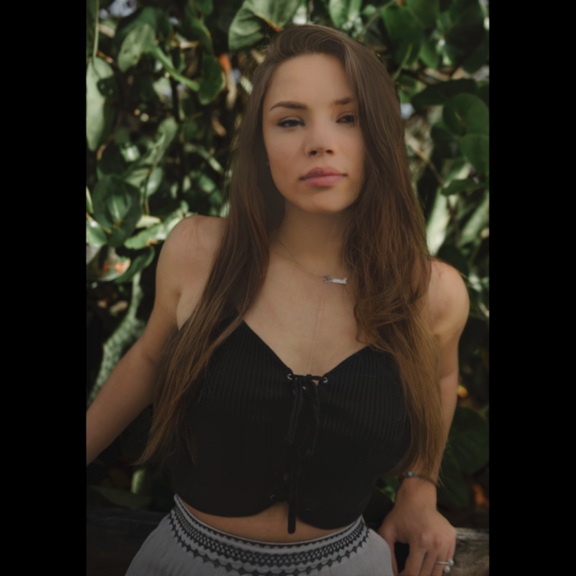} \\
\end{minipage}
\begin{minipage}[t]{0.15\linewidth}
\centering
\includegraphics[width=\linewidth]{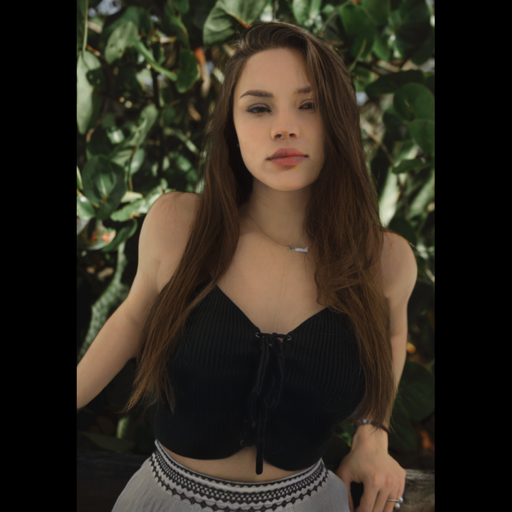} \\
\end{minipage}
\begin{minipage}[t]{0.15\linewidth}
\centering
\includegraphics[width=\linewidth]{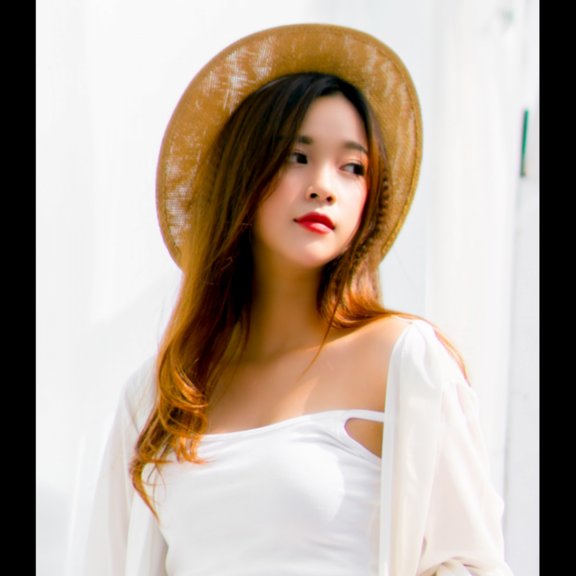} \\
\end{minipage}
\begin{minipage}[t]{0.15\linewidth}
\centering
\includegraphics[width=\linewidth]{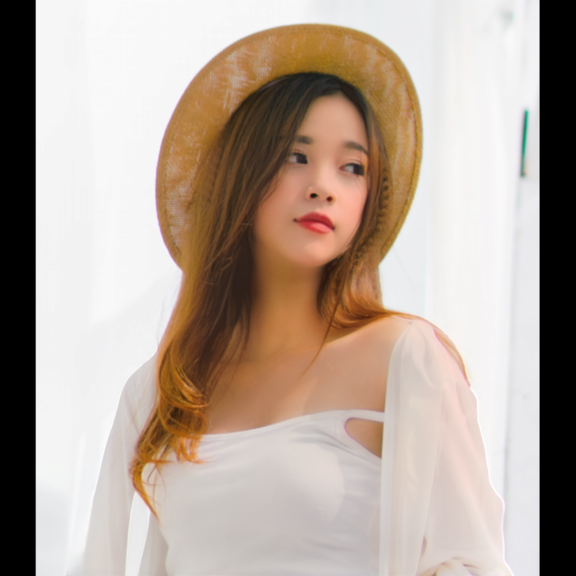} \\
\end{minipage}
\begin{minipage}[t]{0.15\linewidth}
\centering
\includegraphics[width=\linewidth]{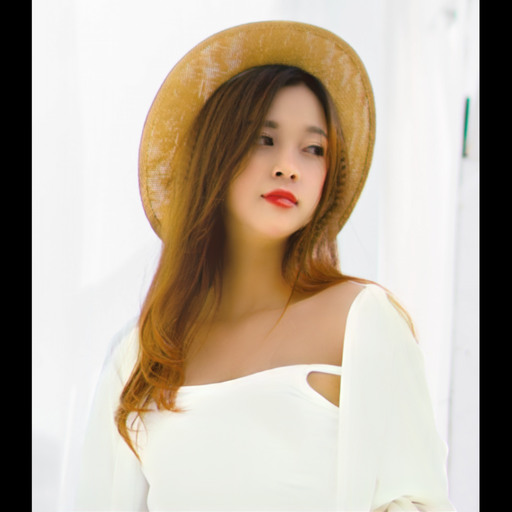} \\
\end{minipage}
\begin{minipage}[t]{0.15\linewidth}
\centering
\includegraphics[width=\linewidth]{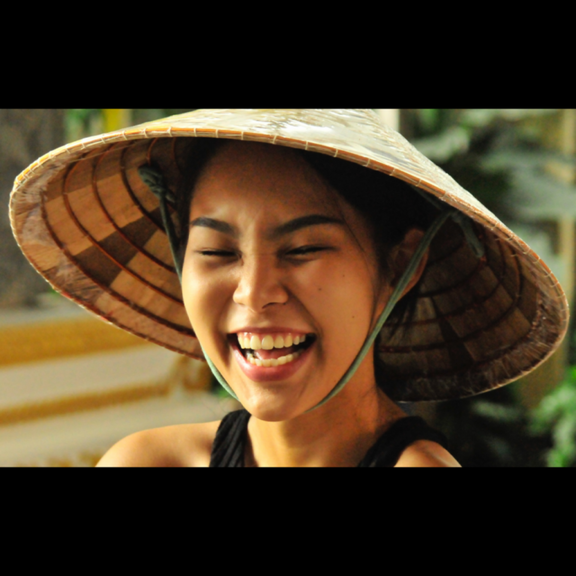} \\
\end{minipage}
\begin{minipage}[t]{0.15\linewidth}
\centering
\includegraphics[width=\linewidth]{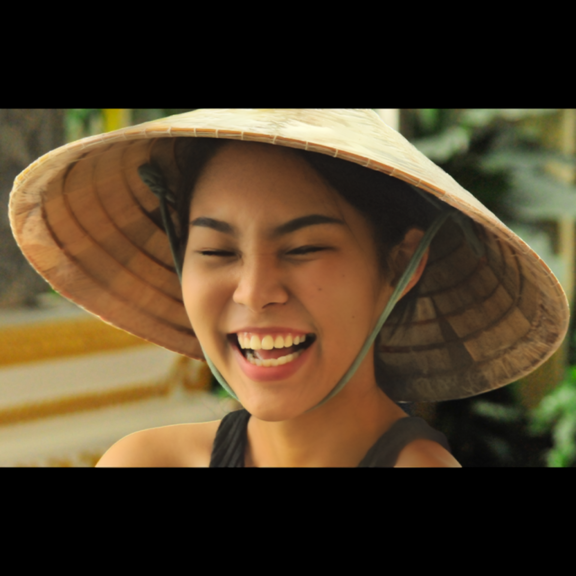} \\
\end{minipage}
\begin{minipage}[t]{0.15\linewidth}
\centering
\includegraphics[width=\linewidth]{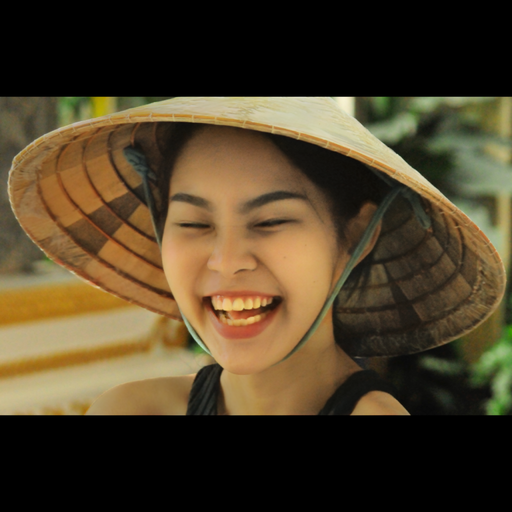} \\
\end{minipage}
\begin{minipage}[t]{0.15\linewidth}
\centering
\includegraphics[width=\linewidth]{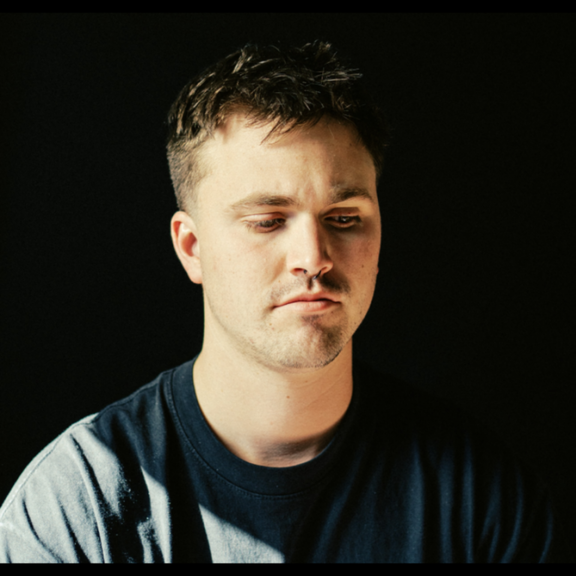} \\
\end{minipage}
\begin{minipage}[t]{0.15\linewidth}
\centering
\includegraphics[width=\linewidth]{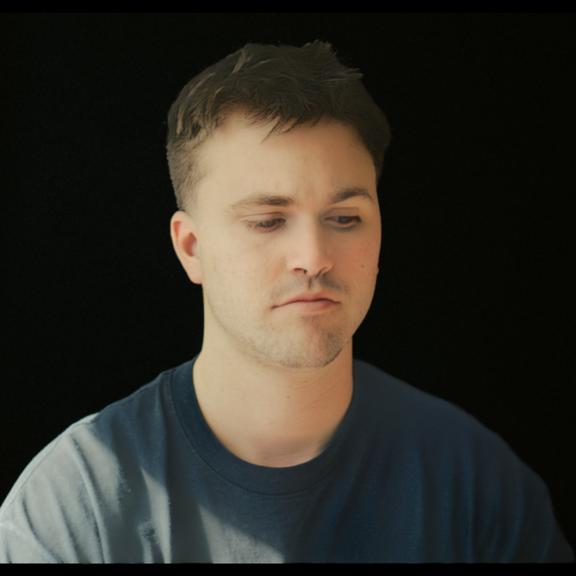} \\
\end{minipage}
\begin{minipage}[t]{0.15\linewidth}
\centering
\includegraphics[width=\linewidth]{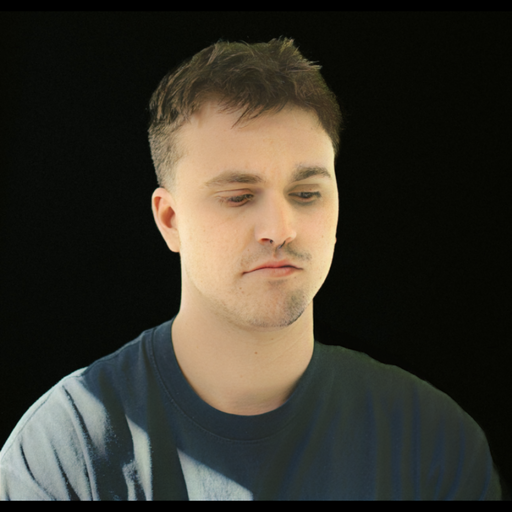} \\
\end{minipage}

\begin{minipage}[t]{0.15\linewidth}
\centering
\includegraphics[width=\linewidth]{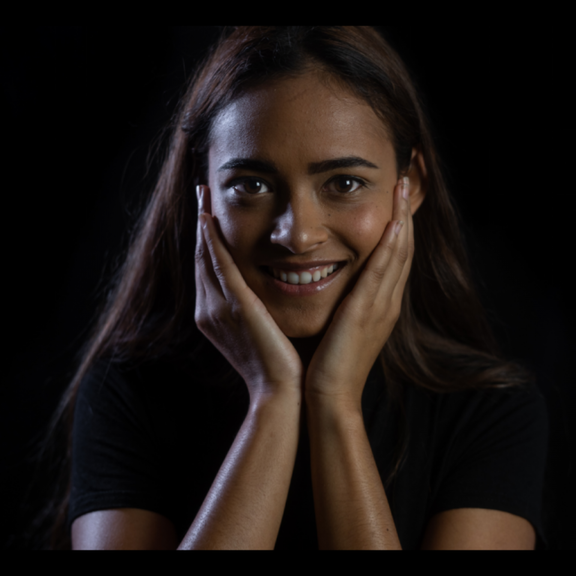} \\ 
\end{minipage}
\begin{minipage}[t]{0.15\linewidth}
\centering
\includegraphics[width=\linewidth]{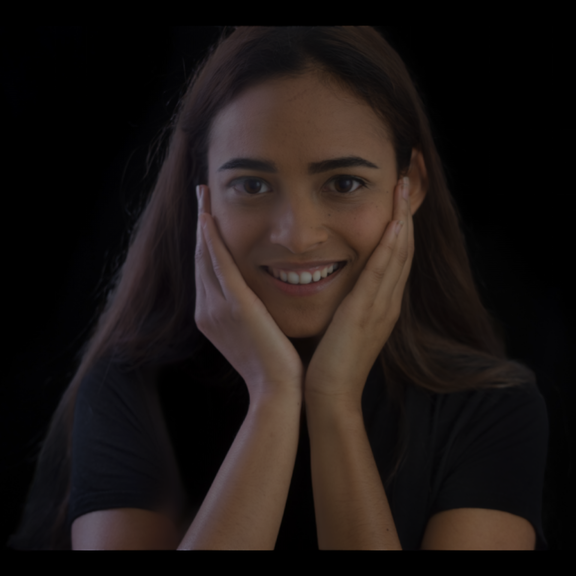} \\
\end{minipage}
\begin{minipage}[t]{0.15\linewidth}
\centering
\includegraphics[width=\linewidth]{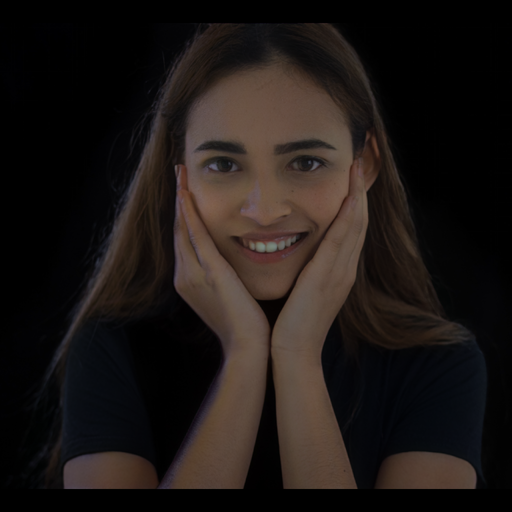} \\
\end{minipage}
\begin{minipage}[t]{0.15\linewidth}
\centering
\includegraphics[width=\linewidth]{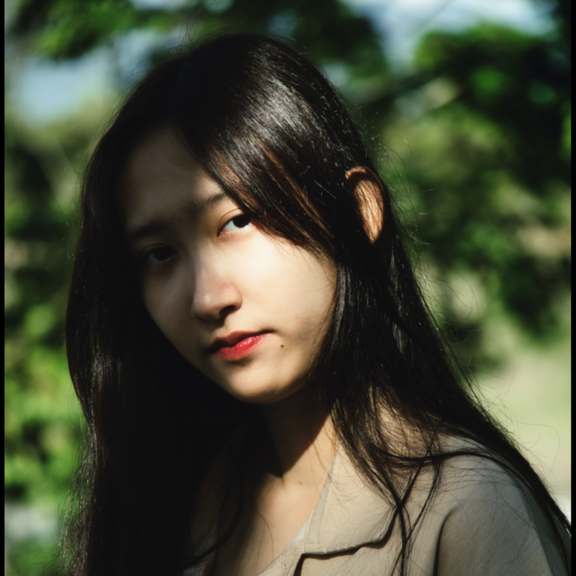} \\
\end{minipage}
\begin{minipage}[t]{0.15\linewidth}
\centering
\includegraphics[width=\linewidth]{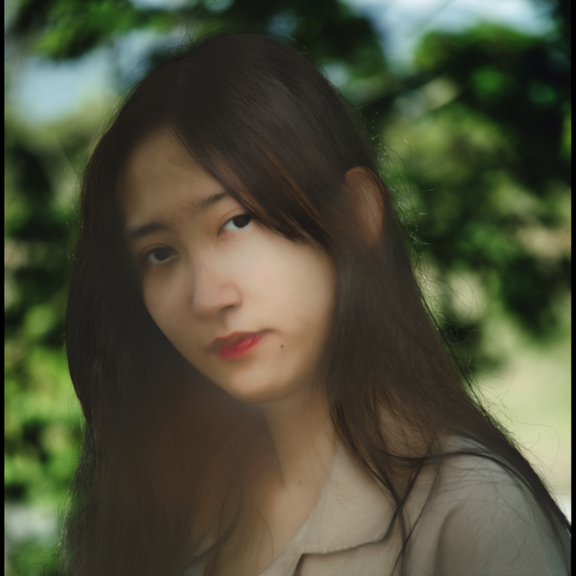} \\
\end{minipage}
\begin{minipage}[t]{0.15\linewidth}
\centering
\includegraphics[width=\linewidth]{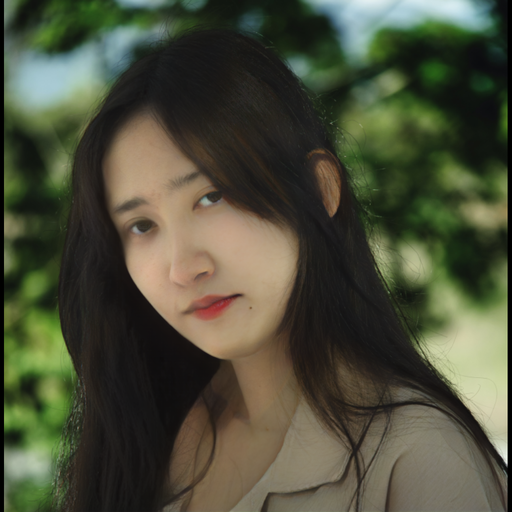} \\
\end{minipage}
\begin{minipage}[t]{0.15\linewidth}
\centering
\includegraphics[width=\linewidth]{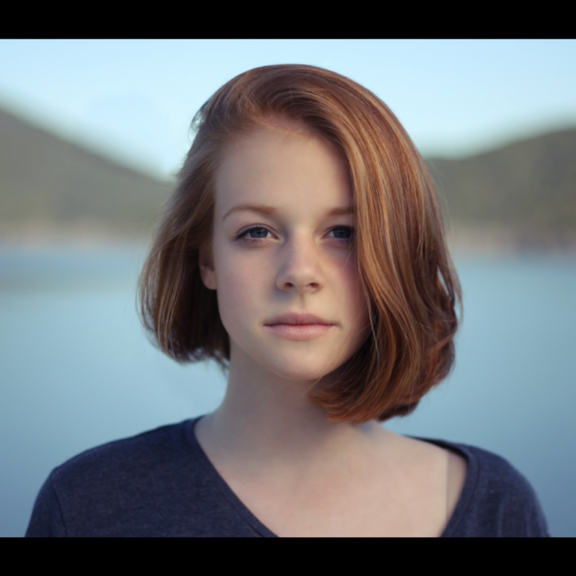} \\
\end{minipage}
\begin{minipage}[t]{0.15\linewidth}
\centering
\includegraphics[width=\linewidth]{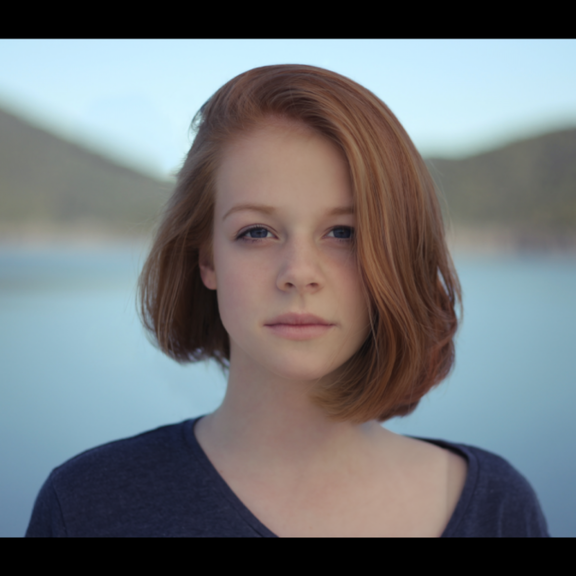} \\
\end{minipage}
\begin{minipage}[t]{0.15\linewidth}
\centering
\includegraphics[width=\linewidth]{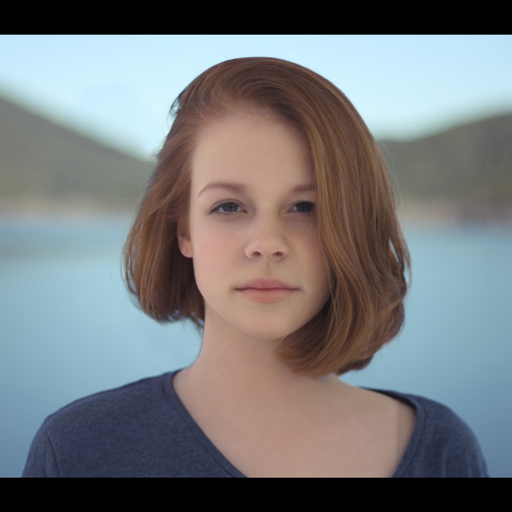} \\
\end{minipage}
\begin{minipage}[t]{0.15\linewidth}
\centering
\includegraphics[width=\linewidth]{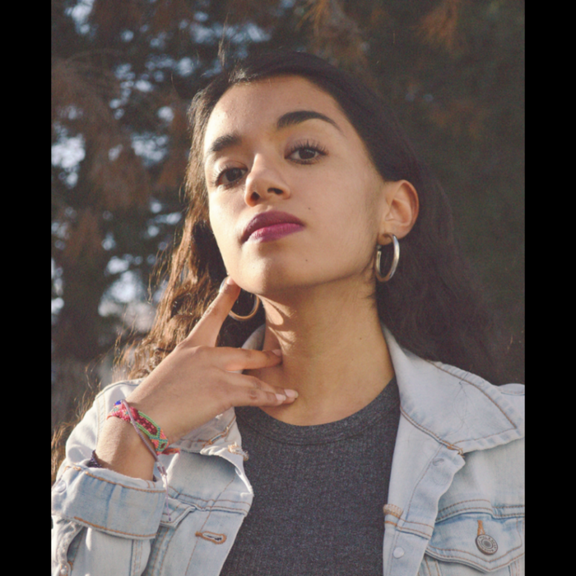} \\
\end{minipage}
\begin{minipage}[t]{0.15\linewidth}
\centering
\includegraphics[width=\linewidth]{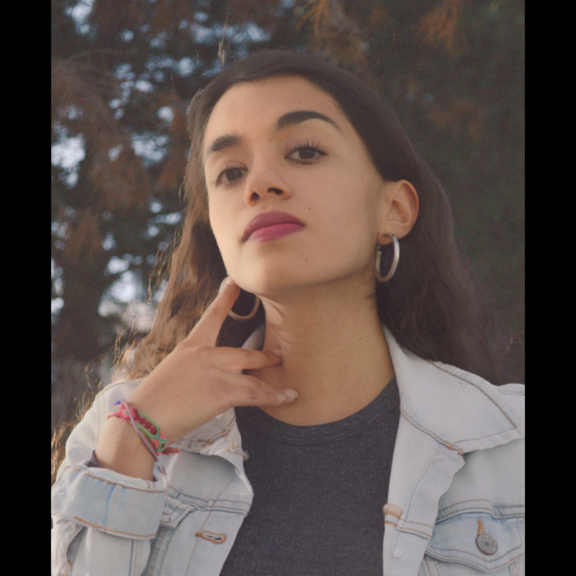} \\
\end{minipage}
\begin{minipage}[t]{0.15\linewidth}
\centering
\includegraphics[width=\linewidth]{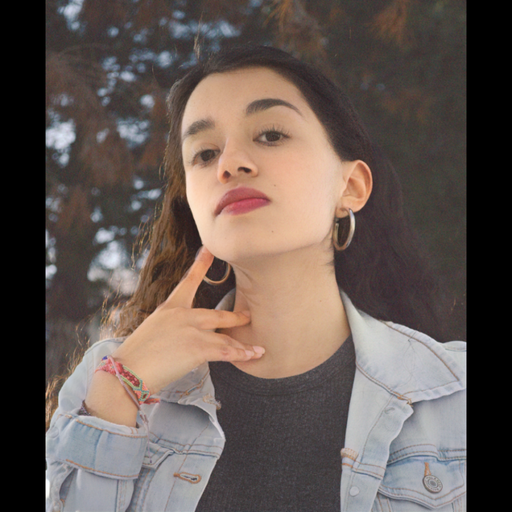} \\
\end{minipage}

\begin{minipage}[t]{0.15\linewidth}
\centering
\includegraphics[width=\linewidth]{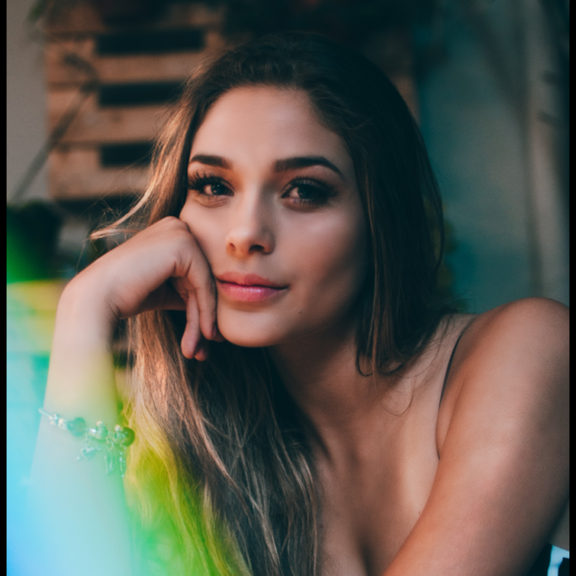} \\ 
\end{minipage}
\begin{minipage}[t]{0.15\linewidth}
\centering
\includegraphics[width=\linewidth]{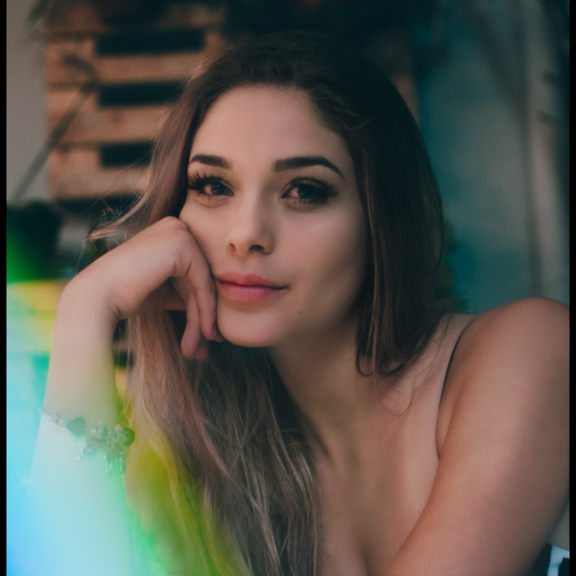} \\
\end{minipage}
\begin{minipage}[t]{0.15\linewidth}
\centering
\includegraphics[width=\linewidth]{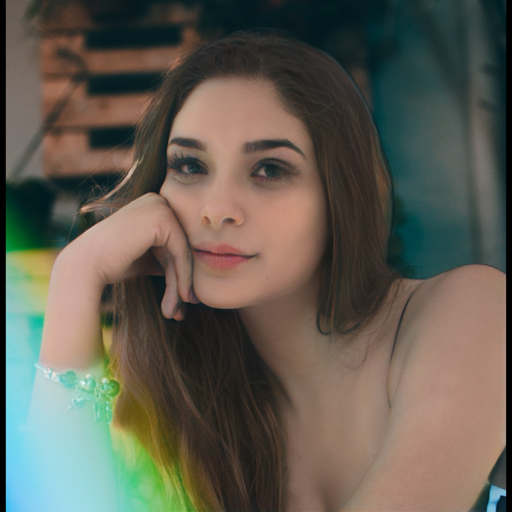} \\
\end{minipage}
\begin{minipage}[t]{0.15\linewidth}
\centering
\includegraphics[width=\linewidth]{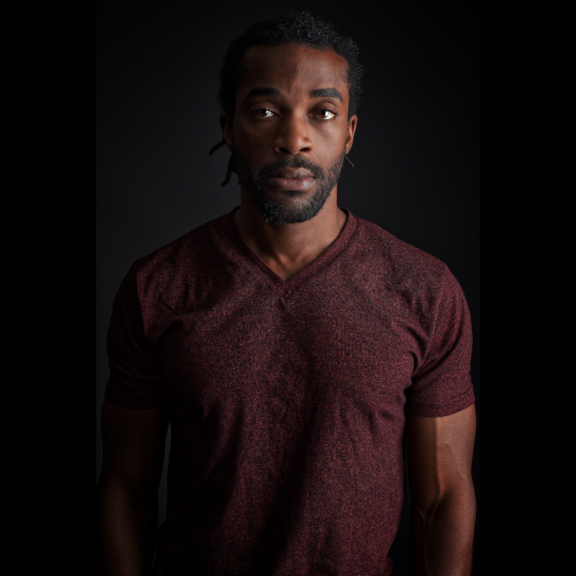} \\
\end{minipage}
\begin{minipage}[t]{0.15\linewidth}
\centering
\includegraphics[width=\linewidth]{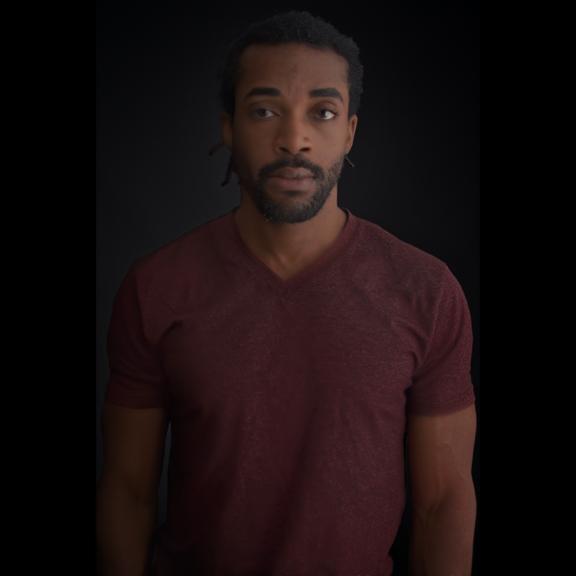} \\
\end{minipage}
\begin{minipage}[t]{0.15\linewidth}
\centering
\includegraphics[width=\linewidth]{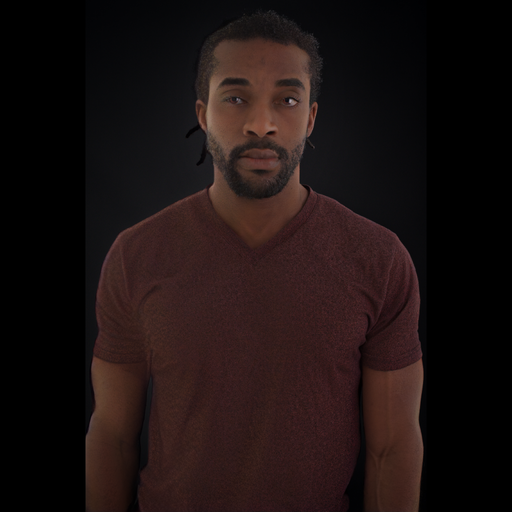} \\
\end{minipage}
\begin{minipage}[t]{0.15\linewidth}
\centering
\includegraphics[width=\linewidth]{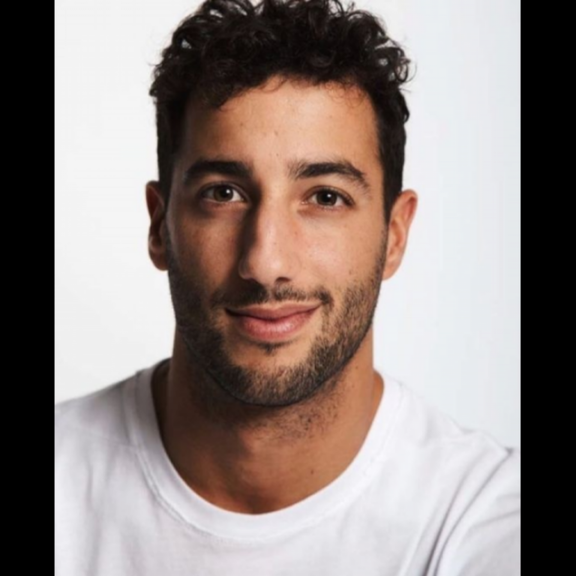} \\
\small a) Source
\end{minipage}
\begin{minipage}[t]{0.15\linewidth}
\centering
\includegraphics[width=\linewidth]{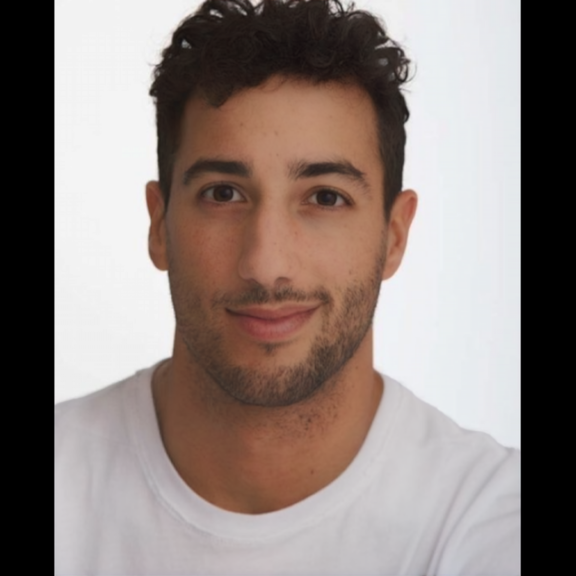} \\
\small b) ~\cite{controllable-light-diffusion}
\end{minipage}
\begin{minipage}[t]{0.15\linewidth}
\centering
\includegraphics[width=\linewidth]{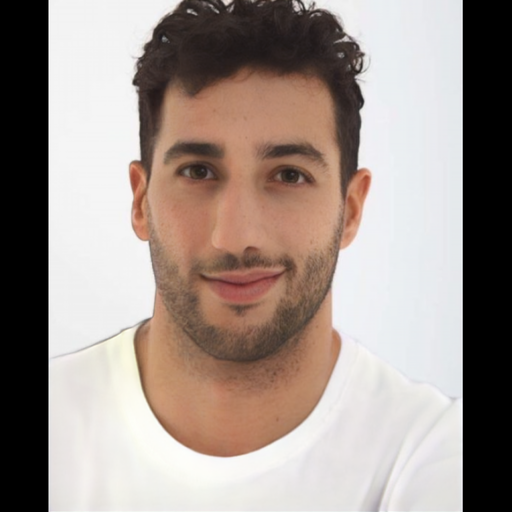} \\
\small c) Ours
\end{minipage}
\begin{minipage}[t]{0.15\linewidth}
\centering
\includegraphics[width=\linewidth]{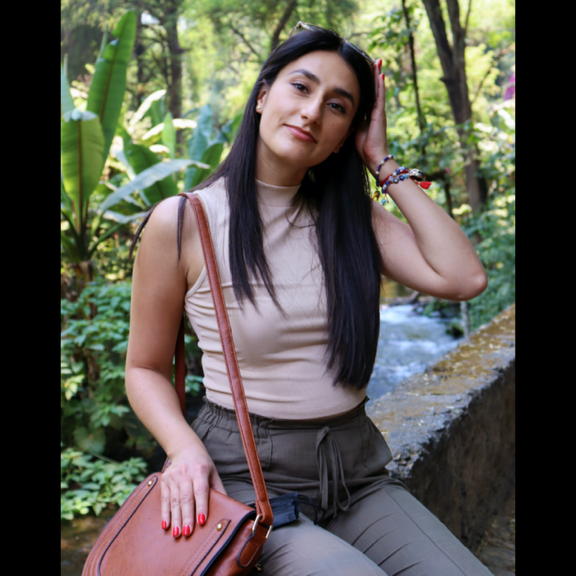} \\
\small d) Source
\end{minipage}
\begin{minipage}[t]{0.15\linewidth}
\centering
\includegraphics[width=\linewidth]{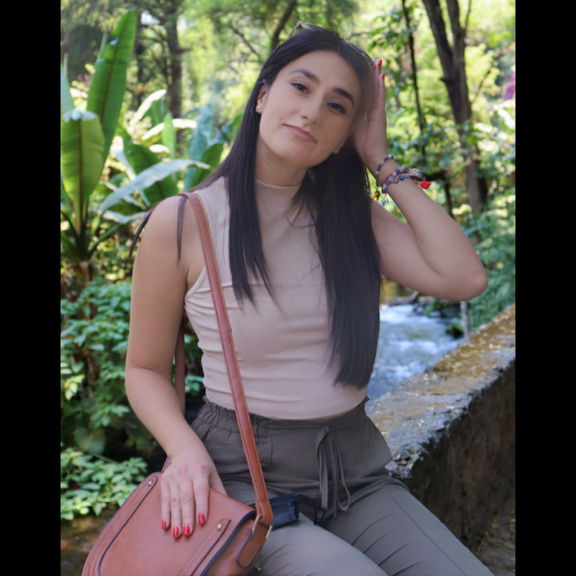} \\
\small e) ~\cite{controllable-light-diffusion}
\end{minipage}
\begin{minipage}[t]{0.15\linewidth}
\centering
\includegraphics[width=\linewidth]{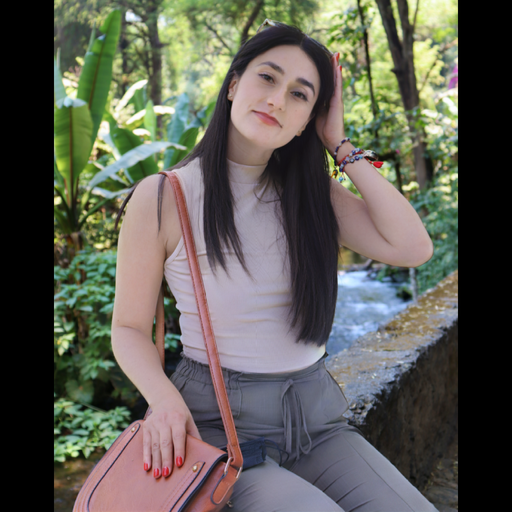} \\
\small f) Ours
\end{minipage}

\vspace{-2mm}
\caption{\small \textbf{Shadow Removal Comparisons with Futschik \textit{et al.}~\cite{controllable-light-diffusion}}. We demonstrate all qualitative shadow removal results against Futschik \textit{et al.}~\cite{controllable-light-diffusion} without cherry picking. COMPOSE is able to leave less of a shadow trace, especially for darker shadows when the goal is complete shadow removal. 
}\label{fig:ShadowRemoval}
\end{center}\vspace{-4mm}
\end{figure*}

\renewcommand{\thefigure}{11}
\begin{figure*}[t]
\vspace{-2mm}
\begin{center}

\begin{minipage}[t]{0.15\linewidth}
\centering
\includegraphics[width=\linewidth]{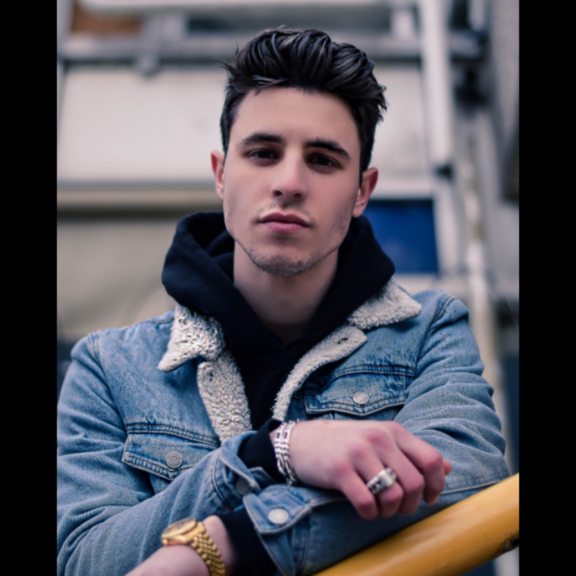} \\ 
\end{minipage}
\begin{minipage}[t]{0.15\linewidth}
\centering
\includegraphics[width=\linewidth]{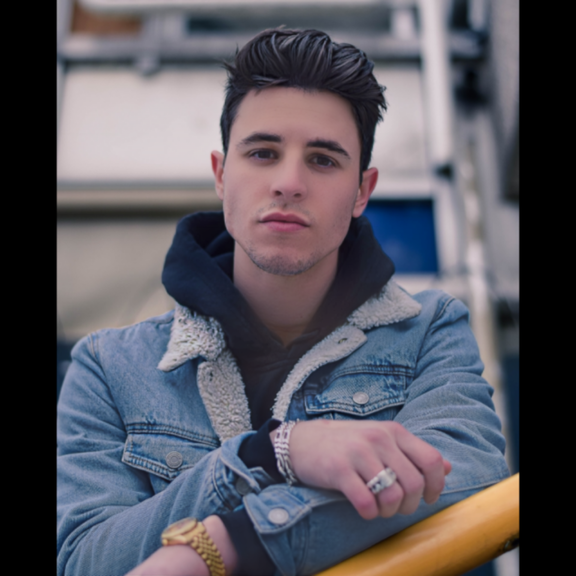} \\
\end{minipage}
\begin{minipage}[t]{0.15\linewidth}
\centering
\includegraphics[width=\linewidth]{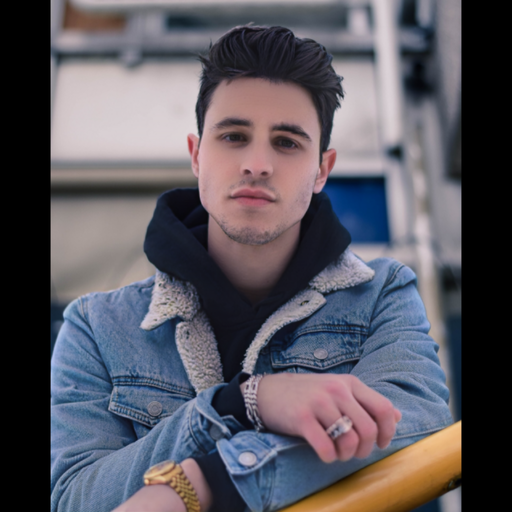} \\
\end{minipage}
\begin{minipage}[t]{0.15\linewidth}
\centering
\includegraphics[width=\linewidth]{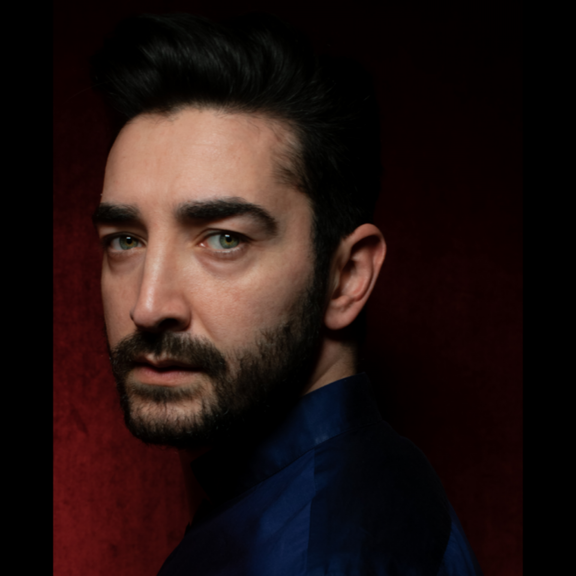} \\
\end{minipage}
\begin{minipage}[t]{0.15\linewidth}
\centering
\includegraphics[width=\linewidth]{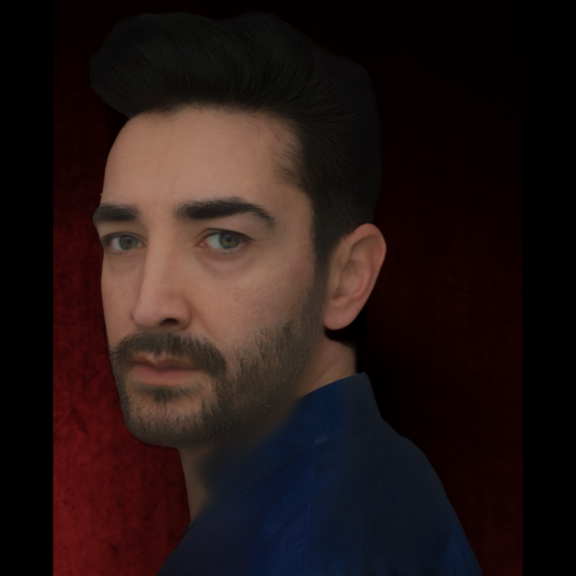} \\
\end{minipage}
\begin{minipage}[t]{0.15\linewidth}
\centering
\includegraphics[width=\linewidth]{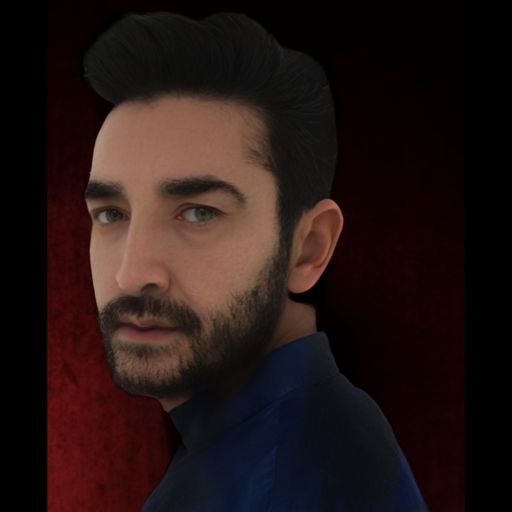} \\
\end{minipage}
\begin{minipage}[t]{0.15\linewidth}
\centering
\includegraphics[width=\linewidth]{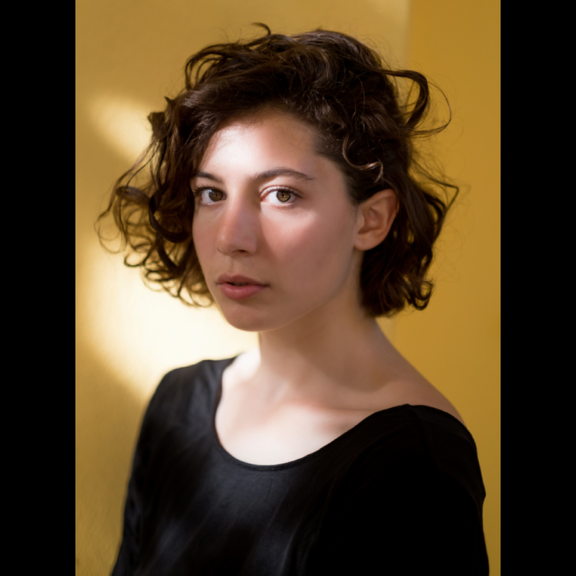} \\
\end{minipage}
\begin{minipage}[t]{0.15\linewidth}
\centering
\includegraphics[width=\linewidth]{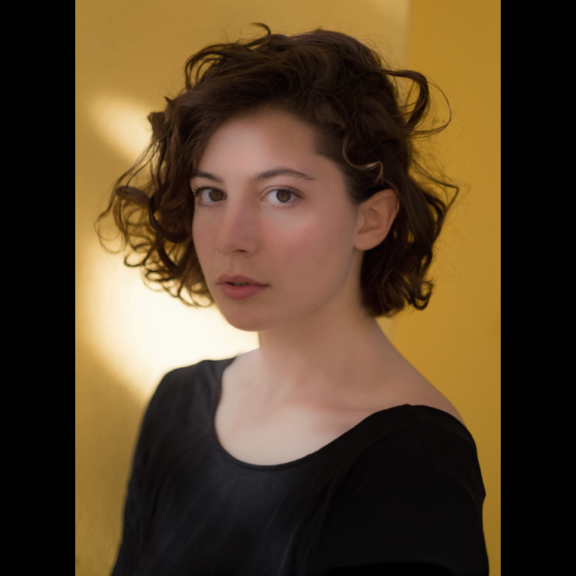} \\
\end{minipage}
\begin{minipage}[t]{0.15\linewidth}
\centering
\includegraphics[width=\linewidth]{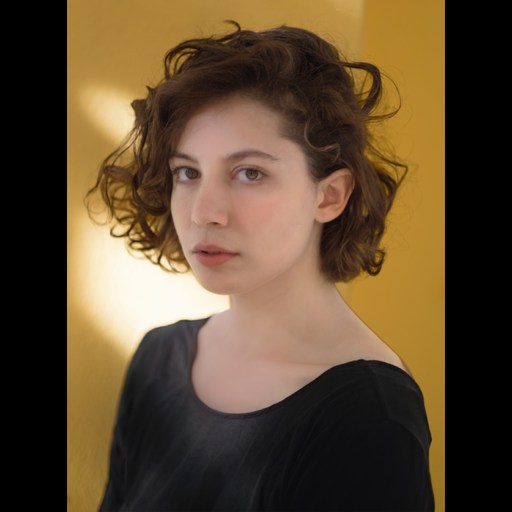} \\
\end{minipage}
\begin{minipage}[t]{0.15\linewidth}
\centering
\includegraphics[width=\linewidth]{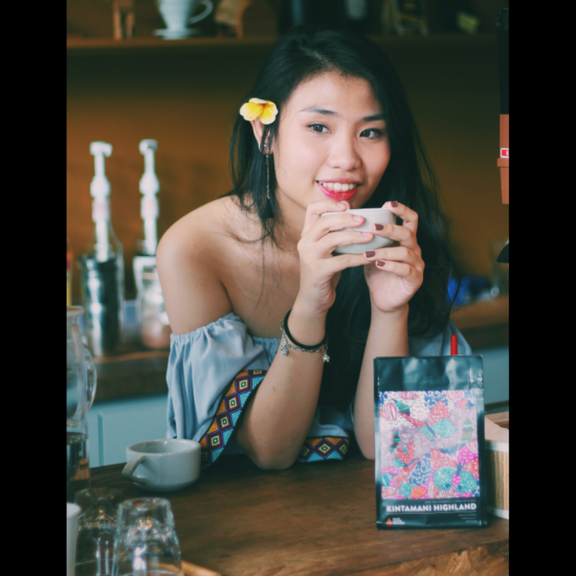} \\
\end{minipage}
\begin{minipage}[t]{0.15\linewidth}
\centering
\includegraphics[width=\linewidth]{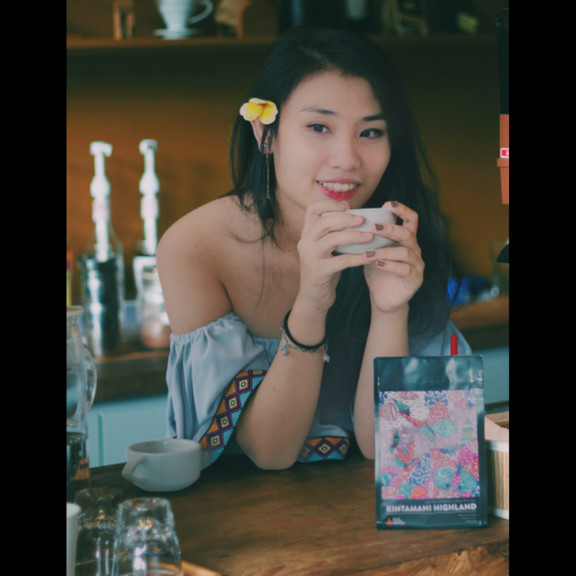} \\
\end{minipage}
\begin{minipage}[t]{0.15\linewidth}
\centering
\includegraphics[width=\linewidth]{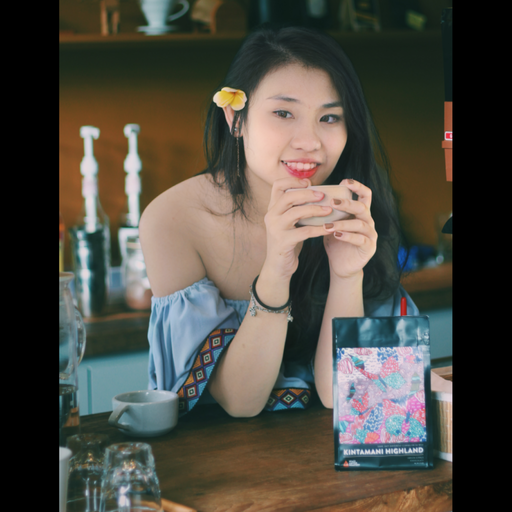} \\
\end{minipage}

\begin{minipage}[t]{0.15\linewidth}
\centering
\includegraphics[width=\linewidth]{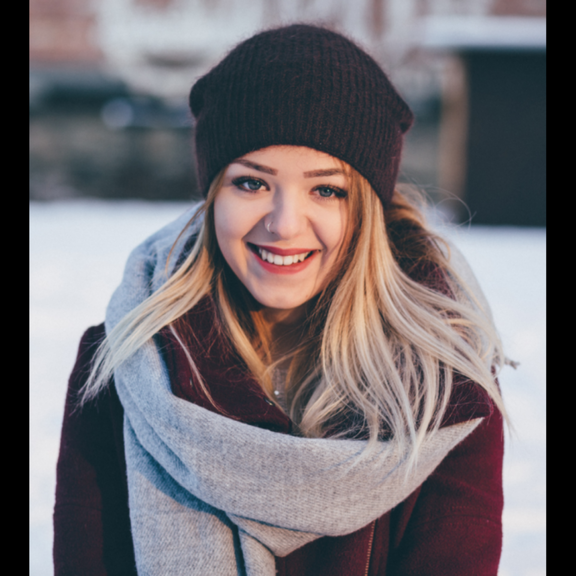} \\ 
\end{minipage}
\begin{minipage}[t]{0.15\linewidth}
\centering
\includegraphics[width=\linewidth]{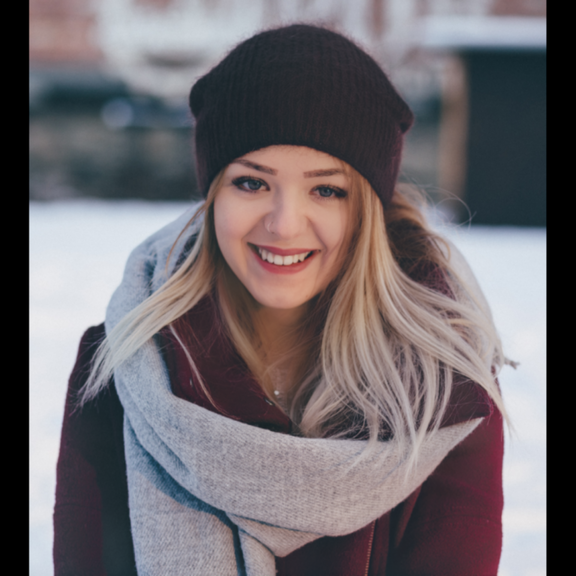} \\
\end{minipage}
\begin{minipage}[t]{0.15\linewidth}
\centering
\includegraphics[width=\linewidth]{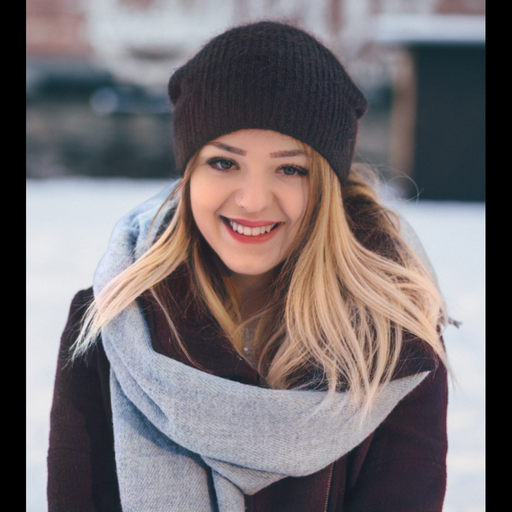} \\
\end{minipage}
\begin{minipage}[t]{0.15\linewidth}
\centering
\includegraphics[width=\linewidth]{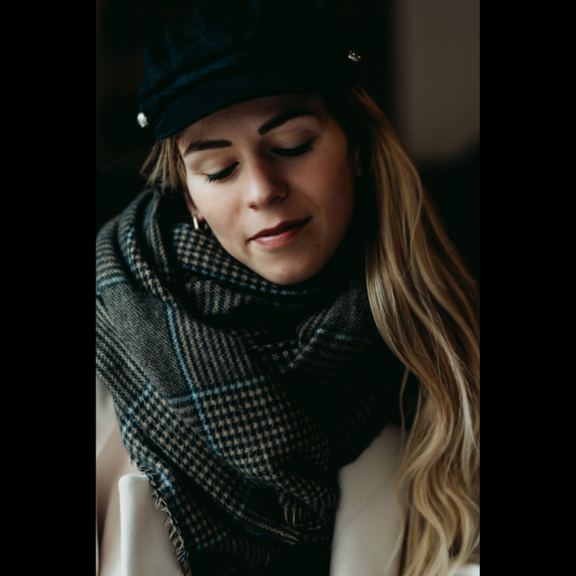} \\ 
\end{minipage}
\begin{minipage}[t]{0.15\linewidth}
\centering
\includegraphics[width=\linewidth]{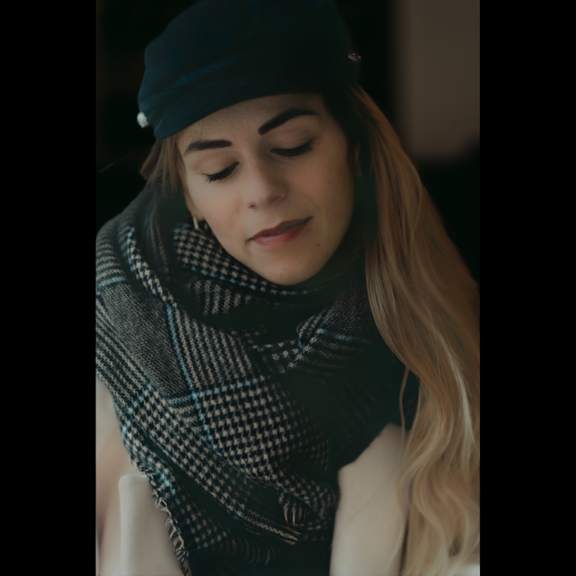} \\
\end{minipage}
\begin{minipage}[t]{0.15\linewidth}
\centering
\includegraphics[width=\linewidth]{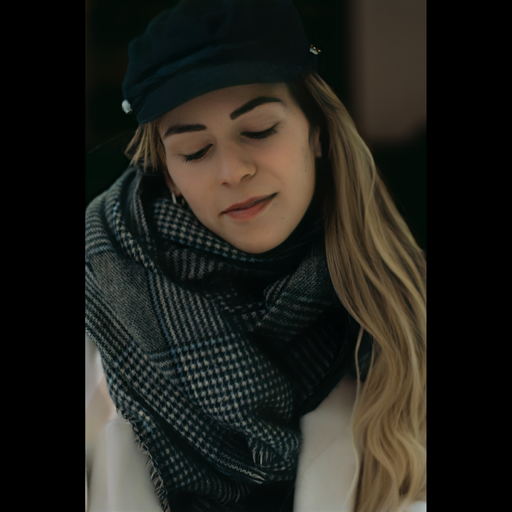} \\
\end{minipage}
\begin{minipage}[t]{0.15\linewidth}
\centering
\includegraphics[width=\linewidth]{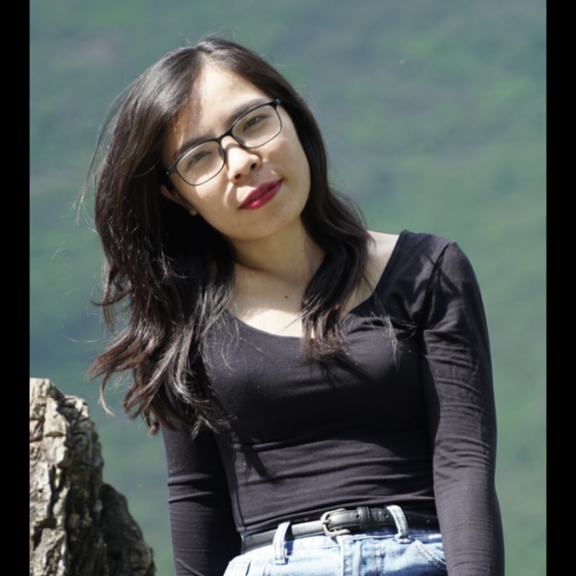} \\
\small a) Source  
\end{minipage}
\begin{minipage}[t]{0.15\linewidth}
\centering
\includegraphics[width=\linewidth]{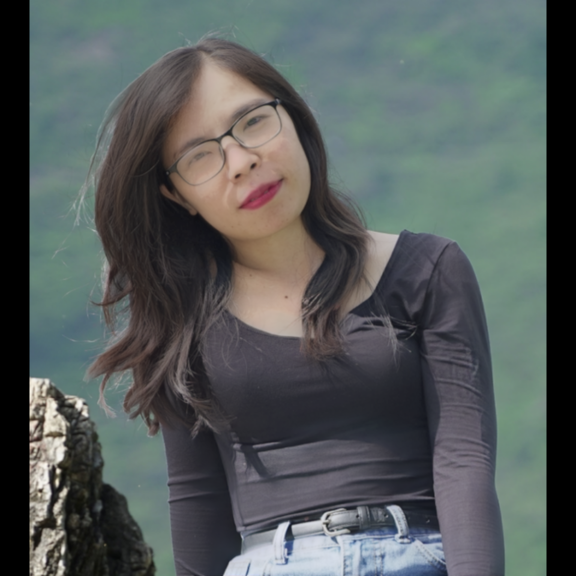} \\
\small b) ~\cite{controllable-light-diffusion}
\end{minipage}
\begin{minipage}[t]{0.15\linewidth}
\centering
\includegraphics[width=\linewidth]{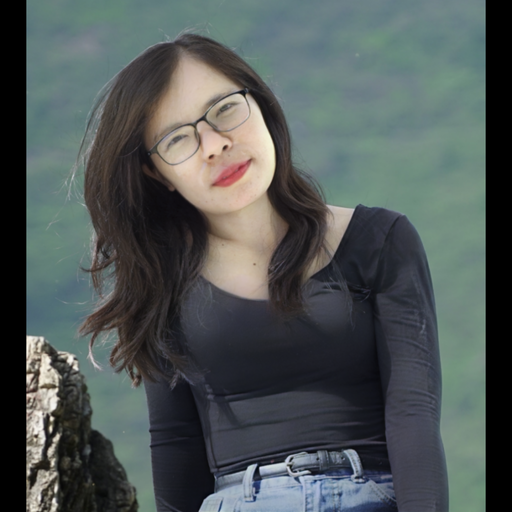} \\
\small c) Ours 
\end{minipage}
\begin{minipage}[t]{0.15\linewidth}
\centering
\includegraphics[width=\linewidth]{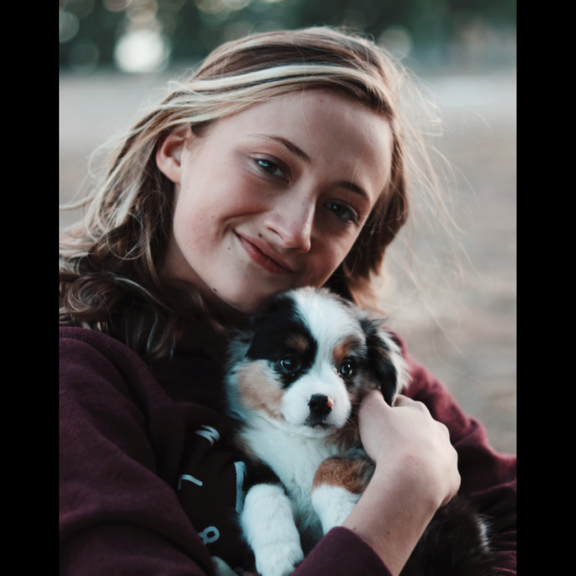} \\
\small d) Source  
\end{minipage}
\begin{minipage}[t]{0.15\linewidth}
\centering
\includegraphics[width=\linewidth]{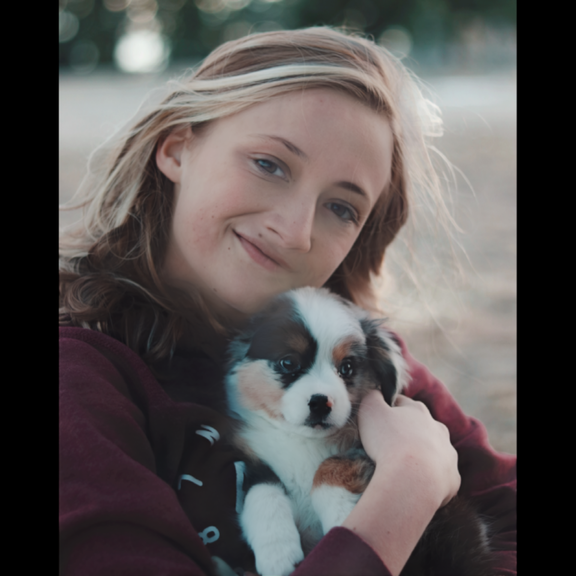} \\
\small e) ~\cite{controllable-light-diffusion}
\end{minipage}
\begin{minipage}[t]{0.15\linewidth}
\centering
\includegraphics[width=\linewidth]{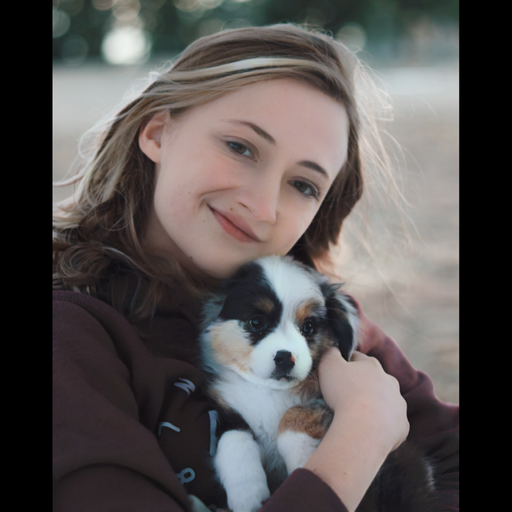} \\
\small f) Ours 
\end{minipage}

\vspace{-2mm}
\caption{\small \textbf{Shadow Removal Comparisons with Futschik \textit{et al.}~\cite{controllable-light-diffusion}}. We demonstrate all qualitative shadow removal results against Futschik \textit{et al.}~\cite{controllable-light-diffusion} without cherry picking. COMPOSE is able to leave less of a shadow trace, especially for darker shadows when the goal is complete shadow removal. 
}\label{fig:ShadowRemoval}
\end{center}\vspace{-4mm}
\end{figure*}

When comparing shadow softening performance with Futschik \textit{et al.}~\cite{controllable-light-diffusion}, we sent a total of $40$ in-the-wild images. As demonstrated in Fig.~\ref{fig:ShadowRemoval}, the main limitation in~\cite{controllable-light-diffusion} is that sometimes it leaves a noticeable shadow trace in the presence of darker shadows at level $0.0$, where the goal is to remove all shadows. In Fig.~\ref{fig:ShadowRemoval}, we show all $40$ shadow softening results compared to our method at level $0.0$. Across all $40$ evaluation images, our method is virtually spotless at removing shadows whereas several examples have shadow traces and artifacts for~\cite{controllable-light-diffusion}. This is largely thanks to our hierarchical transformer design for our light diffusion network, which better handles removing the effects of shadows at all scales compared to the U-Net design of~\cite{controllable-light-diffusion}. 

\section{Clarifications on Evaluation Protocols}

Here, we clarify more about our testing protocols to ensure that we perform fair comparisons in all of our evaluations. When comparing with Total Relighting (TR)~\cite{TotalRelighting} and Hou \textit{et al.}~\cite{face-relighting-with-geometrically-consistent-shadows} in Fig. $4$ and Tab. $1$ of the main paper, we ensure that we provide the most generous possible test setting. The results from TR were provided by the authors upon our request and we followed all instructions to properly prepare testing data. The environment maps were Gaussian lights of varying intensity/position/spread. We allowed the authors to tune the lighting scale to achieve their best possible results and provided sample relit images with their corresponding environment maps to help them measure the appropriate lighting scale to match TR's conventions. The results from Hou \textit{et al.} were also provided by the authors, and in their case the evaluation is simpler since they only model light direction. The qualitative comparisons with Futschik \textit{et al.}~\cite{controllable-light-diffusion} in Fig. $7$ on shadow softening, as mentioned in the main paper, are also provided by the authors. 

\section{Design Choices}

Here we further explain some of our design choices in the COMPOSE pipeline, including motivation and physical plausibility. 

\subsubsection{LDR Representation of Light} 

In our light estimation stage, we regress an LDR environment map instead of an HDR environment map. The reasons for this are that LDR is easier to regress than HDR during training due to the large, out-of-range peak values in HDR and we find LDR is sufficient for estimating light position in our pipeline (we are focused on the dominant light position and not as concerned with regressing every detail in the environment map). When performing gaussian fitting to find the dominant light, the algorithm will converge at roughly the same solution in determining the light center. 

\subsubsection{Single Dominant Light Assumption}

Our assumption of a single dominant light source in the scene stems from our focus on outdoor in-the-wild settings. There's usually one dominant Gaussian light outdoors, namely the sun. Our method can be extended to handle multiple lights by first changing the gaussian fitting step in the light estimation stage to enable fitting multiple gaussians (lights). This could be done using Gaussian Mixture Models (varying k) to determine the number of lights (k) using the lowest fitting error as the metric. Handling more diffuse lighting environments is another interesting direction. This can be done by leveraging the linearity of light to generate diffuse lightings as the sum of many directional lights, where each directional light involves one model inference of COMPOSE. 

\subsubsection{Image Compositing}

We further elaborate on the physical motivation of the fourth stage of COMPOSE: the compositing step. $\mathbf{I}_{D}$ and $\mathbf{I}_{S}$ represent the contributions of ambient and diffuse/directional light respectively. Compositing adjusts the contributions of each type of lighting, which is physically plausible by the Phong Model: $I=L_{amb}*R_{amb}+L_{diff}*R_{diff}+L_{spec}*R_{spec}$. 

\subsubsection{Four Stages}

One question that may arise is whether COMPOSE truly needs to be a four-stage model, which to some may seem unnecessarily complex. However, we argue in favor of this design for several reasons. One is that the inference time of a multi-stage model compared to a single end-to-end pipeline should be similar. Either way, light estimation, delighting, and relighting modules will be required. The advantage of our multi-stage pipeline is greater flexibility for users in terms of applications (\textit{e.g.}, using our light diffusion network alone for shadow removal). It's also easier to integrate components of COMPOSE into other shadow editing and relighting methods. The design is modular and each component can be improved or replaced. This enables better disentanglement, controllability, and explainability during shadow editing. It's also easier to identify which component is causing problems if the shadow editing result is poor. 

\section{Lighting Estimation}

Our lighting estimation stage is primarily responsible for providing the user with an accurate estimation of where the dominant light source of the input image $\mathbf{I}_{N}$ is located. This is useful if the user wants to perform any kind of in-place shadow editing that does not involve changing the light position (\textit{e.g.} shrinking the light, enlarging the light, softening the shadow, and intensifying the shadow). It is important to note that our pipeline is fairly error tolerant in this stage since the user is free to tune the lighting estimation in subsequent stages to achieve better results (\textit{e.g.} tuning the dominant light position in the environment map to better match the shadow positions in the source image). 

\section{Responsible Human Dataset Usage}

We collect and use light stage data, as well as images from Unsplash and Adobe Stock. For the light stage data, we have acquired permission from both the subjects and the capturing studio to include these images in research papers. Unsplash's license mentions photos can be downloaded and used for free for both commercial and non-commercial purposes. For Adobe Stock, we paid for any images that we used and are thus permitted to include them in our submission. Finally, for the subject appearing in Figs. $1$ and $6$, the source image is simply acquired from a colleague who participated in a lighting experiment and we have received permission from them personally to include their photos for publication. In all cases, we collect or use portrait images only that contain no additional personally identifiable information or offensive content. 
\end{document}